\documentclass[10pt,twocolumn,letterpaper]{article}

\usepackage{ijcb}
\usepackage{times}
\usepackage{epsfig}
\usepackage{graphicx}
\usepackage{amsmath}
\usepackage{amssymb}

\usepackage{times}
\usepackage{epsfig}
\usepackage{graphicx}

\usepackage{caption}
\usepackage{makecell}
\usepackage{booktabs}
\usepackage{subfig}
\usepackage{multirow}

\usepackage{multicol}
\usepackage{enumitem}

\usepackage{url}

\usepackage{breakurl}
\usepackage[breaklinks]{hyperref}

\usepackage{algorithm}
\usepackage[noend]{algpseudocode}

\usepackage{adjustbox} 

\newcommand\Mark[1]{\textsuperscript#1}

\ijcbfinalcopy 

\ifijcbfinal\fi
\begin{document}

\title{Face Quality Estimation and Its Correlation to Demographic and Non-Demographic Bias in Face Recognition}

\author{Philipp Terh\"{o}rst\Mark{1}\Mark{2}, Jan Niklas Kolf\Mark{2}, Naser Damer\Mark{1}\Mark{2}, Florian Kirchbuchner\Mark{1}\Mark{2}, Arjan Kuijper\Mark{1}\Mark{2}\\
\Mark{1}Fraunhofer Institute for Computer Graphics Research IGD, Darmstadt, Germany\\
\Mark{2}TU Darmstadt, Darmstadt, Germany\\
Email:{\{philipp.terhoerst, naser.damer, florian.kirchbuchner, arjan.kuijper\}@igd.fraunhofer.de}
}


\maketitle
\thispagestyle{empty}

\vspace{-3mm}
\begin{abstract}
\vspace{-1mm}

Face quality assessment aims at estimating the utility of a face image for the purpose of recognition.
It is a key factor to achieve high face recognition performances.
Currently, the high performance of these face recognition systems come with the cost of a strong bias against demographic and non-demographic sub-groups.
Recent work has shown that face quality assessment algorithms should adapt to the deployed face recognition system, in order to achieve highly accurate and robust quality estimations.
However, this could lead to a bias transfer towards the face quality assessment leading to discriminatory effects e.g. during enrolment.
In this work, we present an in-depth analysis of the correlation between bias in face recognition and face quality assessment.
Experiments were conducted on two publicly available datasets captured under controlled and uncontrolled circumstances with two popular face embeddings.
We evaluated four state-of-the-art solutions for face quality assessment towards biases to pose, ethnicity, and age.
The experiments showed that the face quality assessment solutions assign significantly lower quality values towards subgroups affected by the recognition bias demonstrating that these approaches are biased as well.
This raises ethical questions towards fairness and discrimination which future works have to address.
\end{abstract}

\vspace{-4mm}
\section{Introduction}
\vspace{-1mm}












Face recognition systems are spreading worldwide and have a growing effect on everybody's daily life.
Furthermore, these systems are increasingly involved in critical decision-making processes, such as in forensics and law enforcement.
Current biometric solutions are mainly optimized for maximum overall accuracy  \cite{Jain201650YO} and are heavily biased for certain demographic groups \cite{faceAccurate,DBLP:journals/corr/abs-1809-02169,Furl2002FaceRA,Phillips:2011:OEF:1870076.1870082,pmlr-v81-buolamwini18a, garvie2016perpetual}.
The performance of face recognition is driven by the quality of its captures \cite{DBLP:journals/corr/Best-RowdenJ17}.
Biometric sample quality is defined as the utility of a sample for the purpose of recognition \cite{DBLP:journals/corr/abs-1904-01740, 6712715, 10.1007/978-3-540-74549-5_26, DBLP:journals/corr/Best-RowdenJ17} and is crucial for many applications.
Recent work \cite{CVPR_Terhoerst} has shown that the accuracy and the robustness of face quality estimation can be enhanced drastically by adapting the face quality assessment algorithm to the deployed face recognition model.
However, this can lead to biased face quality assessment algorithms as well.

There are several political regulations to prevent discriminatory decisions.
Article 14 of the European Convention of Human Rights and Article 7 of the Universal Declaration of Human Rights ensure people the right to non-discrimination.
Also the General Data Protection Regulation (GDPR) \cite{Voigt:2017:EGD:3152676} aims at preventing discriminatory effects (article 71).
In spite of these political efforts, several works \cite{Phillips:2011:OEF:1870076.1870082,pmlr-v81-buolamwini18a,faceAccurate,DBLP:journals/corr/abs-1809-02169,Furl2002FaceRA, garvie2016perpetual} showed that open-source \cite{faceAccurate, DBLP:conf/aaai/SernaMFCOR20} as well as commercial \cite{pmlr-v81-buolamwini18a} face recognition solutions are strongly biased towards different demographic groups. The more accurate terms of “differential performance” and “differential outcome” were presented in \cite{BTAS2019_demo} to avoid the unintended interpretation of bias by policy makers and statisticians. Based on these terms, a number of recent works are supporting the notion of differential performance in face recognition systems \cite{FRVT2019, 8636231}

Face quality assessment solutions can possess intended and unintended kinds of biases, e.g. non-demographic and demographic bias.
While non-demographic bias enhances the quality estimation process without discriminative consequences, transferring demographic bias unintentionally to face quality assessment algorithms can have a serious impact on society.
During the enrolment of an individual or for quality-based fusion approaches (e.g. in surveillance scenarios), face quality assessment is needed.
Consequently, a transferred bias to the quality estimation will directly increase discriminative decisions of such quality-based subsystems. 
Moreover, in the operation, face quality estimation can be used as a separate processing step \cite{ISO} and can be trained while having in mind a face recognition system different than the one used in the field. 
Therefore, having a biased quality estimation can add to the bias of the face recognition system, as it might have different biases. 

In this work, we present a detailed analysis of the correlation between bias in face recognition systems and the corresponding face quality assessment.
To the best of our knowledge, this is the first work analysing this relation.
The experiments were conducted on two publicly available datasets under
diverse image capturing conditions.
The correlation analysis was done using two different face recognition solutions with four state-of-the-art face quality assessment algorithms from academia and industry.
Investigating different head poses, ethnicities, and age classes, we found degraded performances, and thus biases, towards certain subclasses for both face recognition systems.
The experiments demonstrated a strong correlation between face recognition bias and face quality assessment.
Face images from the classes affected by the bias were estimated with lower quality values than unbiased images.
Consequently, the bias is transferred to the quality assignment process.

The goal of this work is to point out that current face image quality assessment approaches have to deal with similar bias-related problems than in face recognition.
We point out that the quality of a face image points out a biased ground of a faulty decision. Especially in a controlled environment, such as ABC gates where the image is of good quality, a low face quality must alarm the operator to a high probability of a faulty decision, whether a false match or a false non-match, which might require manual inspection. This faulty decision can be a bias issue given the controlled capture conditions.

\section{Related work}
\label{sec:RelatedWork}

\subsection{Bias in face recognition}
In face biometrics, the bias is usually induced by non-equally distributed classes in training data \cite{Kortylewski_2019_CVPR_Workshops, DBLP:journals/corr/abs-1806-00194}.
Klare et al. \cite{6327355} showed that the performance of face recognition algorithms is strongly influenced by demographic attributes.
In \cite{pmlr-v81-buolamwini18a,faceAccurate,DBLP:conf/aaai/SernaMFCOR20,8636231,FRVT2019, drozdowski2020demographic}, the authors came to the same conclusions for commercial and open-sources face recognition algorithms.
They demonstrated that the person's gender and ethnicity strongly determine their recognition performance.

These findings motivated research towards mitigating bias in face recognition approaches.
For more unbiased face recognition, Zhang and Zhou \cite{Zhang:2010:CFR:1850487.1850593} formulate the face verification problem as a multi-class cost-sensitive learning task and demonstrated that this approach can reduce different kinds of faulty decisions of the system.
In 2017, range loss \cite{DBLP:conf/iccv/ZhangFWLQ17} was proposed to learn robust face representations that can deal with long-tailed training data.
It is designed to reduce overall intrapersonal variations while enlarging interpersonal differences simultaneously.
Recent works published in 2019 aimed at mitigating bias in face recognition through adversarial learning \cite{gong2019debface,Liang_2019_CVPR}, margin-based approaches \cite{wang2019mitigate,DBLP:journals/corr/abs-1806-00194}, or data augmentation \cite{WANG201935,Kortylewski_2019_CVPR_Workshops,DBLP:conf/cvpr/00010S0C19}.

In \cite{gong2019debface}, Gong, Liu, and Jain proposed de-biasing adversarial network.
This network consists of one identity classifier and 3 demographic classifiers 
and learns disentangled feature representations for unbiased face recognition.
Liang et al. \cite{Liang_2019_CVPR} proposed a two-stage method for adversarial bias mitigation.
First, they learn disentangled representations by a one-vs-rest mechanism and second, they enhance the disentanglement by additive adversarial learning.

Also margin-based approaches were proposed to reduce bias in face recognition systems.
In \cite{wang2019mitigate}, Wang et al. apply reinforcement learning to determine a margin that minimizes ethnic bias.
Huang et al. \cite{DBLP:journals/corr/abs-1806-00194} proposed a cluster-based large-margin local embedding approach to reduce the effect of local data imbalance and thus, aims at reducing bias coming from unbalanced training data.

Finally, data augmentation methods were presented for fairer face recognition.
In \cite{WANG201935}, Wang et al. proposed a large margin feature augmentation to balance class distributions.
Kortylewski et al. \cite{Kortylewski_2019_CVPR_Workshops} proposed a data augmentation approach with synthetic data generation and Yin et al. \cite{DBLP:conf/cvpr/00010S0C19} proposed a center-based feature transfer framework to augment under-represented samples.
Although these works were able to reduce decision bias, enhancing this reduction is still an active research topic.


\subsection{Face quality assessment}
Several standards have been proposed to insure face image quality by constraining the capture requirements, such as ISO/IEC 19794-5 \cite{ISO19794-5-2011} and ICAO 9303 \cite{ICAO9303}.
In these standards, quality is divided into \textit{image-based} qualities  (such as illumination, occlusion) and \textit{subject-based} quality measures (such as pose, expression, accessories).
These mentioned standards influenced many face quality assessment approaches that have been proposed recently.

The first generation of face quality assessment algorithms define quality metrics based on image quality factors \cite{10.1007/978-3-540-74549-5_26, 6197711, 7935089, 5424029, 6712715, 6985846, 4341617, 6460821, 6996248}.
However, these approaches have to consider every possible factor manually, and since humans may not know the best characteristics for face recognition systems, recent research focuses on learning-based approaches.

End-to-end learning approaches for face quality assessment were first presented in 2011.
Aggarwal et al. \cite{5981784} proposed an approach for predicting the face recognition performance using a multi-dimensional scaling approach to map space characterization features to genuine scores.
In \cite{5981881}, a patch-based probabilistic image quality approach was designed to work on 2D discrete cosine transform features and trains a Gaussian model on each patch.
In 2015, a rank-based learning approach was proposed by Chen et al. \cite{6877651}.
They define a linear quality assessment function with polynomial kernels and train weights based on a ranking loss.
In \cite{7351562}, face quality assessment was performed based on objective and relative face image qualities.
While the objective quality metric refers to objective visual quality in terms of pose, alignment, blurriness, and brightness, the relative quality metric represents the degree of mismatch between training face images and a test face image.
Best-Rowden and Jain \cite{DBLP:journals/corr/Best-RowdenJ17} proposed an automatic face quality prediction approach in 2018.
They proposed two methods for quality assessment of face images based on (a) human assessments of face image quality and (b) quality values from similarity scores.
Their approach (b) is based on support vector machines applied to deeply learned representations.
In 2019, Hernandez-Ortega et al. proposed FaceQnet \cite{DBLP:journals/corr/abs-1904-01740}, which adapts the quality label generation from Best-Rowden and applies it to fine-tune a face recognition neural network to predict face qualities in a regression task.
Stochastic embedding robustness (SER-FIQ) is a novel face image quality measurement concept proposed in \cite{CVPR_Terhoerst}.
Their method determines the embedding variations generated from random subnetworks of the deployed face recognition model.
The magnitude of these variations define the robustness and thus, the quality.
Their method avoids the need for training and further allows to take into account the decision patterns of the deployed face recognition model.

So far, the best quality estimates were achieved when the systems adapt to the utilized face recognition model.
However, there is a risk of transferring the face recognition bias towards the quality assessment.
Therefore, this work analyses the correlation between face quality assessment and face recognition bias.
To the best of our knowledge, this is the first work that analyses this relationship and its implications on the real use of the technology.

%
%
%
%
%
%
%
%
%
%

\section{Evaluated face quality assessment solutions}

Face quality assessment aims at estimating the usability of an image for the purpose of recognition \cite{DBLP:journals/corr/abs-1904-01740, 6712715, 10.1007/978-3-540-74549-5_26, DBLP:journals/corr/Best-RowdenJ17}.
For our correlation study between face quality and face recognition bias, we choose the four of the latest face quality assessment approaches from academia and industry.
These approaches will be shortly discussed in the following.

\subsection{COTS}
COTS \cite{COTS} is an off the shelf industry product from Neurotechnology, the used version is published in 2019.
Unfortunately, it only provides the application and does not provide any information about its working principles.
However, in \cite{CVPR_Terhoerst}, the authors show that COTS predicted quality synchronise well with FaceNet \cite{DBLP:journals/corr/SchroffKP15} performance, and to a much lower degree with ArcFace \cite{Deng_2019_CVPR} performance.

\subsection{Best-Rowden}
In 2018, Best-Rowden and Jain \cite{DBLP:journals/corr/Best-RowdenJ17} presented two approaches to face quality estimation, with and without human assessments.
We evaluate their approach based on quality labels coming from comparison scores, because the features and comparison scores are matcher dependent and thus, it adapts to the deployed face recognition model.
They define a quality label for query $j$ of subject $i$ as
\begin{align}
z_{ij} = \left( s_{ij}^G - \mu_{ij}^I \right) / \sigma_{ij}^I, \label{eq:BestRowden}
\end{align}
where $s_{ij}^G$ is the genuine score and $\mu_{ij}^I$ and $\sigma_{ij}^I$ are the mean and the standard deviation of the imposter scores.
They use the face embeddings of the deployed face recognition model and, based on these features, they train a support vector regressor to estimate the quality score of an input image.
Following their methodology, we train this approach on the MORPH \cite{1613043} dataset.
The hyperparameters are determined beforehand by a 5-fold cross-validation on this dataset.

\subsection{FaceQnet}
FaceQnet \cite{DBLP:journals/corr/abs-1904-01740} by Hernandez-Ortega et al. was published in 2019.
They adapted the idea of using the comparison score labels (see Equation \ref{eq:BestRowden}) from Best-Rowden et al. and combined them with a ResNet-based deep neural network structure.
Their approach is based on FaceNet embeddings and is trained on VGGFace2 \cite{Cao18}.
In \cite{CVPR_Terhoerst}, it was shown that even if the approach was trained on FaceNet embeddings, FaceQnet shows better synchronisation with ArcFace \cite{Deng_2019_CVPR} performance, indicating some overfitting on FaceNet \cite{DBLP:journals/corr/SchroffKP15} embeddings.
For our experiments, we used the pretrained FaceQnet model\footnote{\url{https://github.com/uam-biometrics/FaceQnet}} provided by the authors.

\subsection{SER-FIQ}


Stochastic embedding robustness (SER) is a face image quality (FIQ) estimation concept presented in \cite{CVPR_Terhoerst}, which avoids the use of inaccurate quality labels.
They defined face image quality based on the robustness of deeply learned features.
Calculating the variations of embeddings coming from random subnetworks of the deployed face recognition model, their solution defines the magnitude of these variations as a robustness measure, and thus, image quality.
Given an input image $I$ and the deployed face recognition model $\mathcal{M}$, their method applies $m=100$ different dropout patterns \cite{Srivastava:2014:DSW:2627435.2670313} to the neural network.
This results in $m$ random subnetworks of $\mathcal{M}$. 
Each of these networks produces a stochastic embedding $x_i$.
The quality
\begin{align}
q(I) = 2 \, \sigma \Big( - \dfrac{2}{m^2} \sum_{i<j} d(x_i, x_j) \Big),
\end{align}
of an input $I$ is then defined as the sigmoid of the negative mean Euclidean distance $d(x_i, x_j)$ between all stochastic embedding pairs.
A greater variation between the stochastic embeddings indicates a lower robustness of the representation and thus, a lower sample quality $q$.
Lower variations between the stochastic embeddings indicate a high robustness in the embedding space and are considered as a high sample quality $q$.
Since it can be directly applied on the deployed face recognition model, it completely avoids any training and further adapts to the decision patterns of the model.
The authors showed that this concept leads to significantly better quality estimations than previous work.
We follow their procedure that applies the dropout pattern repetitively on the last layer of the face recognition network.

\section{Experimental setup} 
\label{sec:ExperimentalSetup}

\paragraph{Database}
To evaluate the correlation between face quality assessment and bias in face recognition systems under controlled and unconstrained conditions, we conducted experiments on the two publicly available datasets, ColorFERET \cite{ColorFERET} and Adience \cite{Eidinger:2014:AGE:2771306.2772049}.
ColorFERET \cite{ColorFERET} consists of 14k images of 1.2k different individuals with different poses under controlled conditions.
The dataset further includes a variety of face poses, facial expressions, and lighting conditions.
The Adience dataset \cite{Eidinger:2014:AGE:2771306.2772049} consists of over 26.5k images of over 2.2k different individuals in an unconstrained environment.
Both databases contain information about identity, gender and age.
ColorFeret also provides labels regarding the subject's ethnicities and head posses.
In the experiments, this information is used to investigate how face quality assessment algorithms affect the recognition performance under diverse circumstances. 

The investigated face quality assessment solutions are based on three databases, MORPH \cite{1613043}, VGGFace2 \cite{Cao18}, and MS1M \cite{DBLP:journals/corr/GuoZHHG16}.
MORPH \cite{1613043} contains 55k frontal face images of more than 13k individuals.
80.4\% of the faces belong to the ethnicity black, 19.2\% to white, and 0.4\% to others.
The individual's age vary from 16-77 years. 79.4\% of the faces are within an age-range of $[20,50]$.
The VGGFace2 \cite{Cao18} database contains faces from over 9k subjects with over 3 million images.
The dataset contains a large variety of pose, age, and ethnicity.
Over 40\% of the face are frontal and over 50\% are half-frontal.
Most images belong to individuals over 18 years old and around 40\% belong to the age group of $[25,34]$.
The MS1M \cite{DBLP:journals/corr/GuoZHHG16} contains over 100k subjects with 10 million images.
The faces cover a large variance of age.
Over 50\% of the faces belong to white subjects.
The faces are mostly frontal.
This information will be used to discuss the influence of the training data on quality predictions.


\paragraph{Evaluation metrics}
In order to evaluate the face quality assessment performance, we follow the methodology by Grother et al. \cite{DBLP:journals/pami/GrotherT07} using error versus reject curves.
These curves show the verification error-rate (y-axis) achieved when unconsidering a certain percentage of face images (x-axis).
Based on the predicted quality values, these unconsidered images are these with the lowest predicted quality and the error rate is calculated on the remaining images.
Error versus reject curves indicate good quality estimation when the verification error decreases consistently when increasing the ratio of unconsidered images.

In order to prove that a face recognition system is biased towards some classes, the verification error is reported for all classes.
The verification error is reported in terms of false non-match rate (FNMR) at fixed false match rates (FMR).
The FMR is reported at 0.1\% FMR threshold as recommended by the best practice guidelines for automated border control of European Border Guard Agency Frontex \cite{FrontexBestPractice}.
To show the correlation between face quality assessment and biased face verification performance, 
the proportion of subgroups is continuously analysed over quality thresholds.
The proportion of biased subgroups will decrease fast if the face quality assessment algorithm assigns them lower quality values than unbiased subgroups.
To get a deeper understanding of the correlation between biased and quality, quality distributions for the different subgroups are illustrated.
These allow validating shifts and separations between biased and unbiased subgroups.



\paragraph{Face recognition networks}
To get the face embedding for a given face image, the image has to be aligned, scaled, and cropped. 
Then, the preprocessed image is passed to a face recognition model to extract the embeddings.
In this work, we use two face recognition models, FaceNet \cite{DBLP:journals/corr/SchroffKP15} and ArcFace \cite{Deng_2019_CVPR}.
For FaceNet, the preprocessing is done as described in \cite{Kazemi2014OneMF}.
To extract the embeddings, a pretrained model\footnote{\url{https://github.com/davidsandberg/facenet}} was used.
For ArcFace, the image preprocessing was done as described in \cite{DBLP:journals/corr/abs-1812-01936} and a pretrained model\footnote{\url{https://github.com/deepinsight/insightface}}  is used, which is provided by the authors of ArcFace.
Both models were trained on the MS1M database \cite{DBLP:journals/corr/GuoZHHG16}.
The output size is 128 for FaceNet and 512 for ArcFace.
The identity verification is done by comparing two embeddings using cosine-similarity.


\paragraph{Investigations}
This work aims at investigating the correlation between face recognition bias and face quality estimation.
This is done using two popular face embeddings, FaceNet \cite{DBLP:journals/corr/SchroffKP15} and ArcFace \cite{Deng_2019_CVPR}.
Since the face quality assessment performance strongly influences the interpretation of the correlation analysis, the quality estimation performance is analysed in the first step.
The second step aims at demonstrating that there is bias in the utilized face recognition systems.
Therefore, the face verification performance of these systems is analysed based on poses, ethnicities, and age classes.
After the bias between these classes is identified, the correlation between the face quality assessment and the face recognition bias is investigated in the third step.
Moreover, the separability in the quality space of the biased and unbiased classes is analysed.

%
%

\section{Results}
\label{sec:Results}

\subsection{Face quality assessment performance}
\label{chap:FQA-performance}

\begin{figure*}[h]
\centering  
\subfloat[ColorFeret - FaceNet \label{fig:FQA_ColorFeret_FaceNet}]{%
       \includegraphics[width=0.24\textwidth]{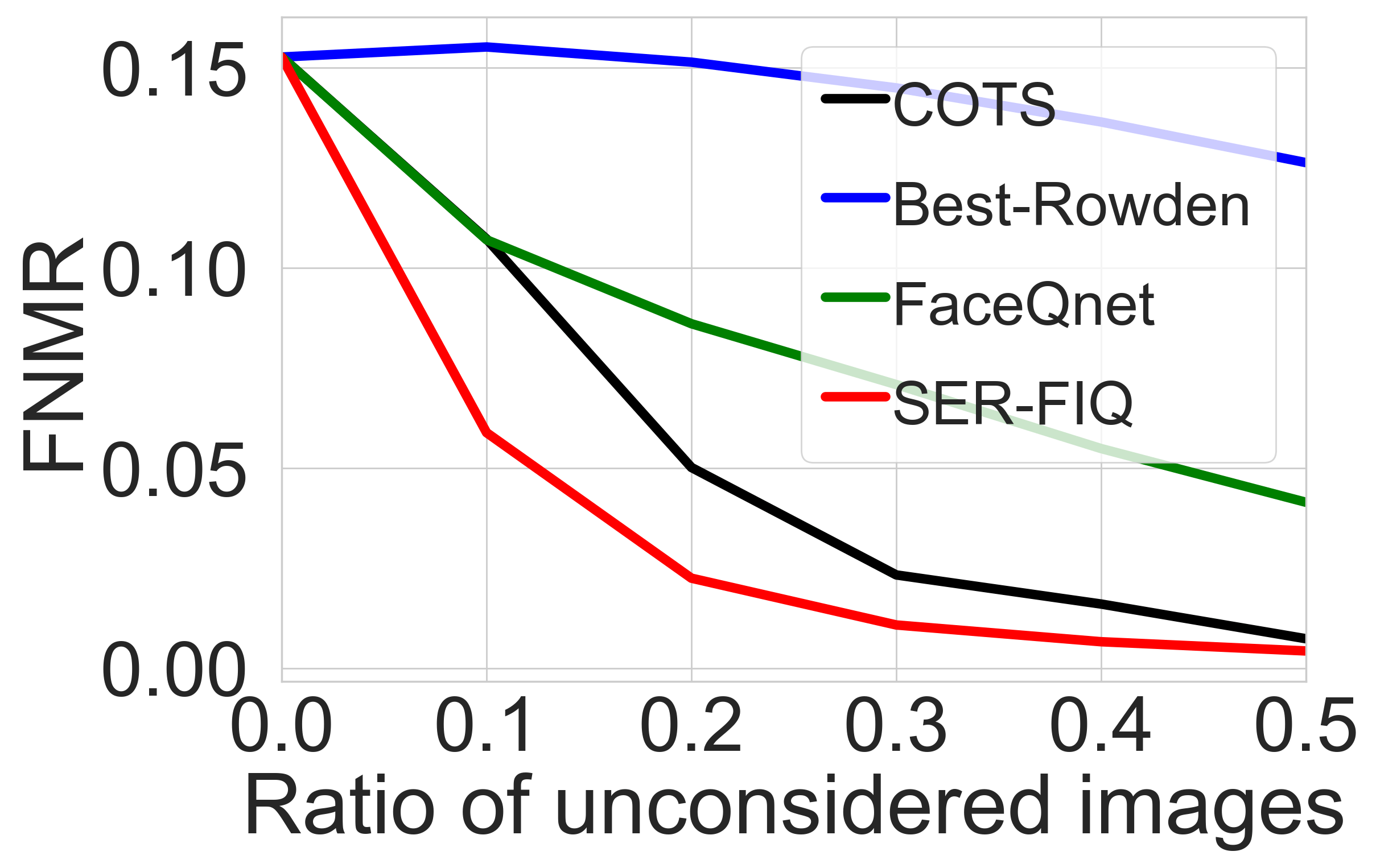}}
\subfloat[ColorFeret - ArcFace \label{fig:FQA_ColorFeret_ArcFace}]{%
       \includegraphics[width=0.24\textwidth]{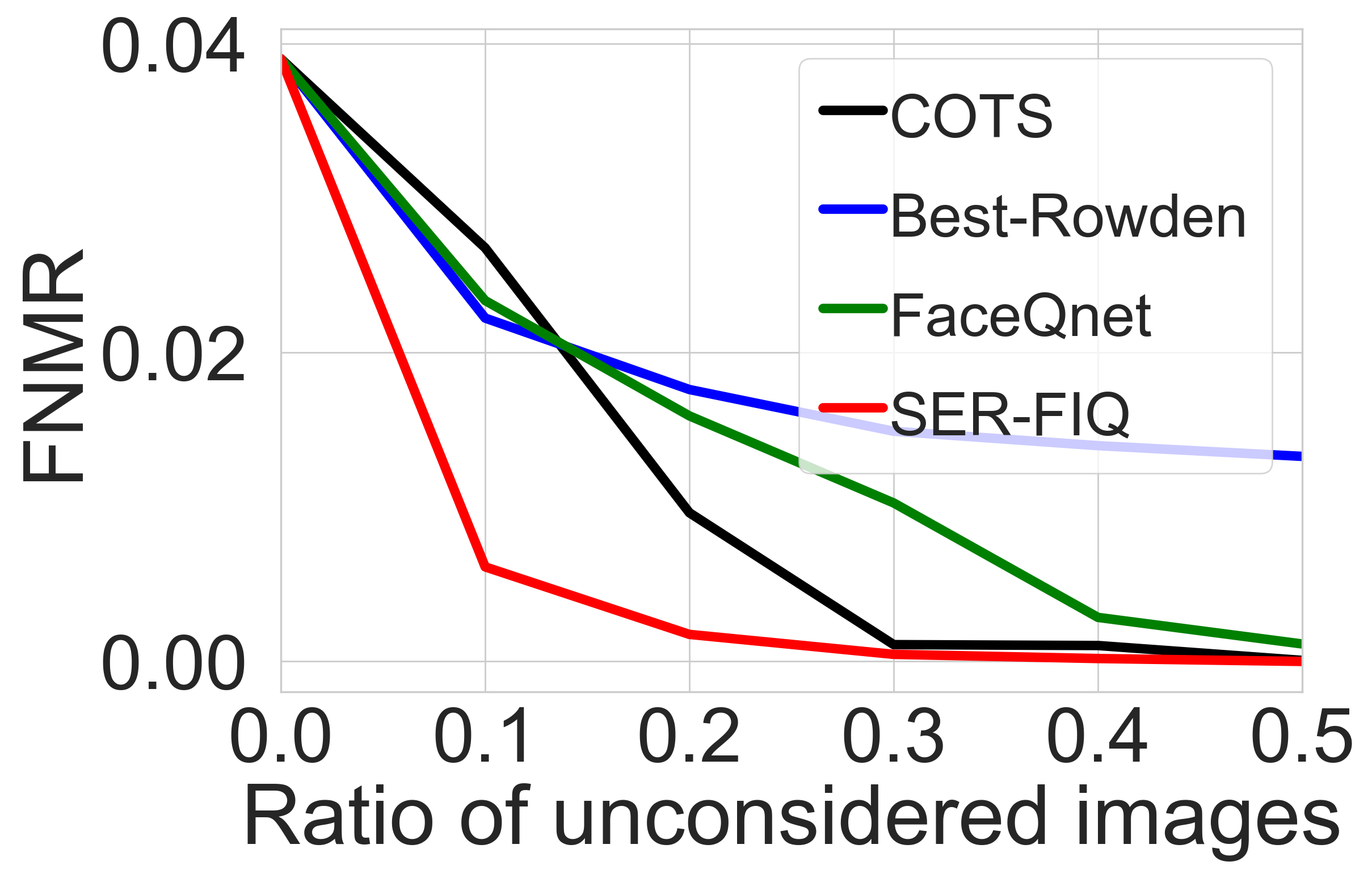}}
\subfloat[Adience - FaceNet \label{fig:FQA_Adience_FaceNet}]{%
       \includegraphics[width=0.24\textwidth]{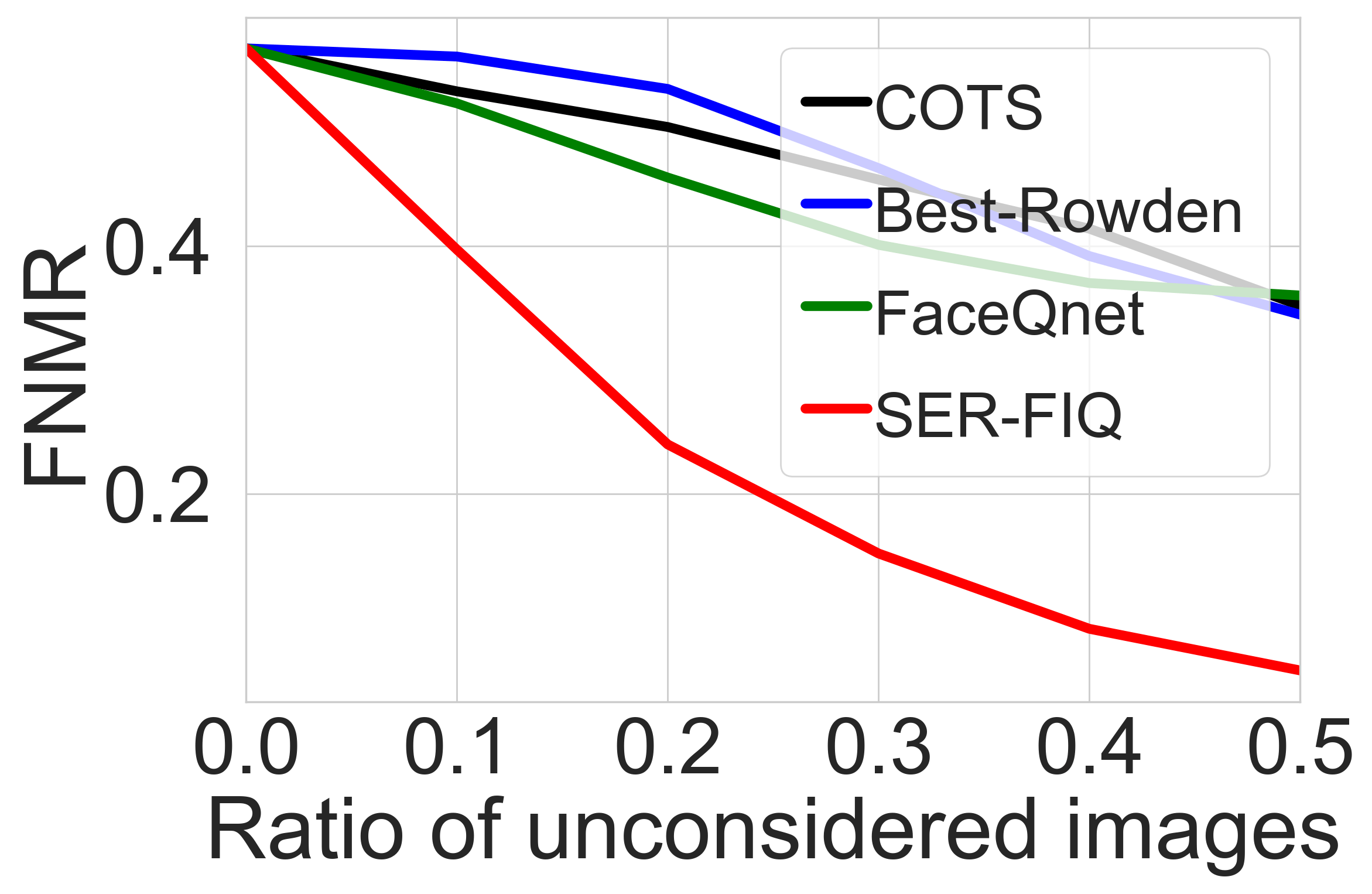}}
\subfloat[Adience - ArcFace \label{fig:FQA_Adience_ArcFace}]{%
       \includegraphics[width=0.24\textwidth]{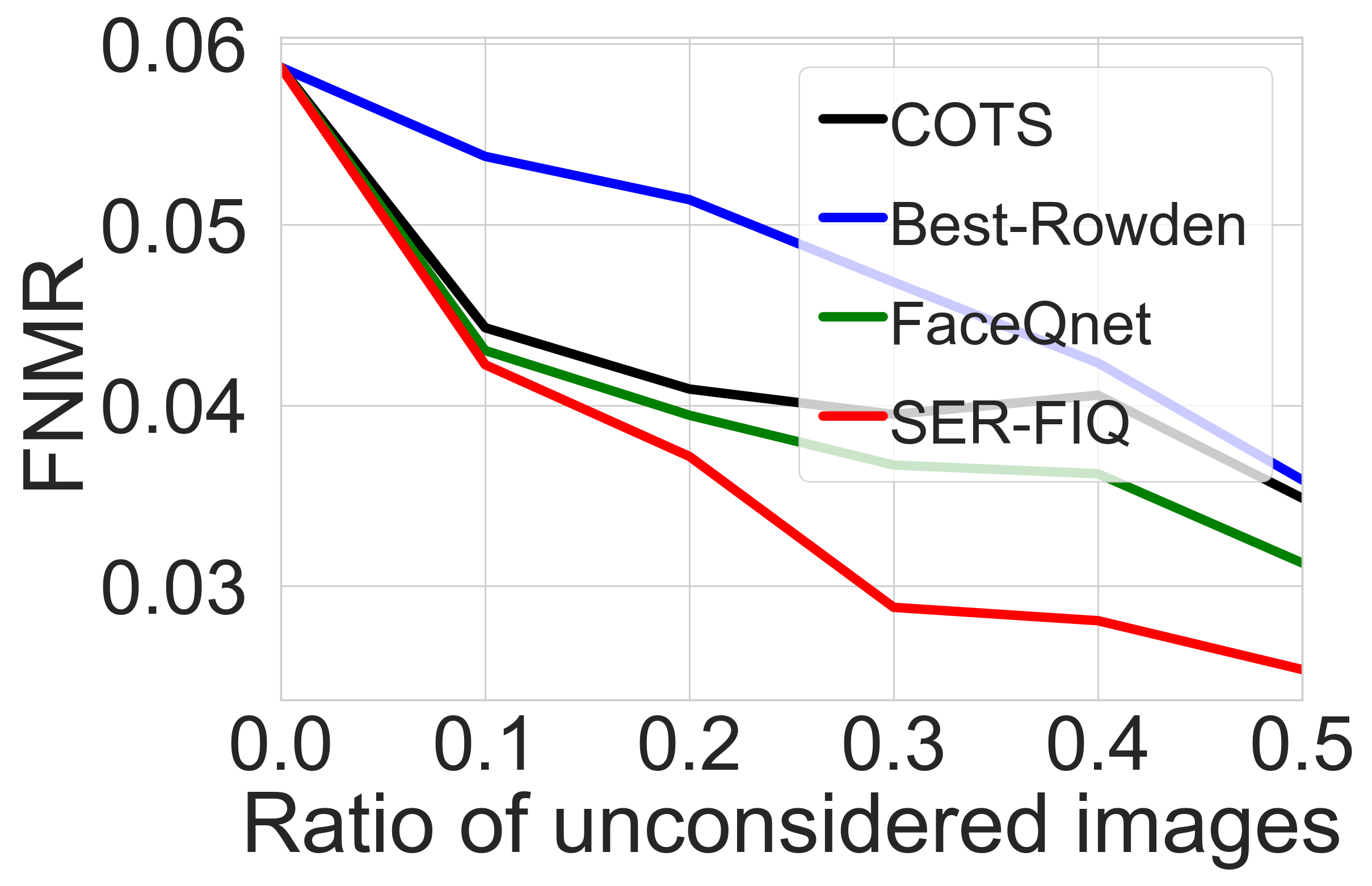}}    
\vspace{-2mm}
\caption{Face quality assessment performance on the ColorFeret and Adience datasets using two face embeddings, FaceNet and Arcface. The FNMR is reported at a FMR of 0.1\%. }
\vspace{-3mm}
\label{fig:FQA performance}
\end{figure*}

Figure \ref{fig:FQA performance} shows the face quality assessment performance for the four discussed solutions.
The performance is reported in terms of FNMR at FMR of 0.1\% as recommended by the European  Border Guard Agency Frontex \cite{FrontexBestPractice}.
It can be seen that COTS shows a better quality estimation performance under constrained scenarios (Figure \ref{fig:FQA_ColorFeret_FaceNet} and \ref{fig:FQA_ColorFeret_ArcFace}).
The approach of Best-Rowden shows a better quality prediction performance on ArcFace embeddings than on FaceNet.
This might be because Best-Rowden was trained on a frontal face database and ArcFace is more robust to these variations.
FaceQnet uses the same kind of training labels than Best-Rowden, but trained a deep learning model to make more advanced predictions.
This approach shows a solid performance in all cases.
Similar to the results from \cite{CVPR_Terhoerst}, SER-FIQ shows the best performance in all scenarios.
This is probably because this method exploits the decision patterns of the deployed model is therefore able to estimate how robust the model is about the input.

\subsection{Identifying biases in pose, ethnicity, and age}

\begin{table}[h]
\footnotesize
\renewcommand{\arraystretch}{1.1}
\setlength{\tabcolsep}{2.8pt}
\centering
\caption{Face verification performance within certain subgroups. The FNMR is evaluated at two FMR thresholds for two face recognition models. In each category, at least one subgroup shows a significantly higher error rate indicating a strong bias in the face embeddings.}
\label{tab:PerformanceBias}
\begin{tabular}{lllrrrr}
\Xhline{2\arrayrulewidth}
 & & & \multicolumn{2}{c}{FaceNet} & \multicolumn{2}{c}{ArcFace}         \\
 \cmidrule{4-5} \cmidrule{6-7}
 &  & Classes & 0.1\%FMR & 1\%FMR & 0.1\%FMR & 1\%FMR \\
\hline
\multirow{10}{*}{\rotatebox{90}{ColorFeret}} & \multirow{5}{*}{\rotatebox{90}{Pose}} & Frontal & 0.40\% & 0.00\% & 0.00\% & 0.00\% \\
 &  & Half & 1.78\% & 0.15\% & 0.07\% & 0.04\% \\
 &  & Profile & 30.95\% & 10.14\% & 12.29\% & 7.55\% \\
 &  & Rotated & 0.07\% & 0.03\% & 1.39\% & 0.00\% \\
\cmidrule{2-7}
 & \multirow{4}{*}{\rotatebox{90}{Ethnicity}} & White & 10.79\% & 3.41\% & 2.55\% & 1.80\% \\
 &  & Asian & 33.90\% & 12.06\% & 6.63\% & 4.19\% \\
 &  & Black & 31.34\% & 16.54\% & 6.41\% & 3.66\% \\
 &  & Others & 12.15\% & 6.29\% & 3.53\% & 2.08\% \\
 \cmidrule{2-7}
 & & All & 16.22\% & 3.92\% & 4.15\% & 2.98\% \\
 \Xhline{2\arrayrulewidth}
\multirow{9}{*}{\rotatebox{90}{Adience}} & \multirow{8}{*}{\rotatebox{90}{Age}} & {[}0,2{]} & 80.02\% & 59.88\% & 18.81\% & 9.73\% \\
 &  & {[}4,6{]} & 63.95\% & 36.80\% & 13.19\% & 6.46\% \\
 &  & {[}8,12{]} & 37.16\% & 17.27\% & 9.92\% & 4.79\% \\
 &  & {[}15,20{]} & 89.78\% & 52.51\% & 10.30\% & 6.15\% \\
 &  & {[}25,32{]} & 28.37\% & 4.58\% & 5.31\% & 4.81\% \\
 &  & {[}38,43{]} & 16.48\% & 4.07\% & 2.68\% & 2.07\% \\
 &  & {[}48,53{]} & 20.94\% & 5.85\% & 1.92\% & 1.39\% \\
 &  & {[}60,100{]} & 11.32\% & 2.97\% & 1.67\% & 0.66\% \\
 \cmidrule{2-7}
 & & All & 55.99\% & 16.28\% & 5.99\% & 3.24\%\\
 \Xhline{2\arrayrulewidth}
\end{tabular}
\end{table}

In order to identify biased classes in the two utilized face embeddings, Table \ref{tab:PerformanceBias} shows the face verification performance at two decision thresholds for FaceNet and ArcFace embeddings. 
The performance is evaluated over four different head poses, four ethnicities, and eight age classes.
In the case of poses, all poses show very low error rates, with the exception of the profile view.
Here, the error rates are more than 10 times higher than the next highest class.
This shows that there is a strong bias towards profile face images.
In the case of ethnicities, face images of white individuals show the smallest error rate, followed by the class others.
For the ethnicities asian and black, the error rates are strongly increased and thus, indicate a strong ethnic bias.
This might come from a training process that mainly involved face images of white individuals.
In the cases of the age classes, there are higher error rates among young people (below 7 years) compared to older individuals.
This bias might come from the lack of appropriate training material as well as the fact that faces at this age are not yet fully developed.

Over all three attributes pose, ethnicity, and age, ArcFace shows significantly lower error rates than FaceNet. 
However, for both face embeddings, it is demonstrated that there exists high biases towards certain classes.

%
%
%

\subsection{The correlation study - bias versus quality}

\begin{figure*}[h]
\captionsetup[subfloat]{farskip=5pt,captionskip=1pt}
\centering
\subfloat[Pose - COTS \label{fig:StackedCharts_ColorFeret_Pose_FaceNet_COTS}]{%
       \includegraphics[width=0.24\textwidth]{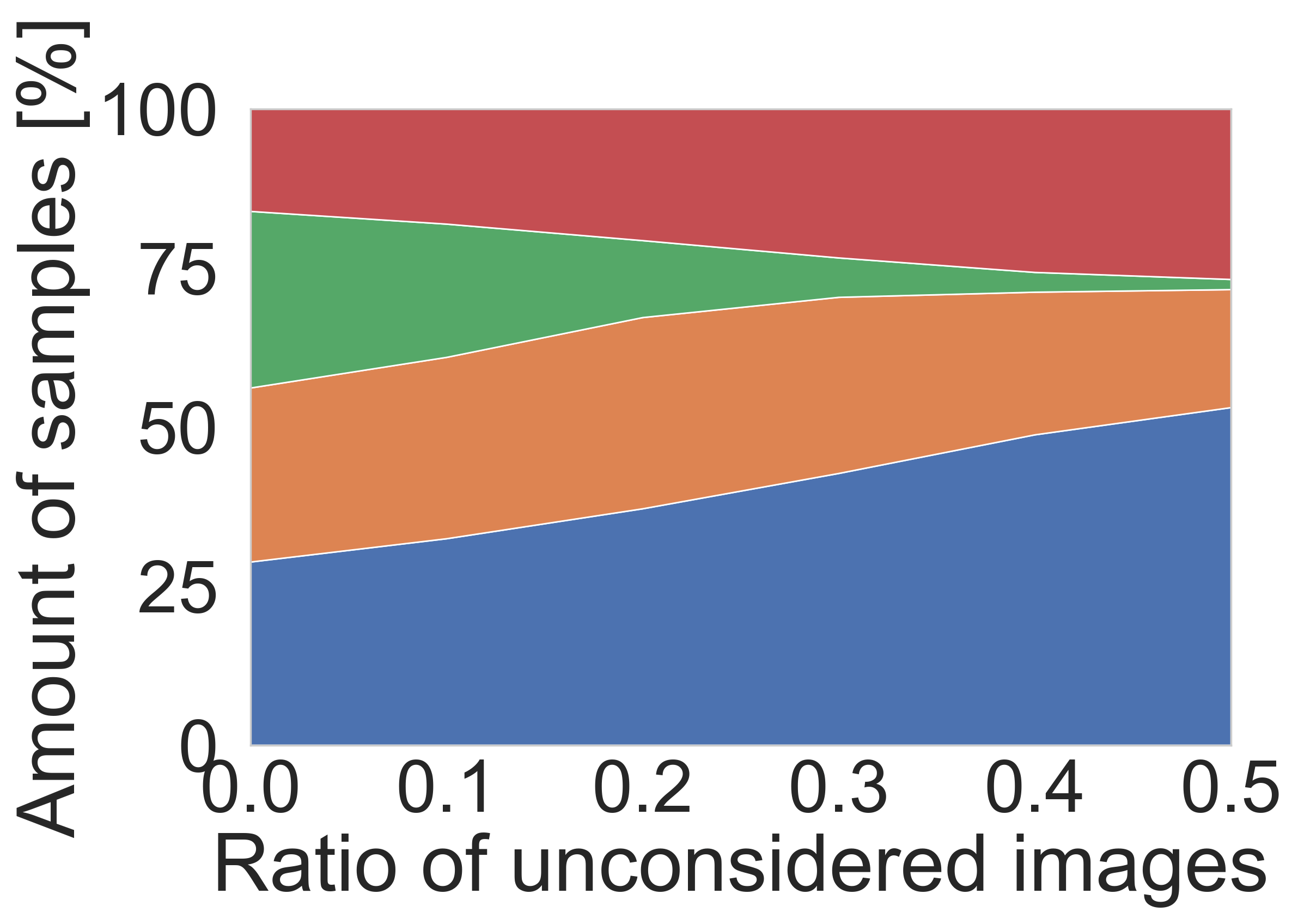}}
\subfloat[Pose - Best-Rowden \label{fig:StackedCharts_ColorFeret_Pose_FaceNet_BestRowden}]{%
       \includegraphics[width=0.24\textwidth]{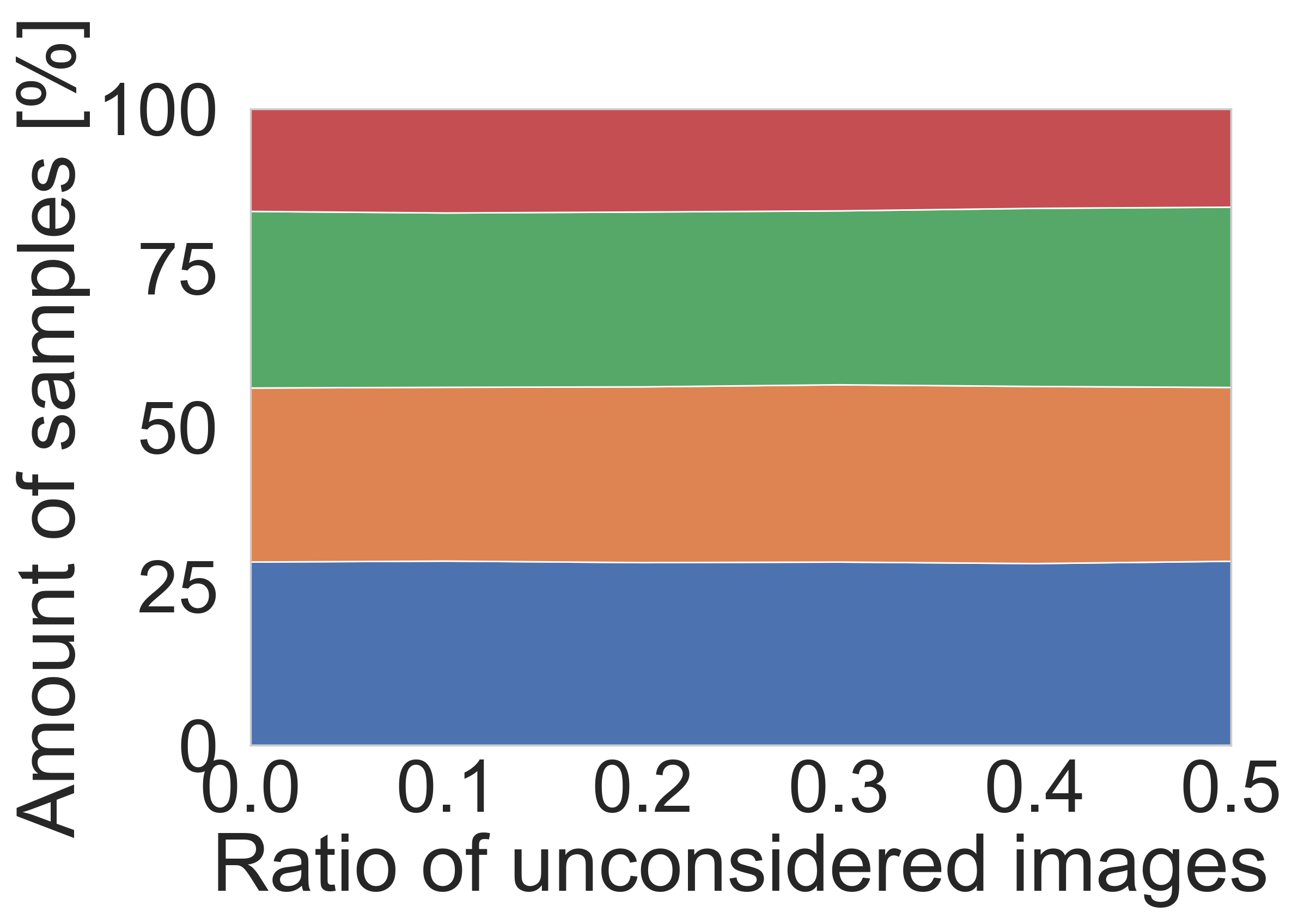}}
\subfloat[Pose - FaceQnet \label{fig:StackedCharts_ColorFeret_Pose_FaceNet_FaceQnet}]{%
       \includegraphics[width=0.24\textwidth]{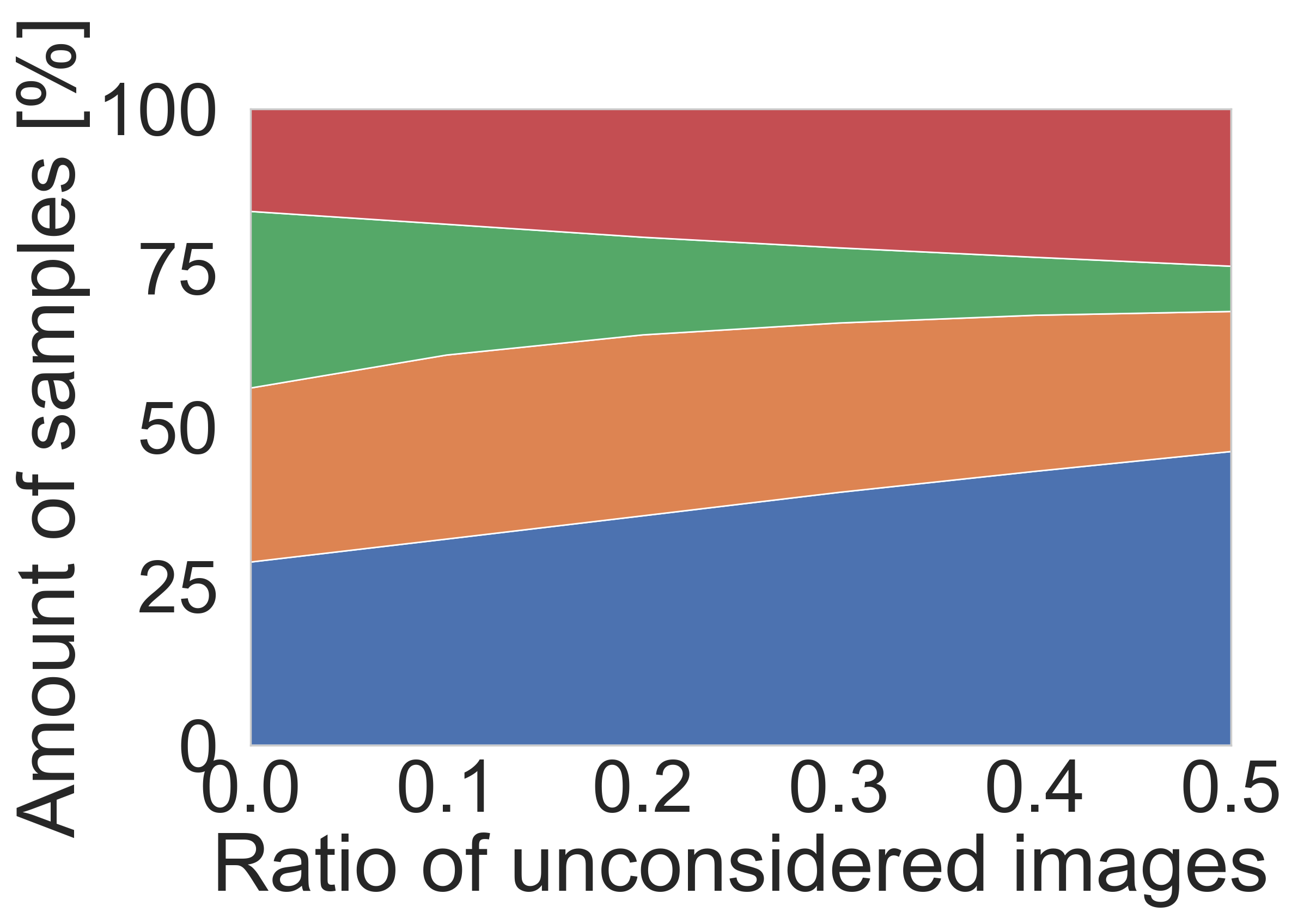}}
\subfloat[Pose - SER-FIQ   \label{fig:StackedCharts_ColorFeret_Pose_FaceNet_SER-FIQ}]{%
       \includegraphics[width=0.24\textwidth]{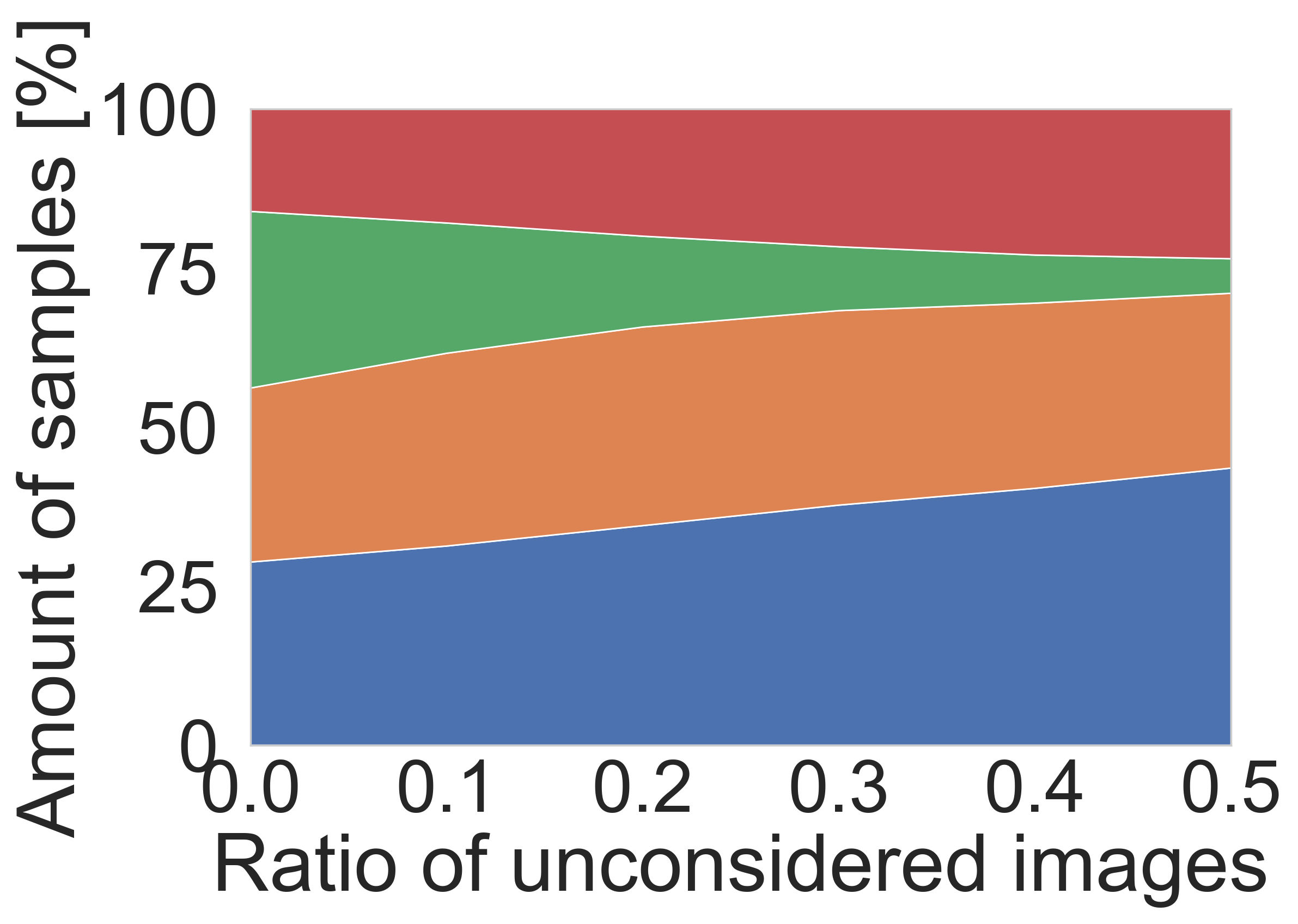}} 
       \adjustbox{raise=6mm}{\includegraphics[height=19mm]{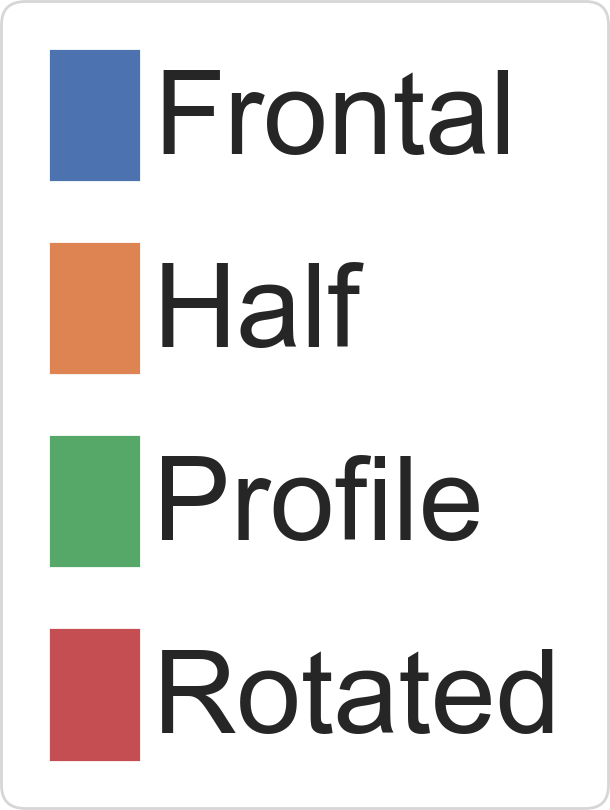}}           
\vspace{-3mm}
             
\subfloat[Ethnicity - COTS \label{fig:StackedCharts_ColorFeret_Ethnicity_FaceNet_COTS}]{%
       \includegraphics[width=0.24\textwidth]{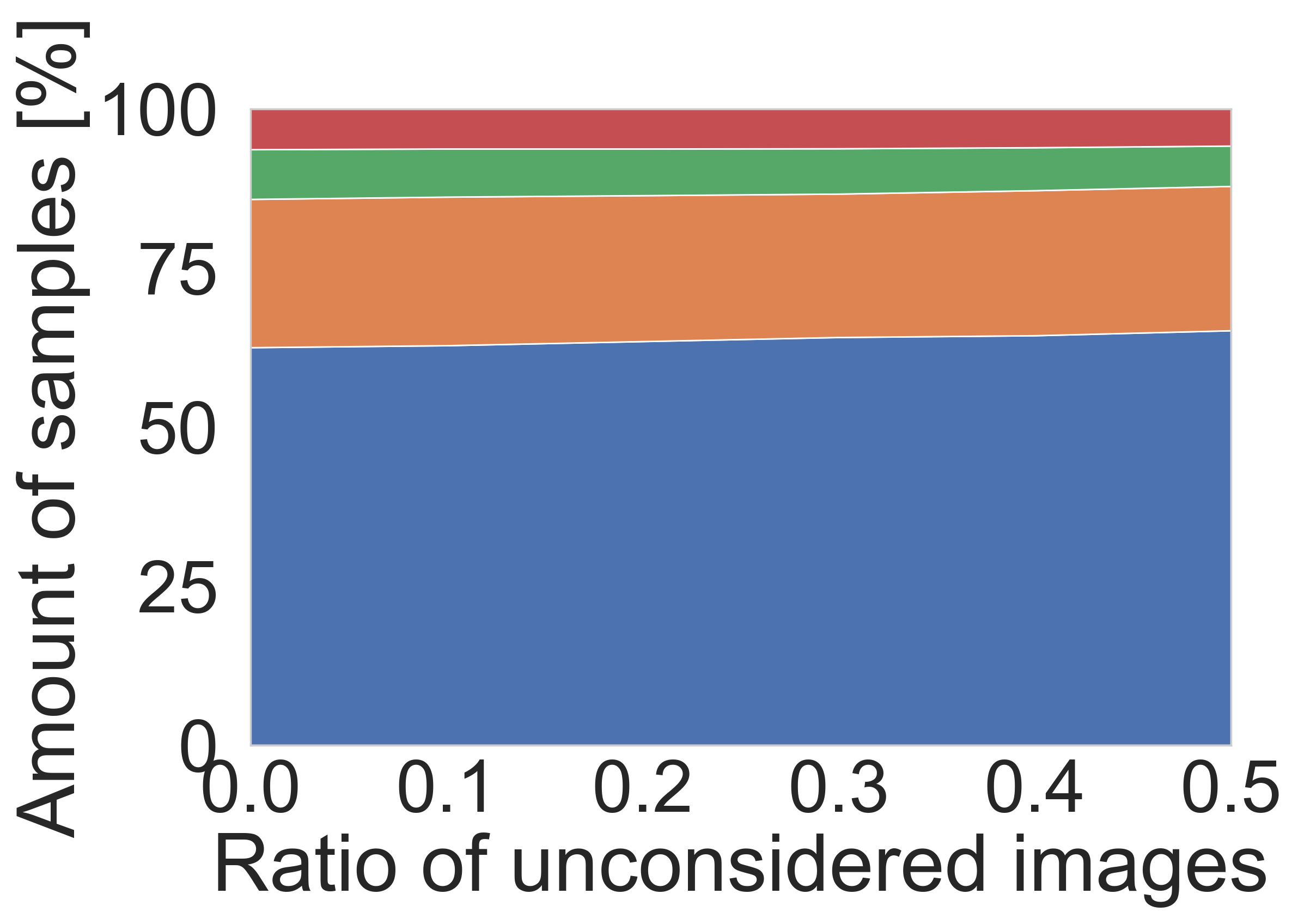}}
\subfloat[Ethnicity - Best-Rowden \label{fig:StackedCharts_ColorFeret_Ethnicity_FaceNet_BestRowden}]{%
       \includegraphics[width=0.24\textwidth]{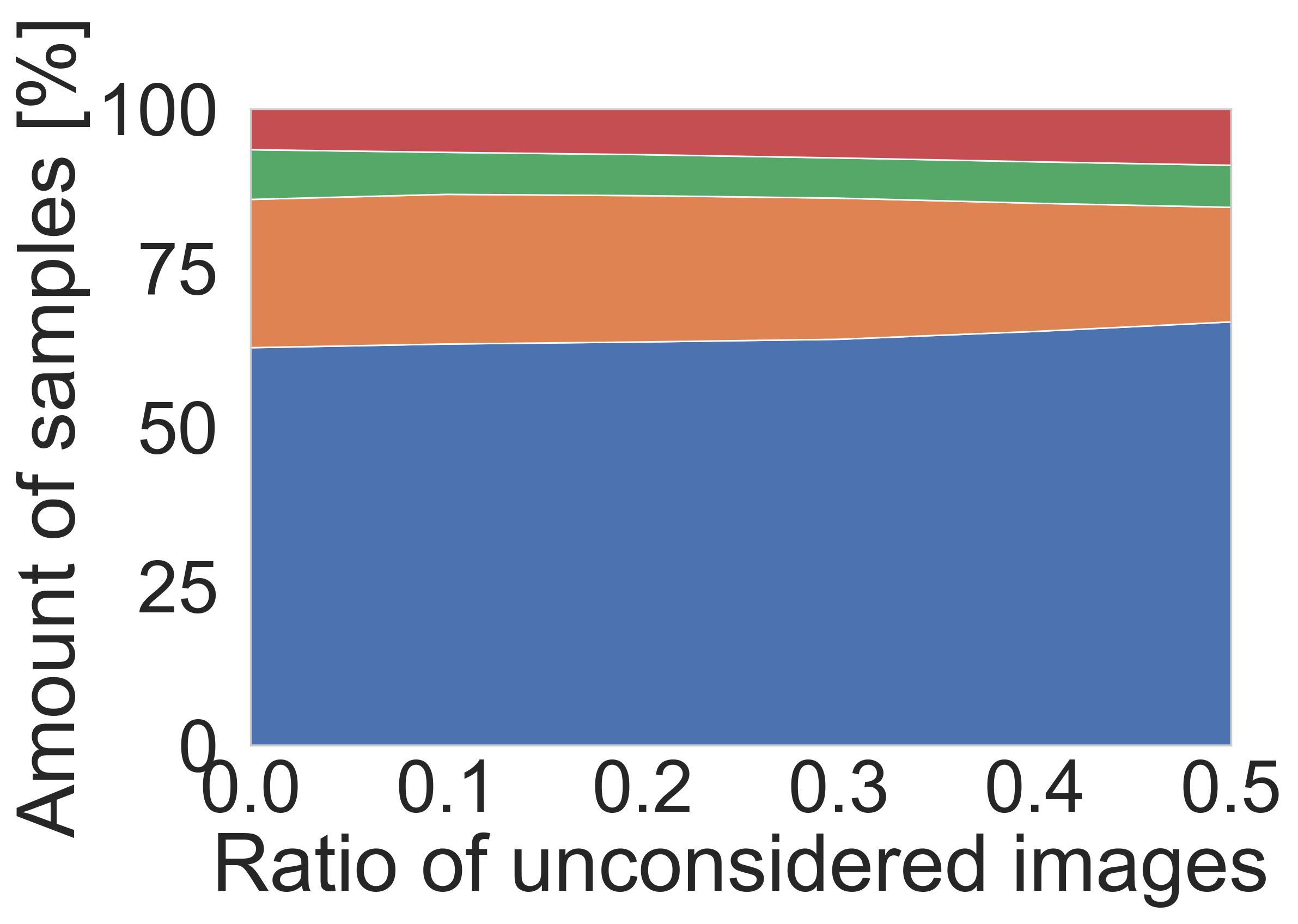}}
\subfloat[Ethnicity - FaceQnet \label{fig:StackedCharts_ColorFeret_Ethnicity_FaceNet_FaceQnet}]{%
       \includegraphics[width=0.24\textwidth]{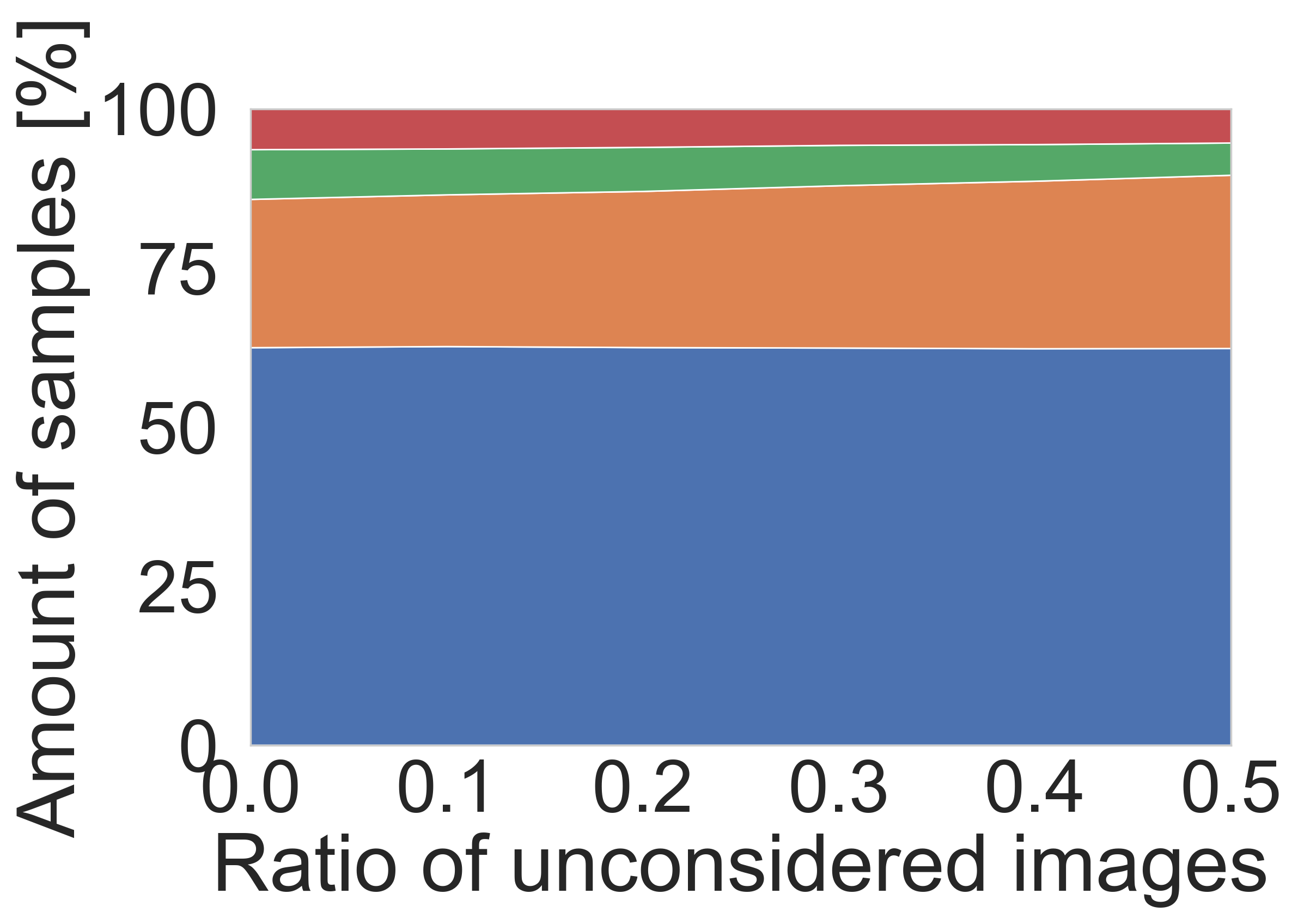}}   
\subfloat[Ethnicity - SER-FIQ   \label{fig:StackedCharts_ColorFeret_Ethnicity_FaceNet_SER-FIQ}]{%
       \includegraphics[width=0.24\textwidth]{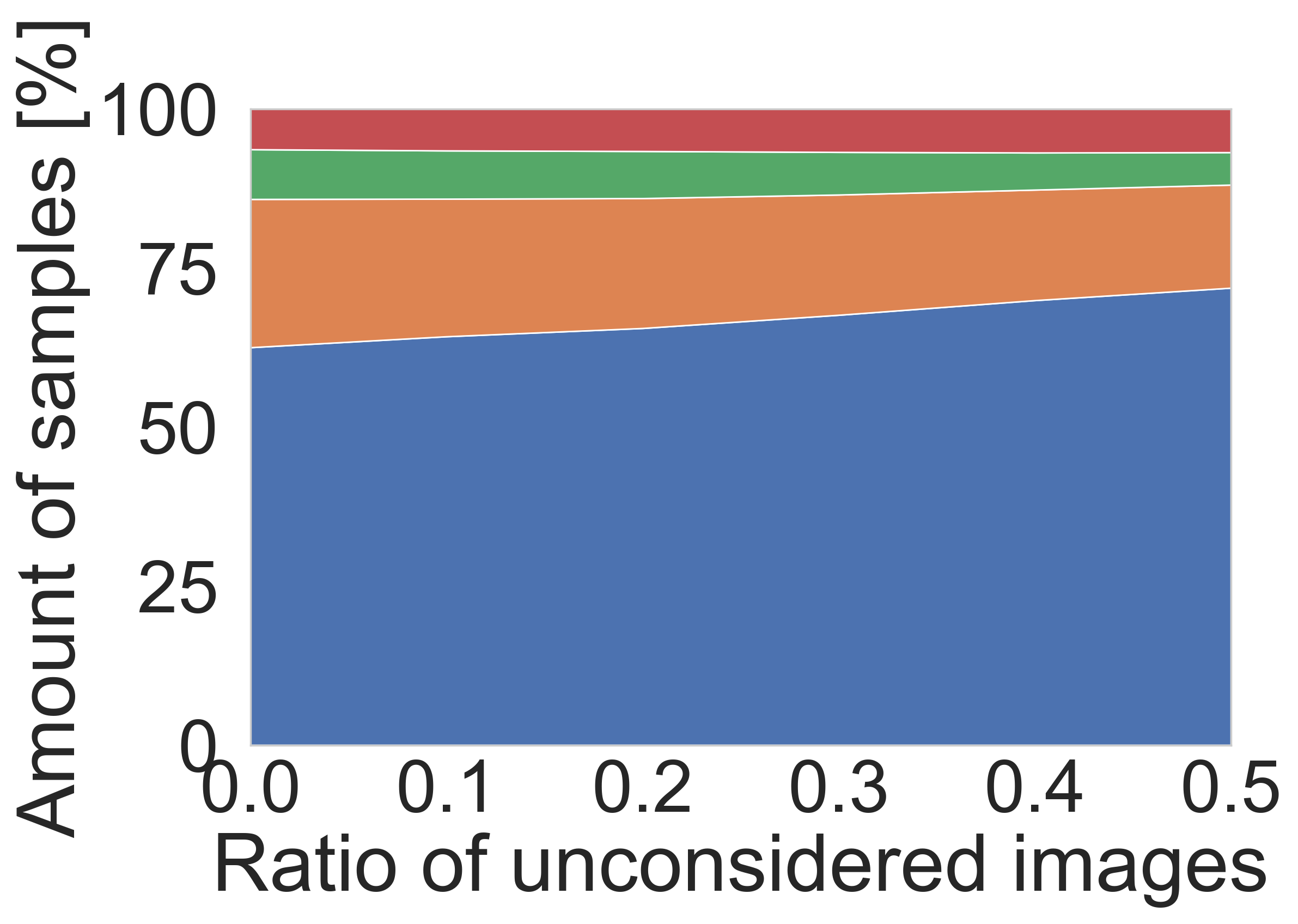}}
       \adjustbox{raise=6mm}{\includegraphics[height=19mm]{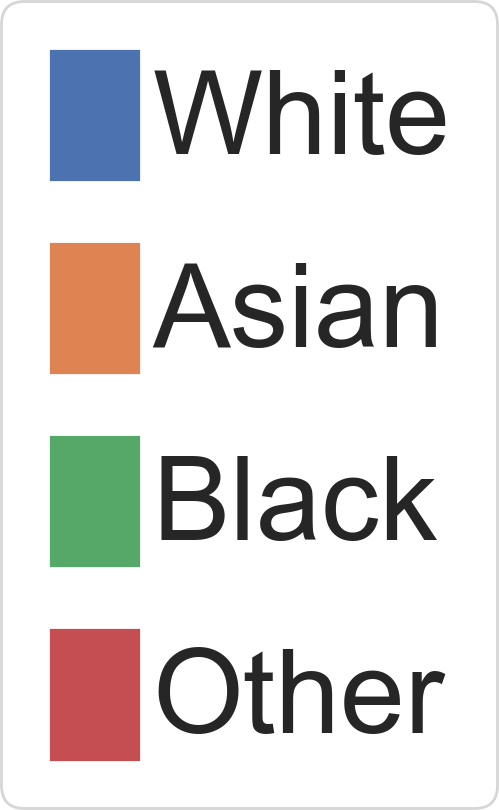}} 
\vspace{-3mm}
              
\subfloat[Age - COTS \label{fig:StackedCharts_Adience_Age_FaceNet_COTS}]{%
       \includegraphics[width=0.24\textwidth]{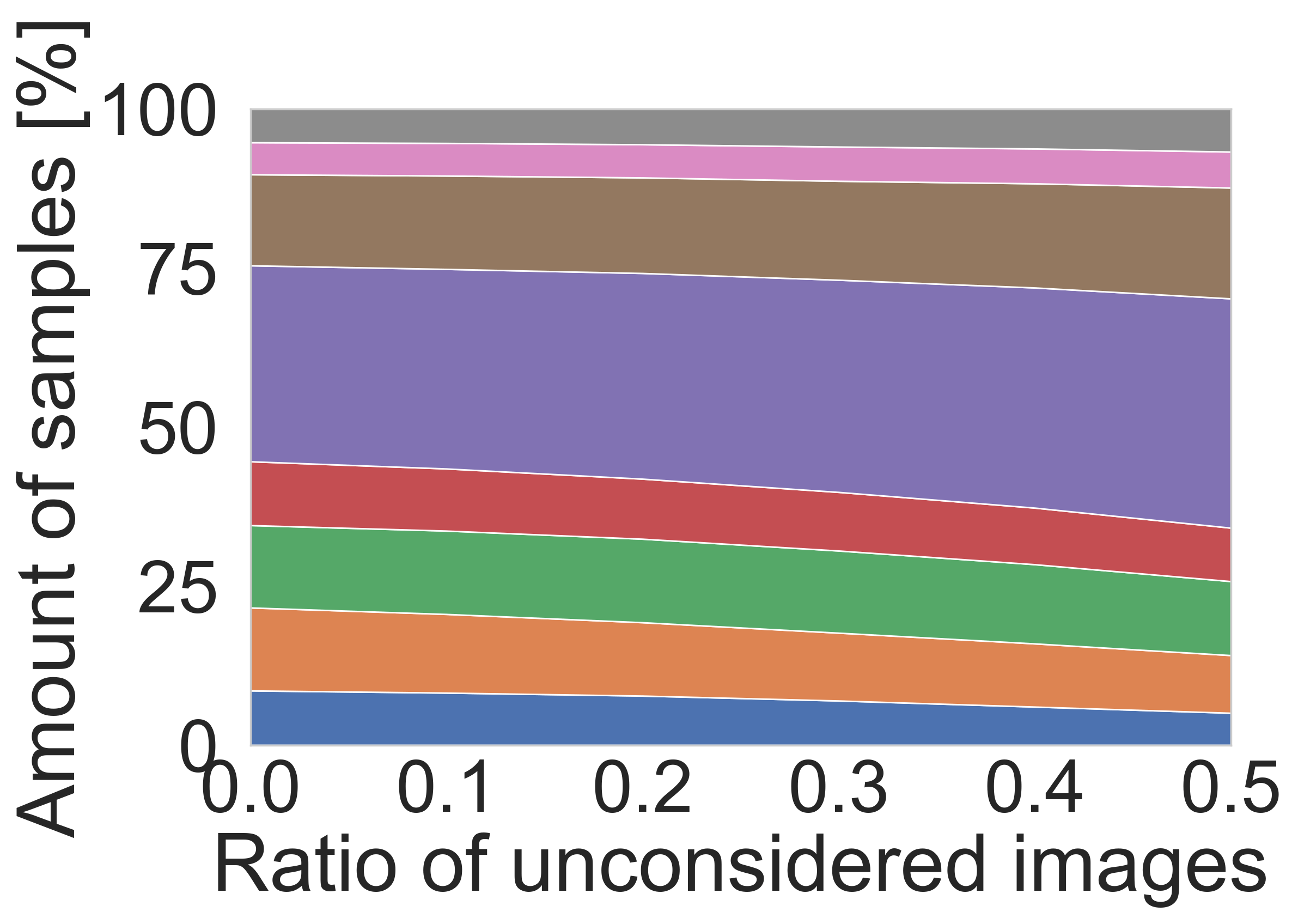}}
\subfloat[Age - Best-Rowden \label{fig:StackedCharts_Adience_Age_FaceNet_BestRowden}]{%
       \includegraphics[width=0.24\textwidth]{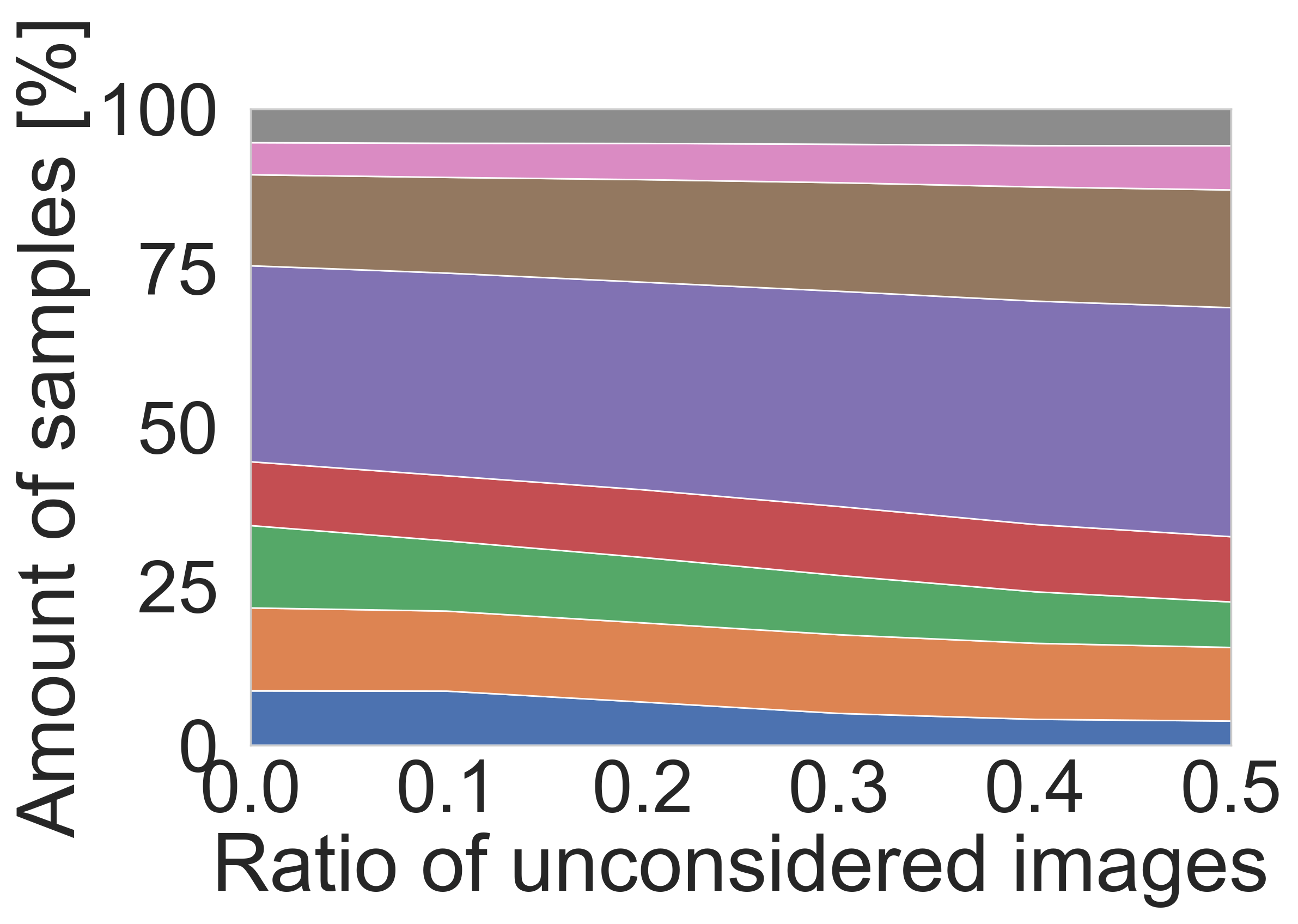}}
\subfloat[Age - FaceQnet \label{fig:StackedCharts_Adience_Age_FaceNet_FaceQnet}]{%
       \includegraphics[width=0.24\textwidth]{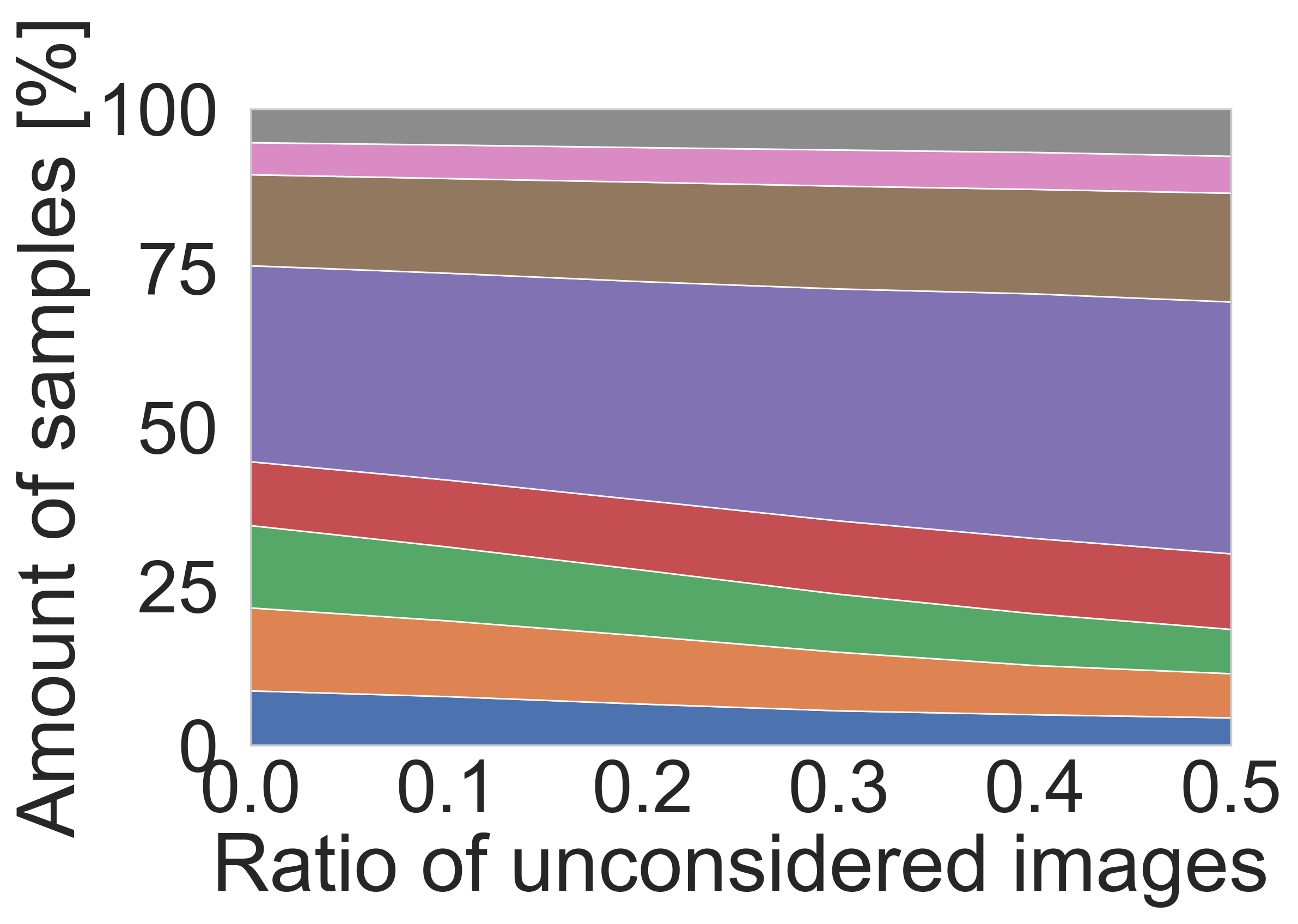}}
\subfloat[Age - SER-FIQ   \label{fig:StackedCharts_Adience_Age_FaceNet_SER-FIQ}]{%
       \includegraphics[width=0.24\textwidth]{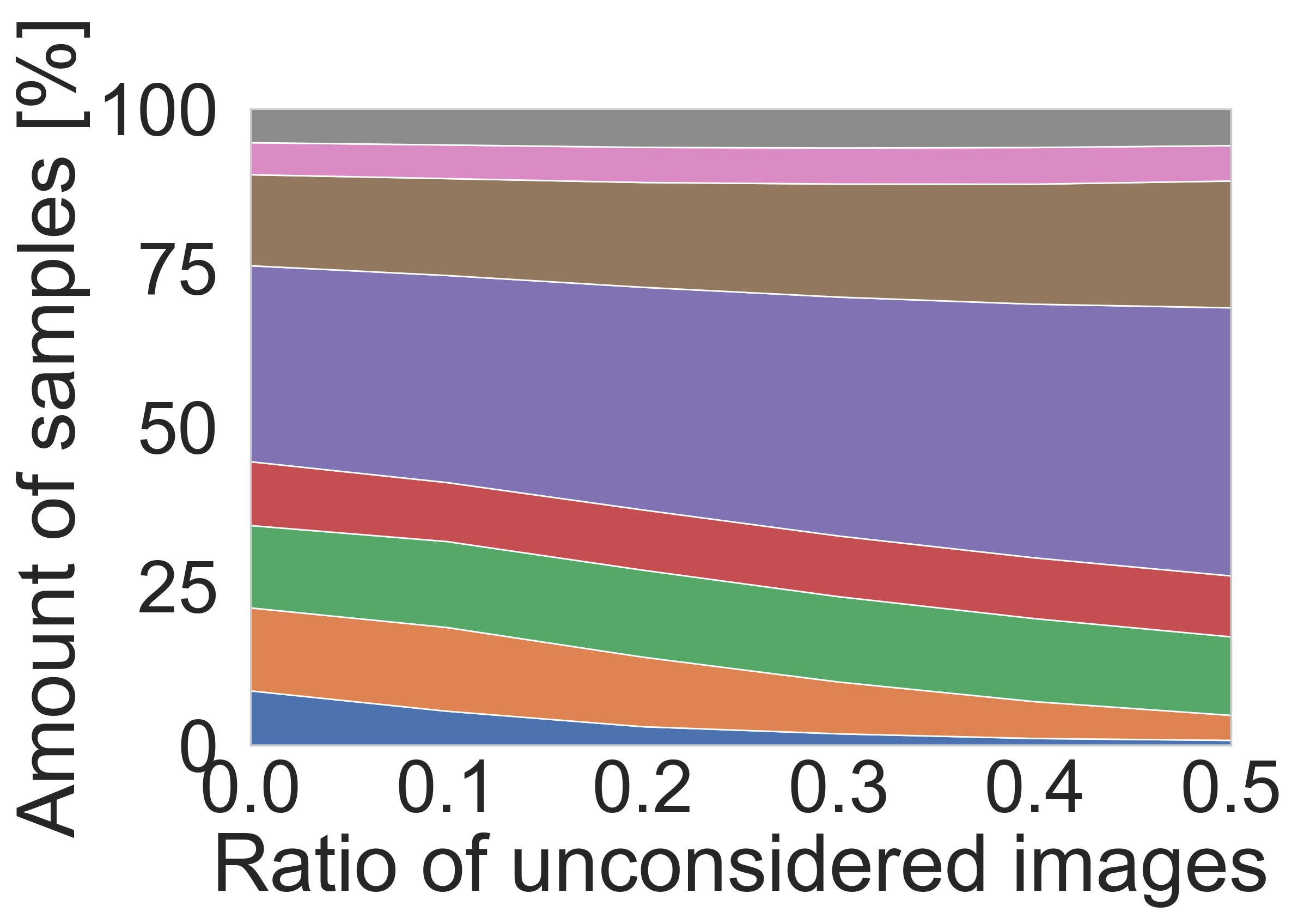}}  
       \adjustbox{raise=0mm}{\includegraphics[height=28mm]{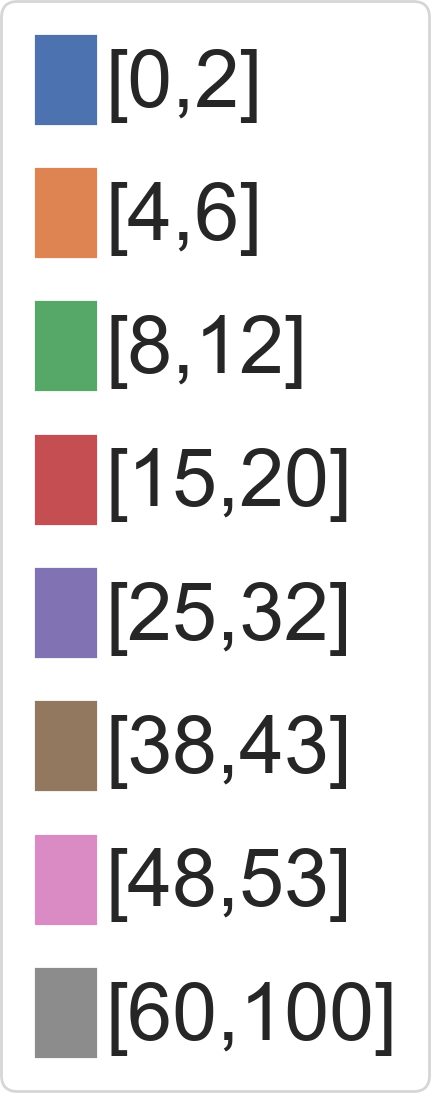}} 
\vspace{-2mm}   
\caption{Analysis of the proportion of subgroups for FaceNet embeddings. The pose (a-d), ethnicities (e-h), and age (i-l) proportions are shown when applying several quality thresholds.}
\label{fig:StackedCharts_FaceNet}
\vspace{-3mm} 
\end{figure*}

\begin{figure*}[h]
\captionsetup[subfloat]{farskip=5pt,captionskip=1pt}
\centering
\subfloat[Pose - COTS \label{fig:StackedCharts_ColorFeret_Pose_ArcFace_COTS}]{%
       \includegraphics[width=0.24\textwidth]{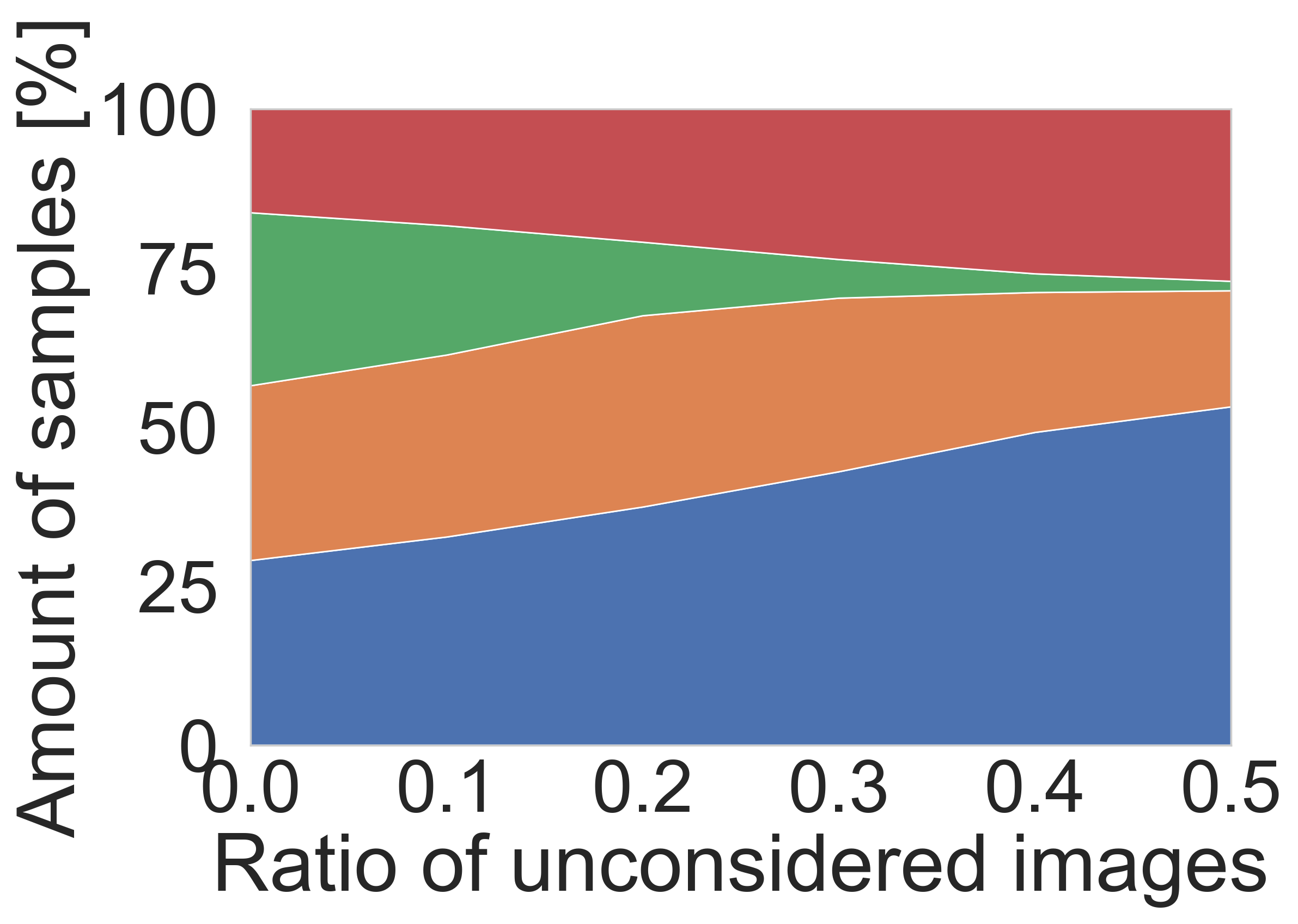}}   
\subfloat[Pose - Best-Rowden \label{fig:StackedCharts_ColorFeret_Pose_ArcFace_BestRowden}]{%
       \includegraphics[width=0.24\textwidth]{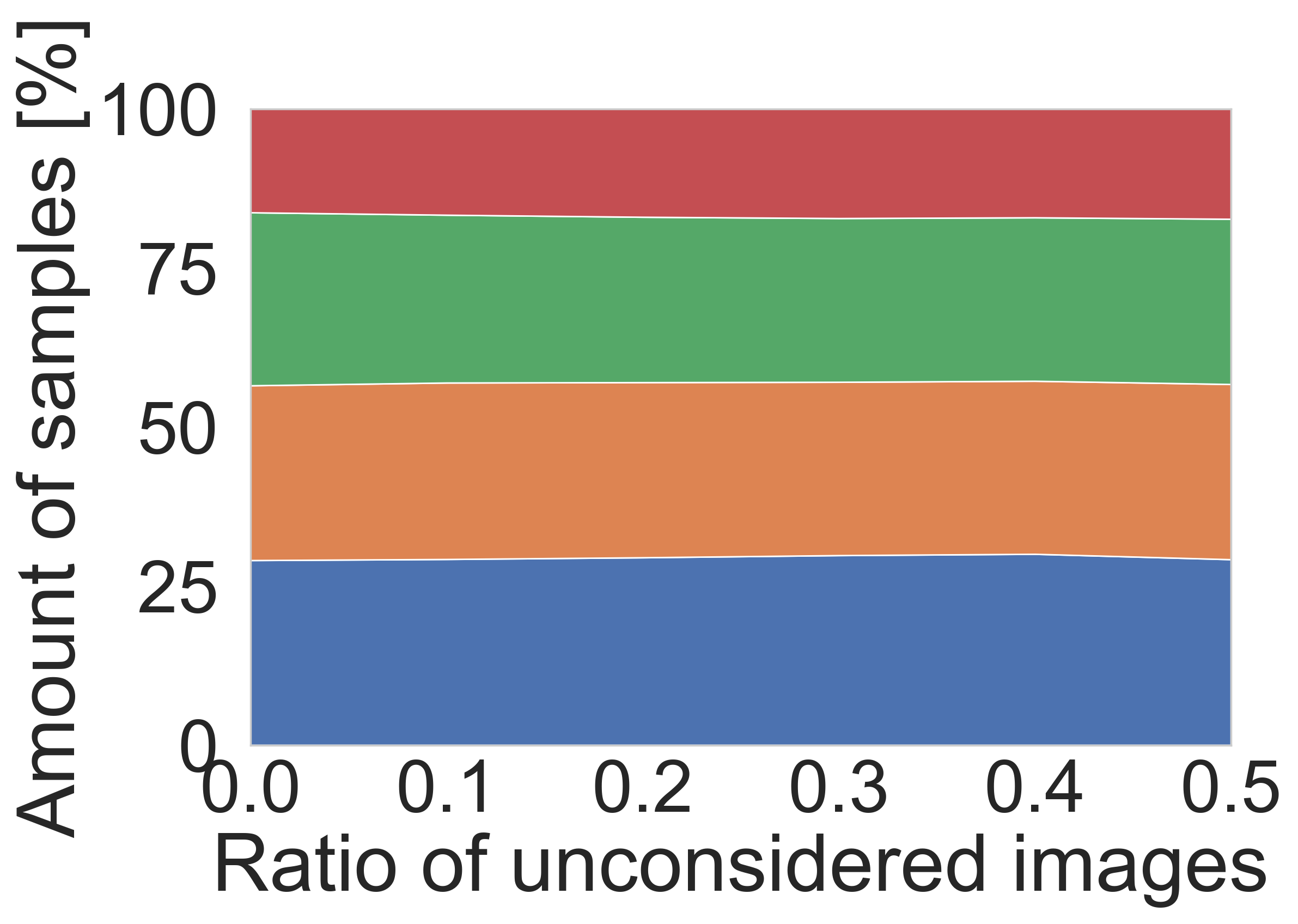}}
\subfloat[Pose - FaceQnet \label{fig:StackedCharts_ColorFeret_Pose_ArcFace_FaceQnet}]{%
       \includegraphics[width=0.24\textwidth]{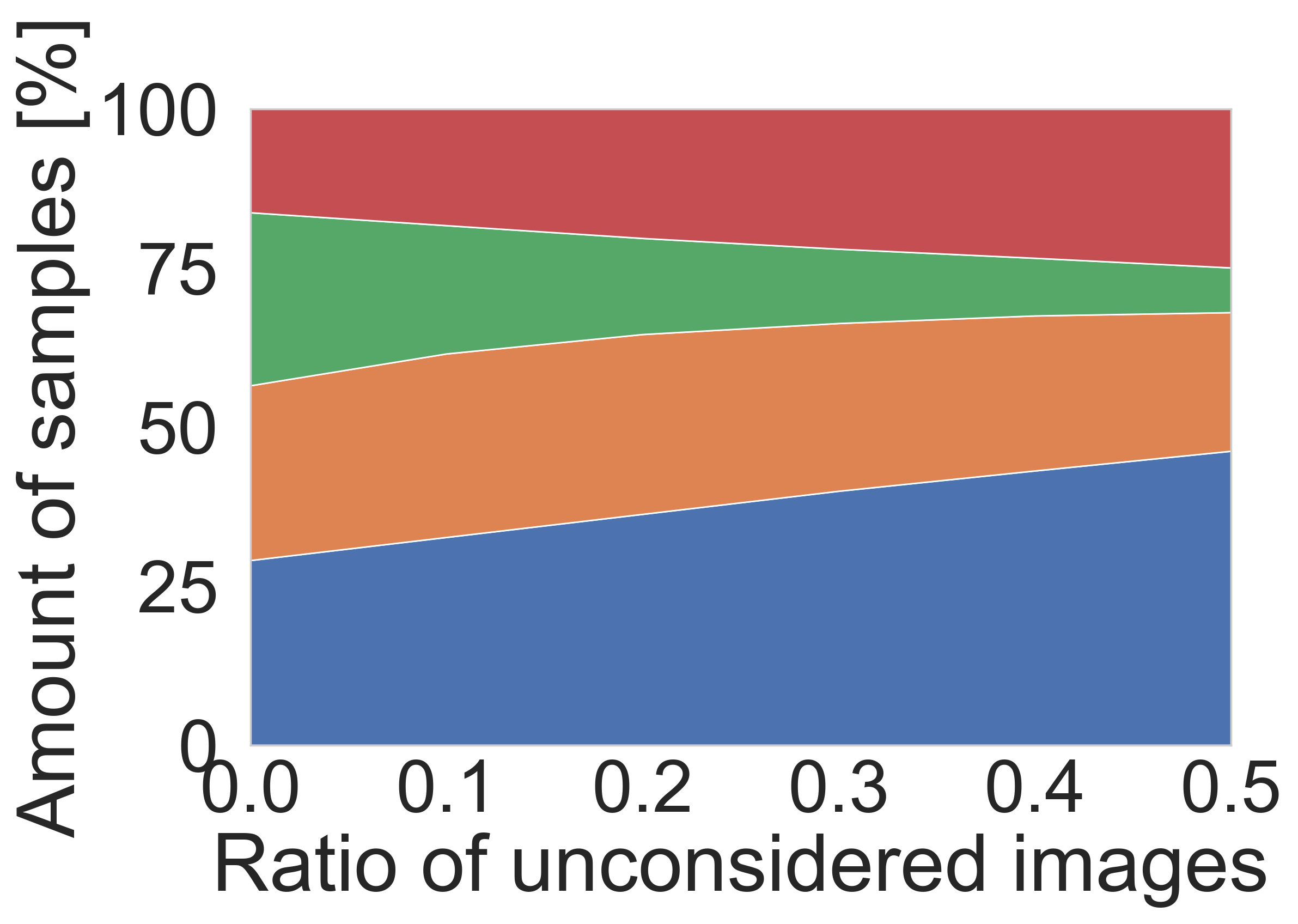}}
\subfloat[Pose - SER-FIQ   \label{fig:StackedCharts_ColorFeret_Pose_ArcFace_SER-FIQ}]{%
       \includegraphics[width=0.24\textwidth]{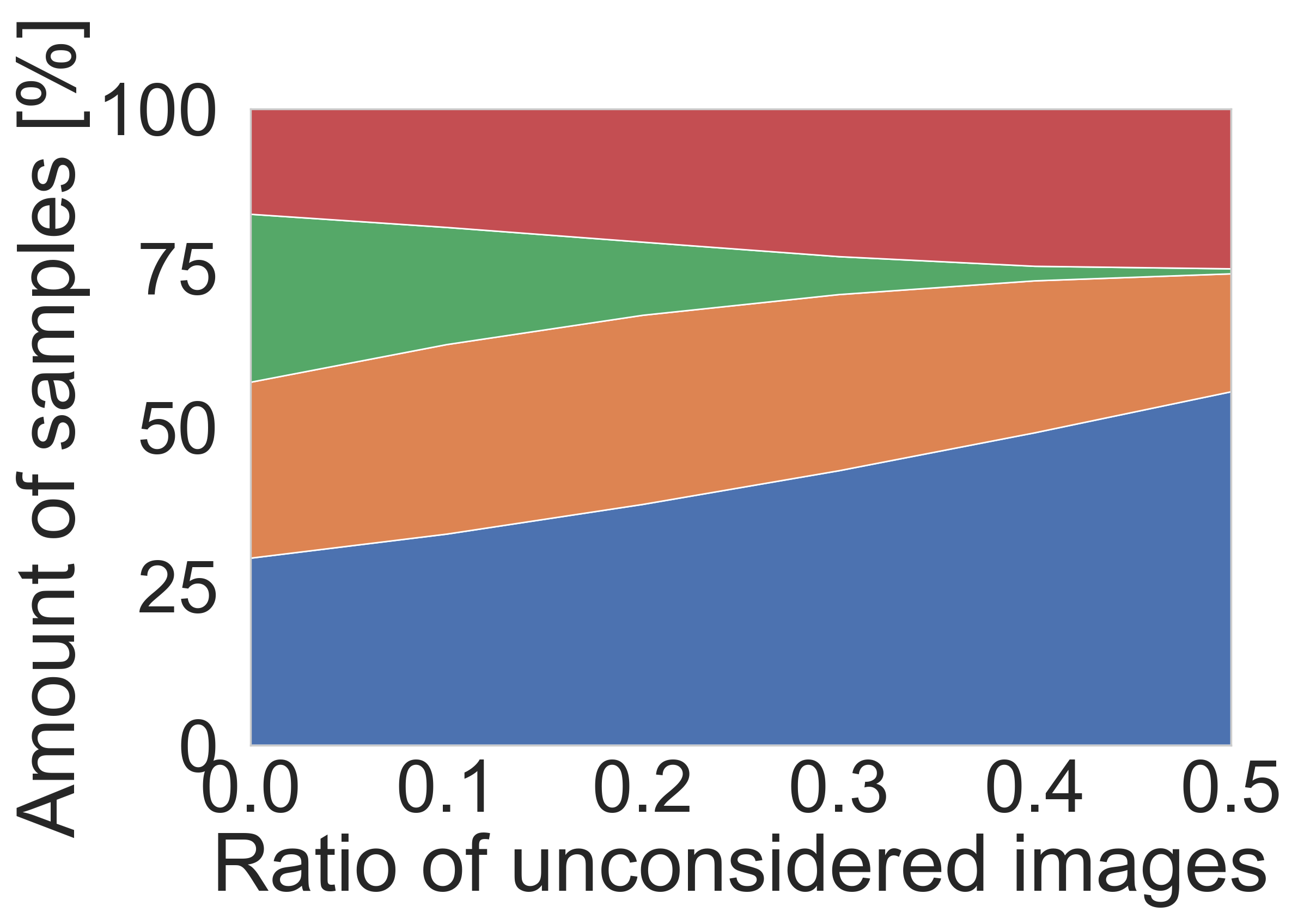}}
       \adjustbox{raise=6mm}{\includegraphics[height=19mm]{colorferet_pose.png}}    
\vspace{-3mm}
       
\subfloat[Ethnicity - COTS \label{fig:StackedCharts_ColorFeret_Ethnicity_ArcFace_COTS}]{%
       \includegraphics[width=0.24\textwidth]{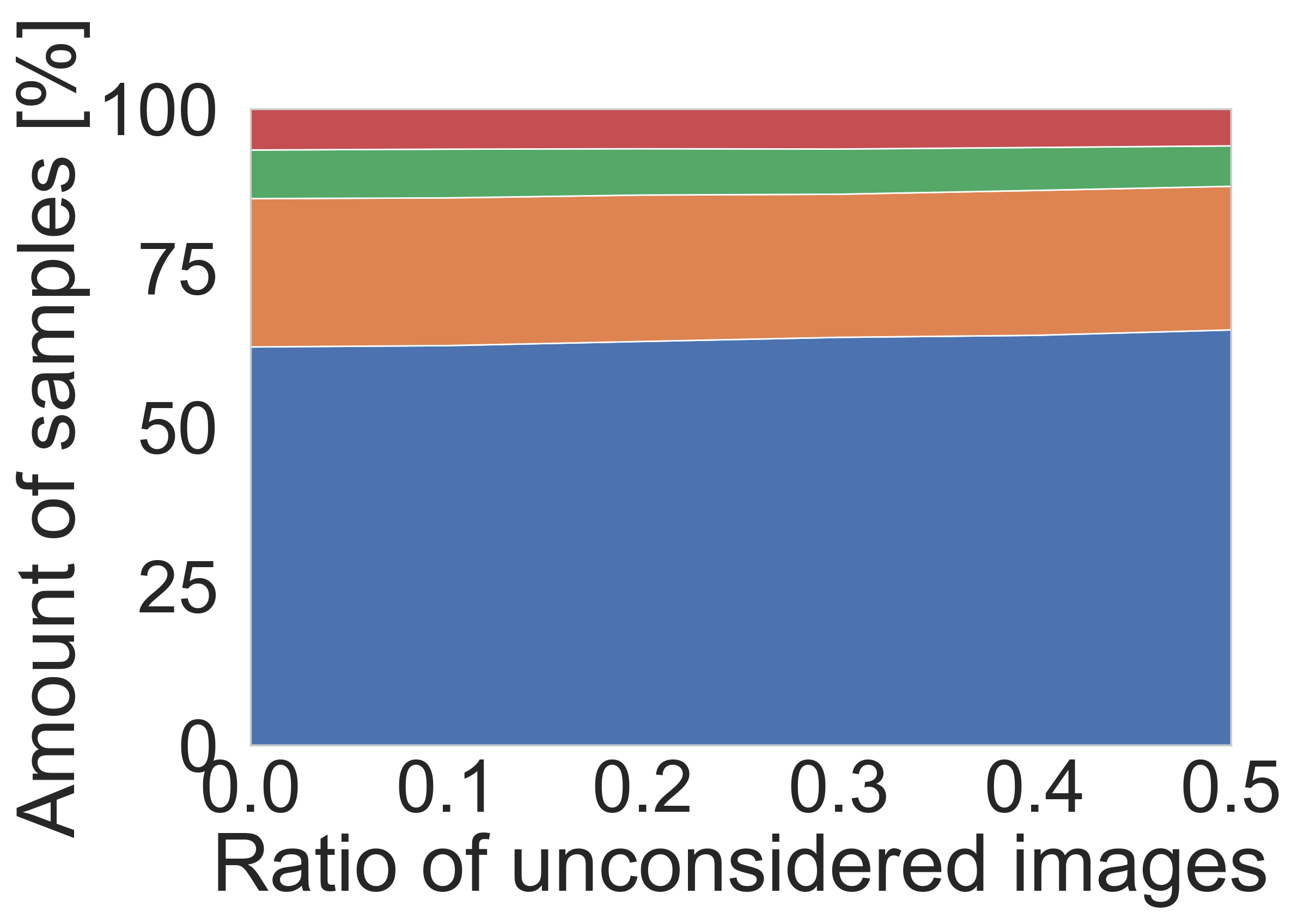}} 
\subfloat[Ethnicity - Best-Rowden \label{fig:StackedCharts_ColorFeret_Ethnicity_ArcFace_BestRowden}]{%
       \includegraphics[width=0.24\textwidth]{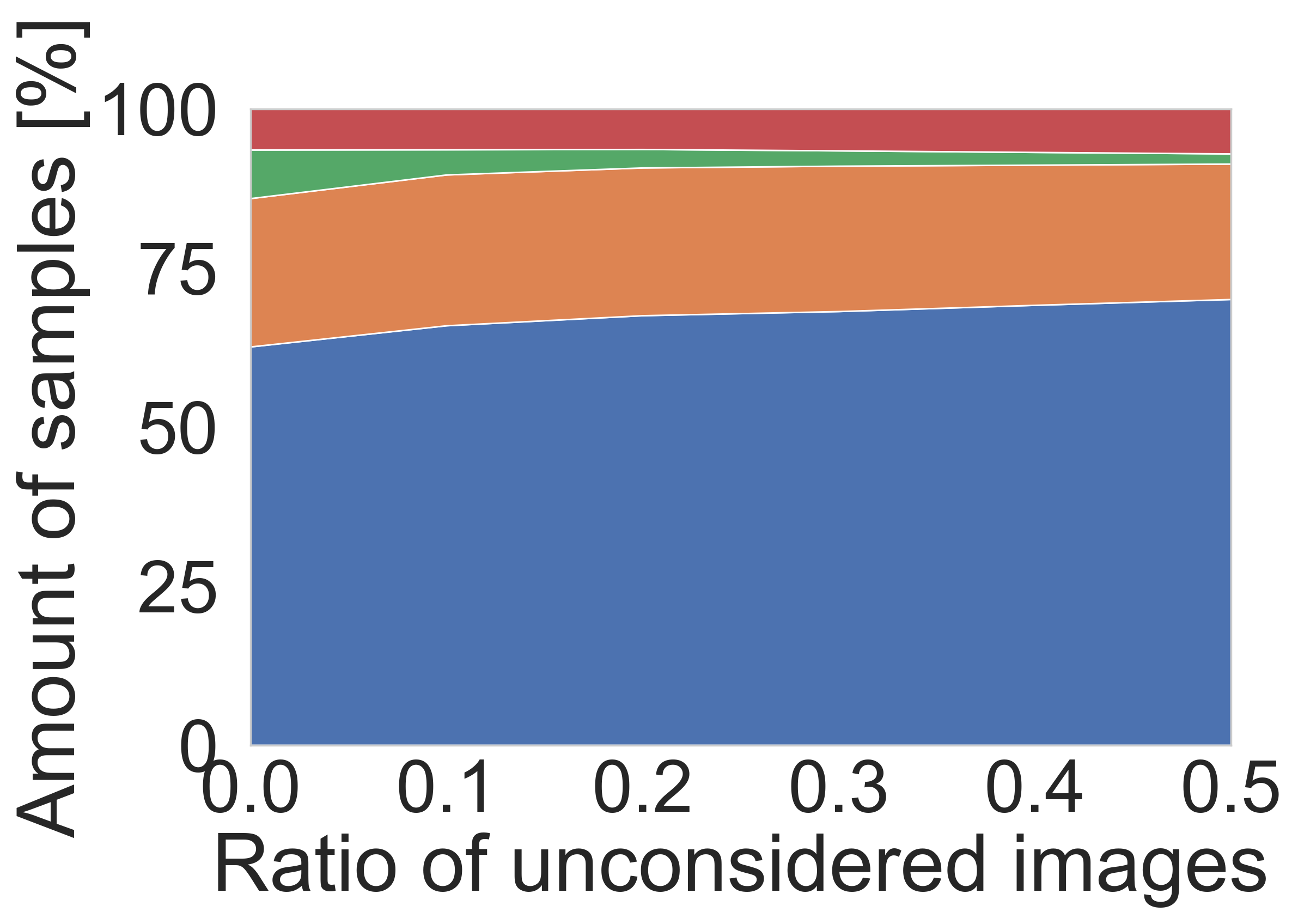}}
\subfloat[Ethnicity - FaceQnet \label{fig:StackedCharts_ColorFeret_Ethnicity_ArcFace_FaceQnet}]{%
       \includegraphics[width=0.24\textwidth]{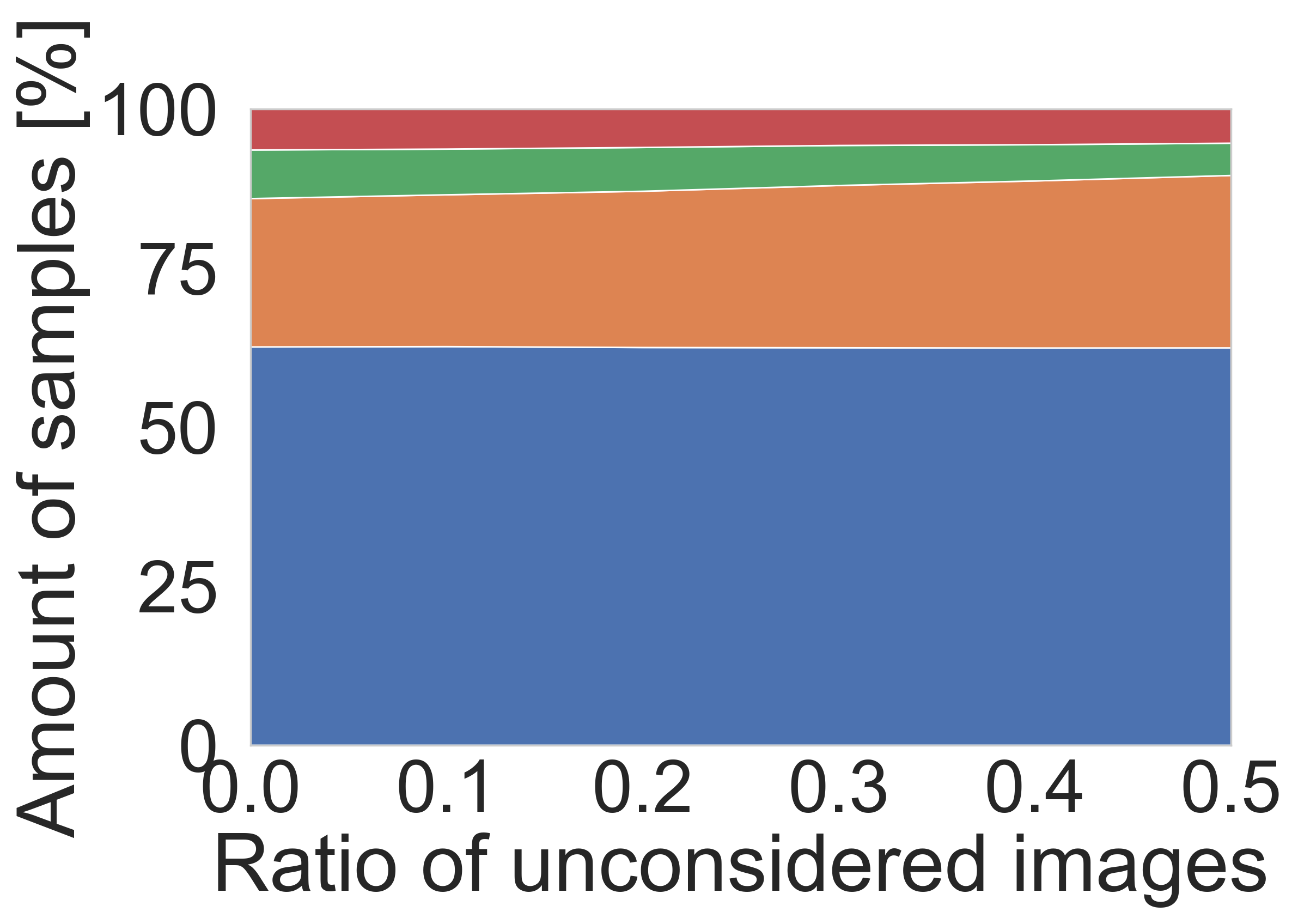}}
\subfloat[Ethnicity - SER-FIQ   \label{fig:StackedCharts_ColorFeret_Ethnicity_ArcFace_SER-FIQ}]{%
       \includegraphics[width=0.24\textwidth]{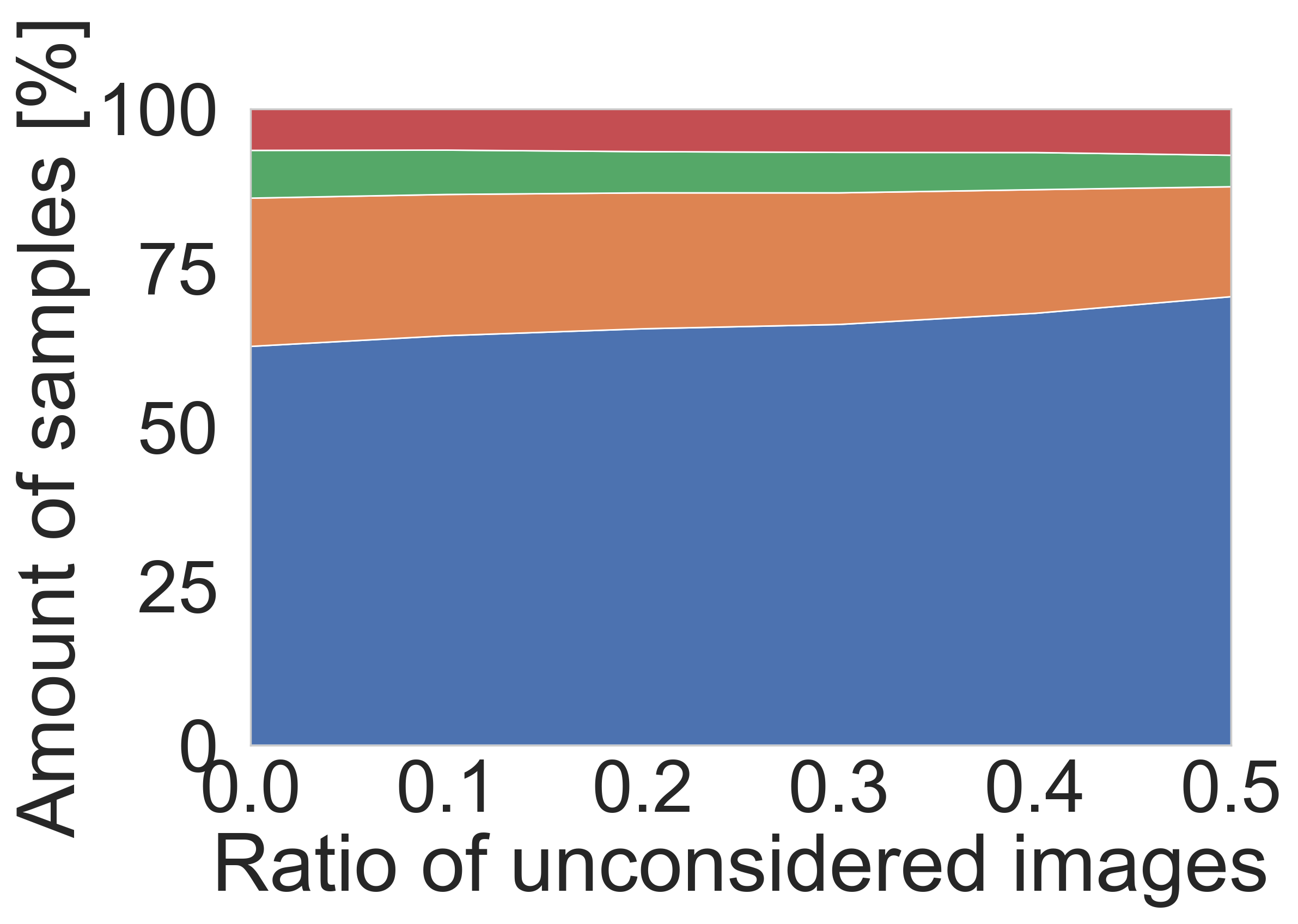}}
       \adjustbox{raise=6mm}{\includegraphics[height=19mm]{colorferet_ethnics.png}} 
\vspace{-3mm}
       
\subfloat[Age - COTS \label{fig:StackedCharts_Adience_Age_ArcFace_COTS}]{%
       \includegraphics[width=0.24\textwidth]{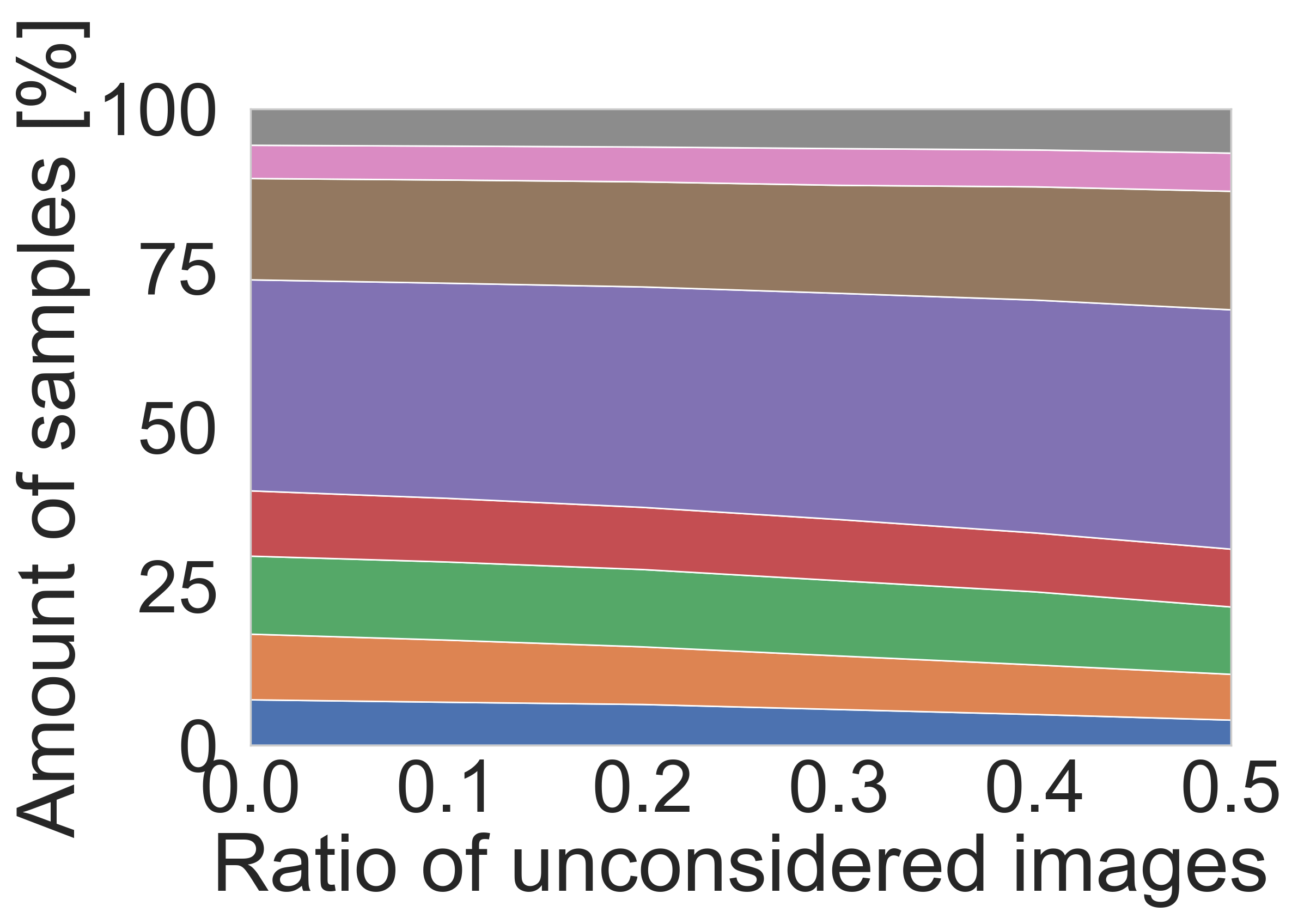}}
\subfloat[Age - Best-Rowden \label{fig:StackedCharts_Adience_Age_ArcFace_BestRowden}]{%
       \includegraphics[width=0.24\textwidth]{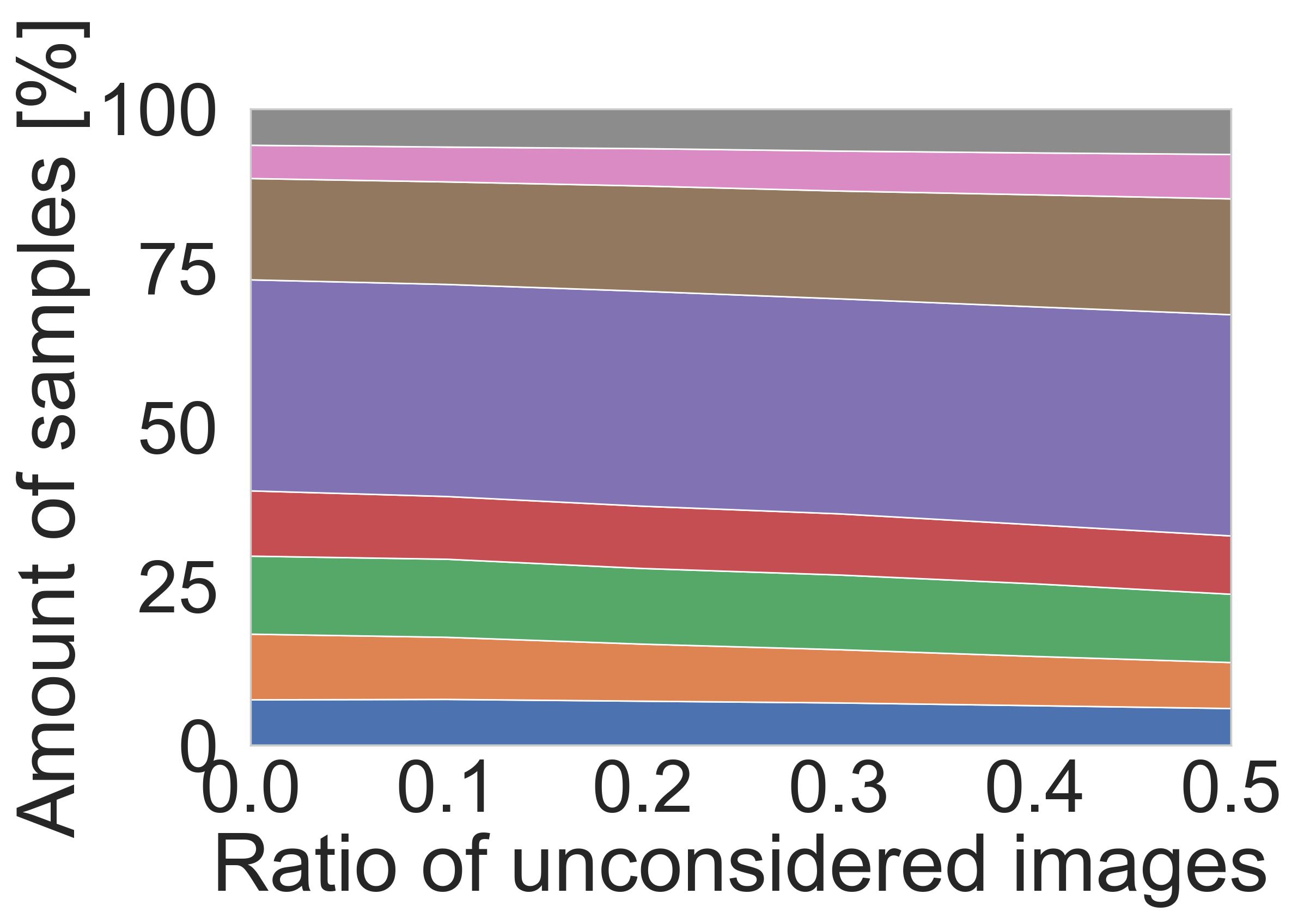}}
\subfloat[Age - FaceQnet \label{fig:StackedCharts_Adience_Age_ArcFace_FaceQnet}]{%
       \includegraphics[width=0.24\textwidth]{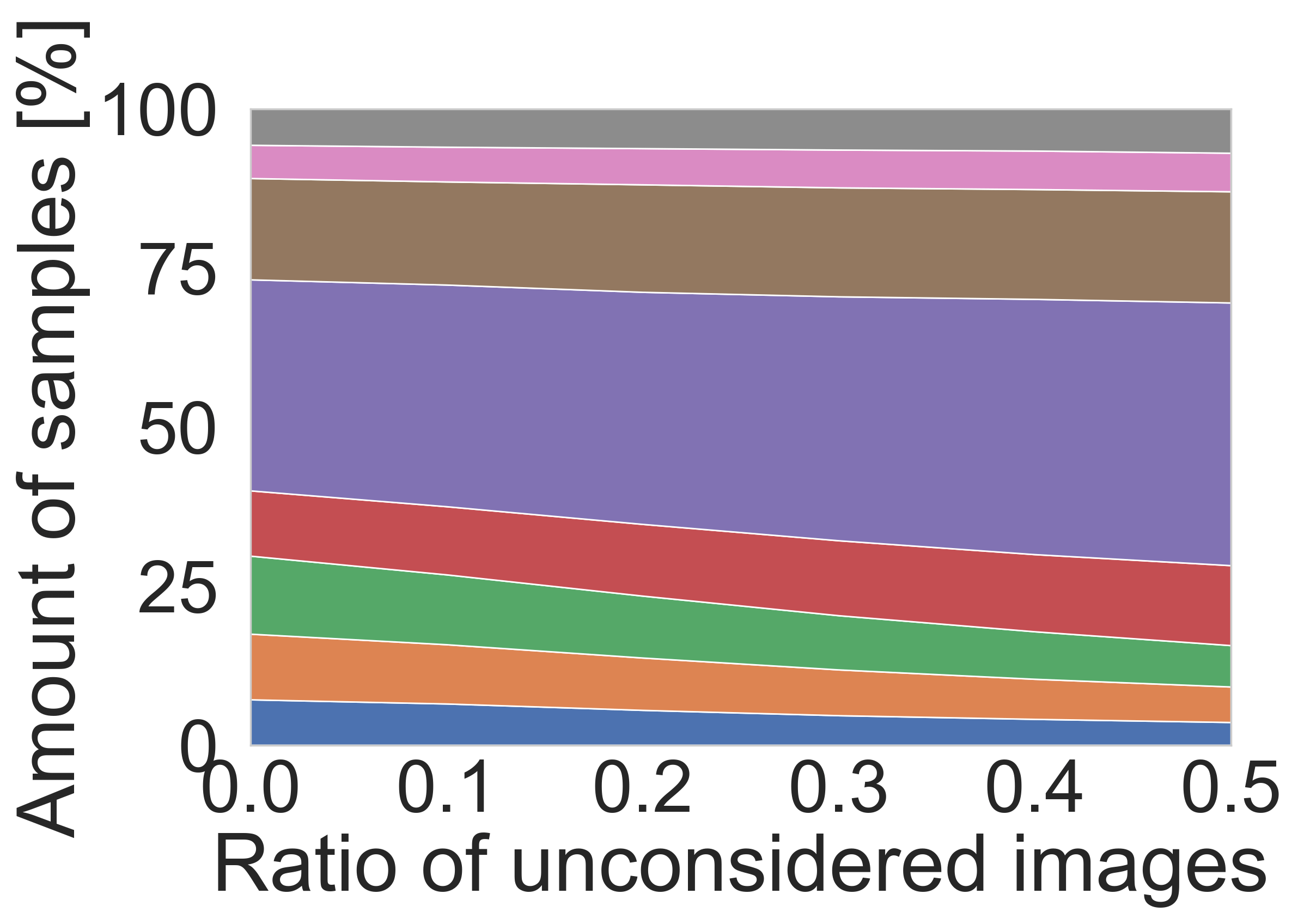}}
\subfloat[Age - SER-FIQ   \label{fig:StackedCharts_Adience_Age_ArcFace_SER-FIQ}]{%
       \includegraphics[width=0.24\textwidth]{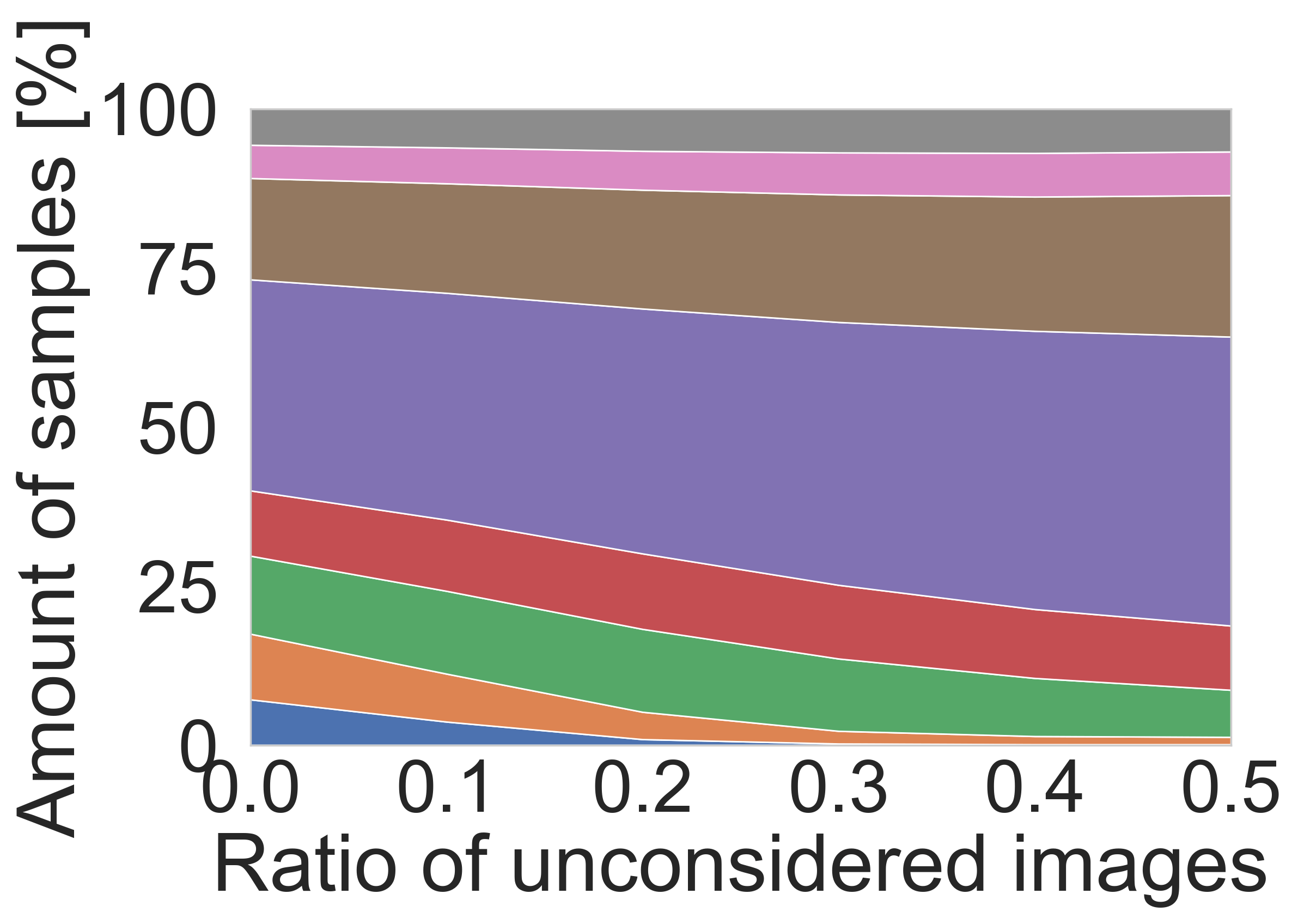}}
       \adjustbox{raise=0mm}{\includegraphics[height=28mm]{adience_age.png}} 
\vspace{-2mm}
\caption{Analysis of the proportion of subgroups for ArcFace embeddings. The pose (a-d), ethnicities (e-h), and age (i-l) proportions are shown when applying several quality thresholds.}
\label{fig:StackedCharts_ArcFace}
\vspace{-3mm} 
\end{figure*}

\begin{figure*}[h]
\captionsetup[subfloat]{farskip=2pt,captionskip=1pt}
\centering
\subfloat[Pose - COTS \label{fig:QualityDitribution_ColorFeret_Pose_ArcFace_COTS}]{%
       \includegraphics[width=0.24\textwidth]{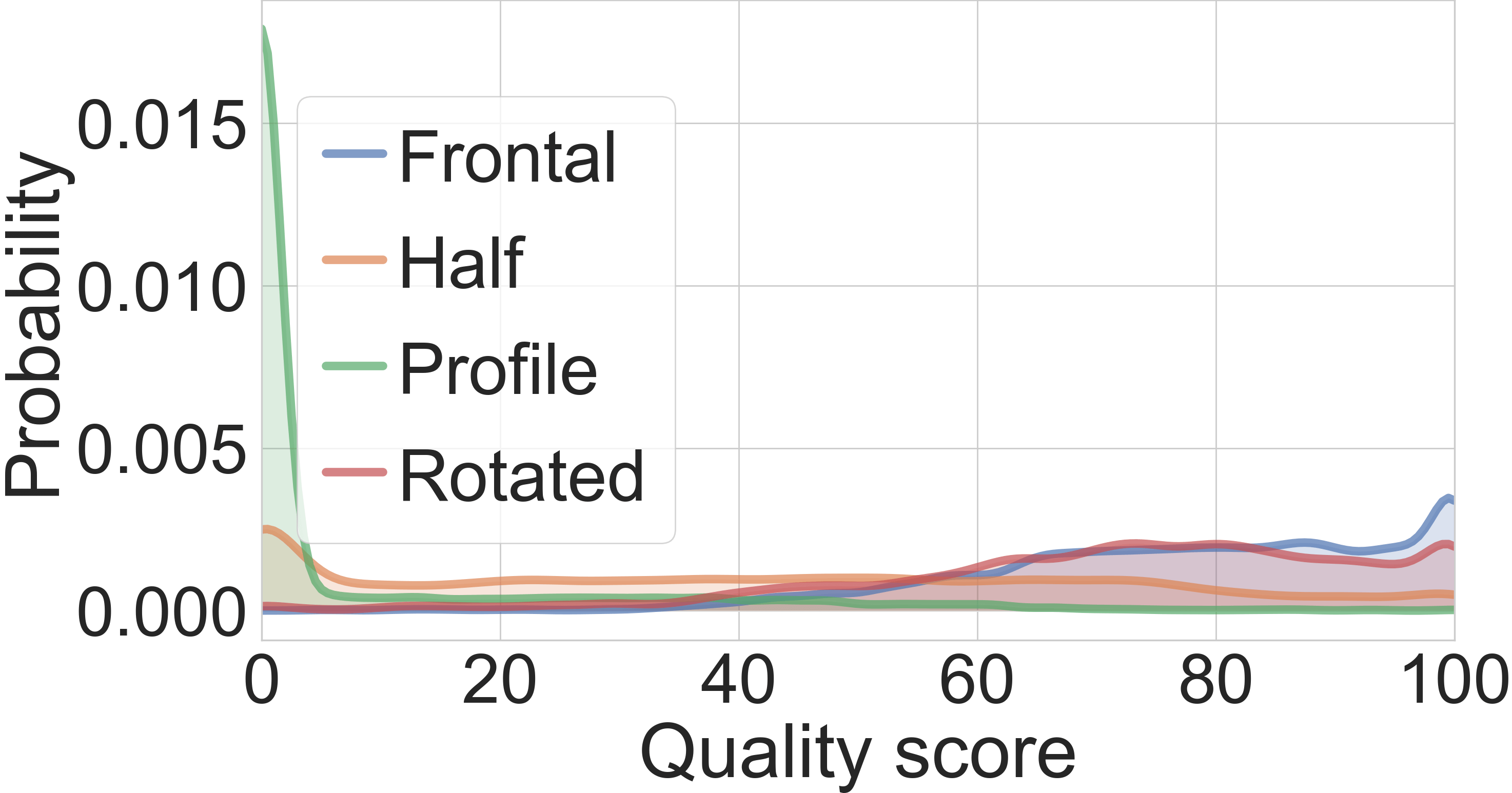}} \hspace{5mm}
\subfloat[Ethnics - COTS \label{fig:QualityDitribution_ColorFeret_Ethnicity_ArcFace_COTS}]{%
       \includegraphics[width=0.24\textwidth]{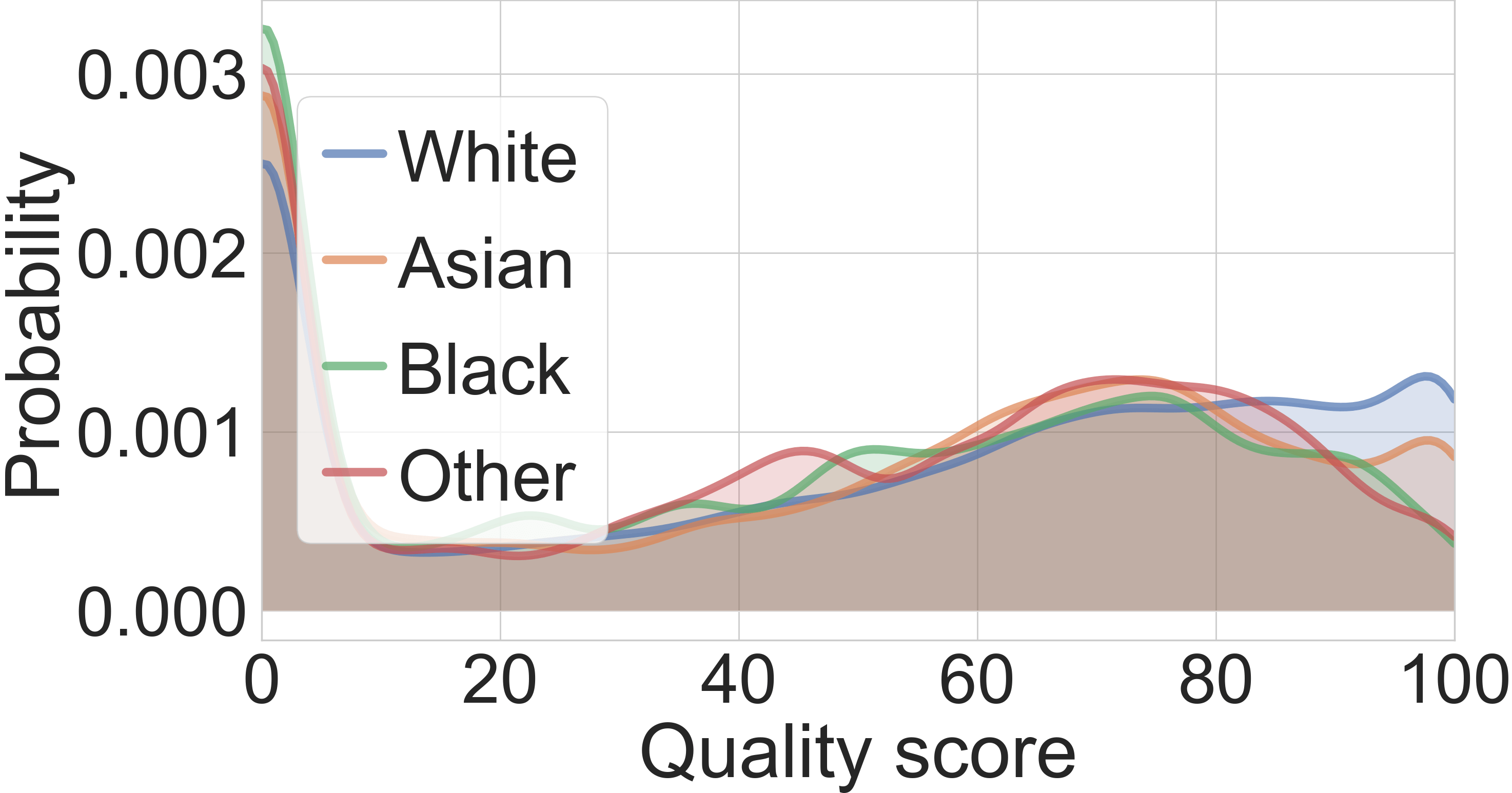}}     \hspace{5mm}  
\subfloat[Age - COTS \label{fig:QualityDitribution_Adience_Age_ArcFace_COTS}]{%
       \includegraphics[width=0.24\textwidth]{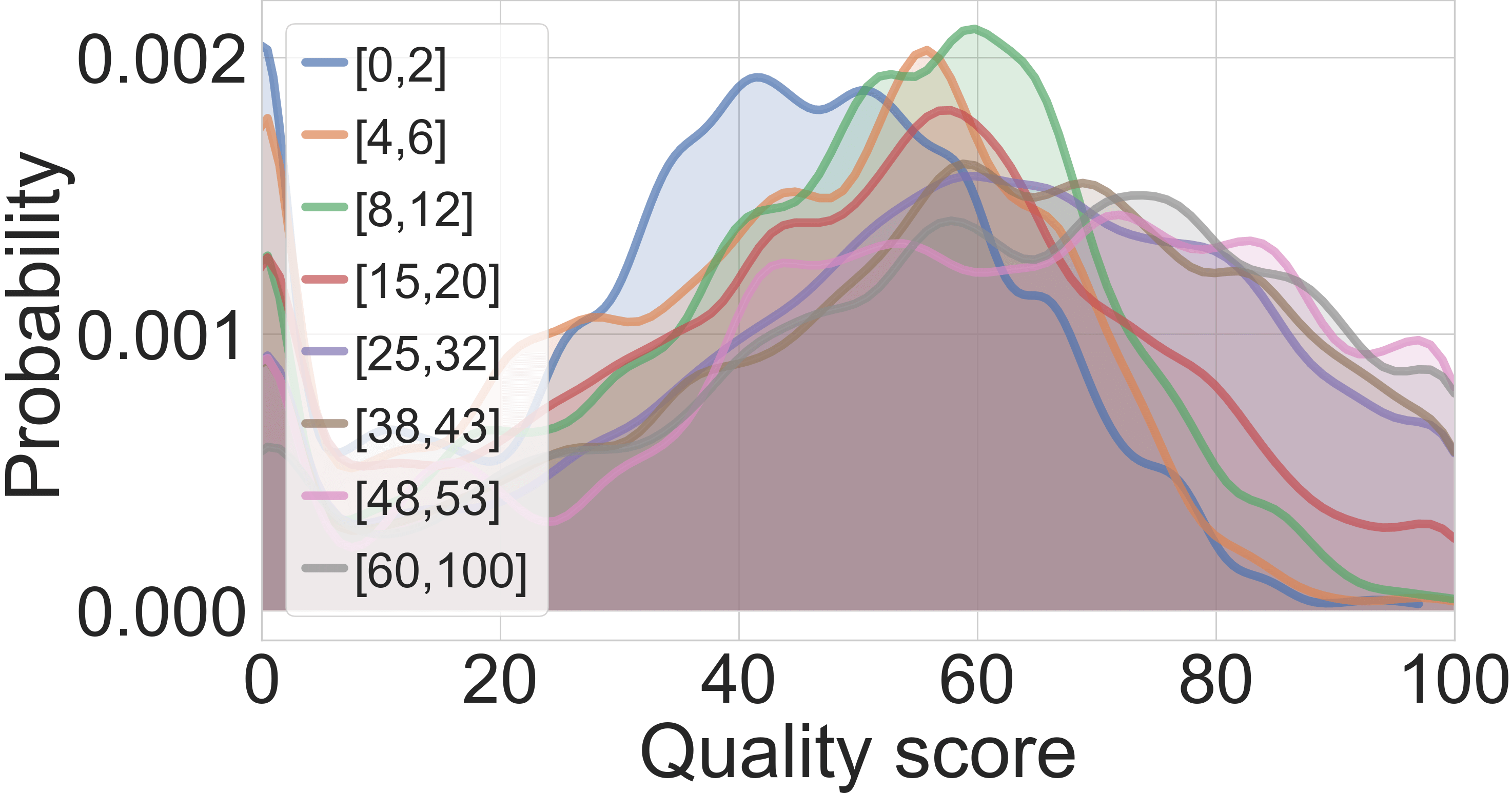}}
       
\subfloat[Pose - Best-Rowden - FaceNet \label{fig:QualityDitribution_ColorFeret_Pose_FaceNet_BestRowden}]{%
       \includegraphics[width=0.24\textwidth]{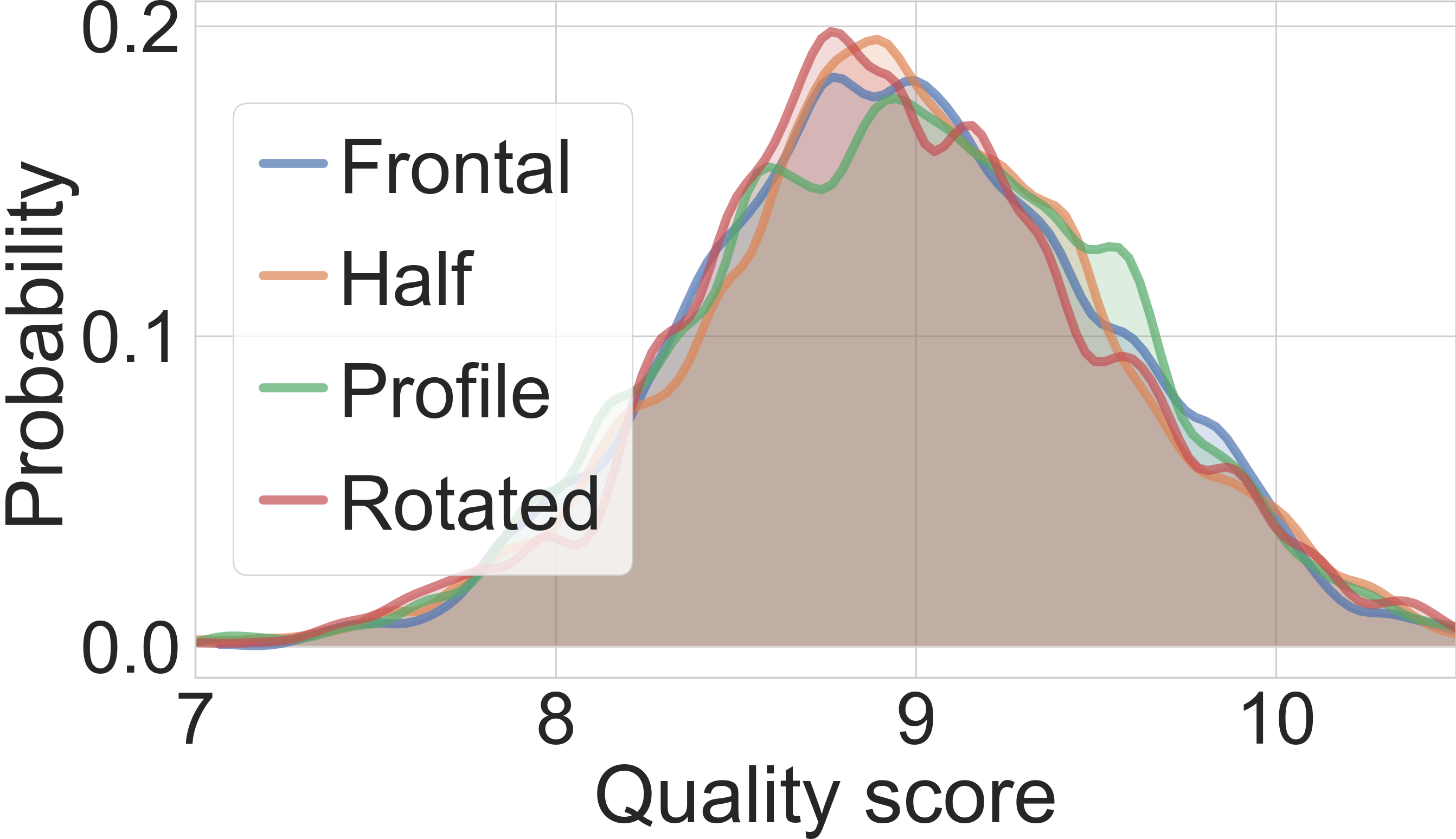}}\hspace{5mm}
\subfloat[Ethnics - Best-Rowden - FaceNet \label{fig:QualityDitribution_ColorFeret_Ethnicity_FaceNet_BestRowden}]{%
       \includegraphics[width=0.24\textwidth]{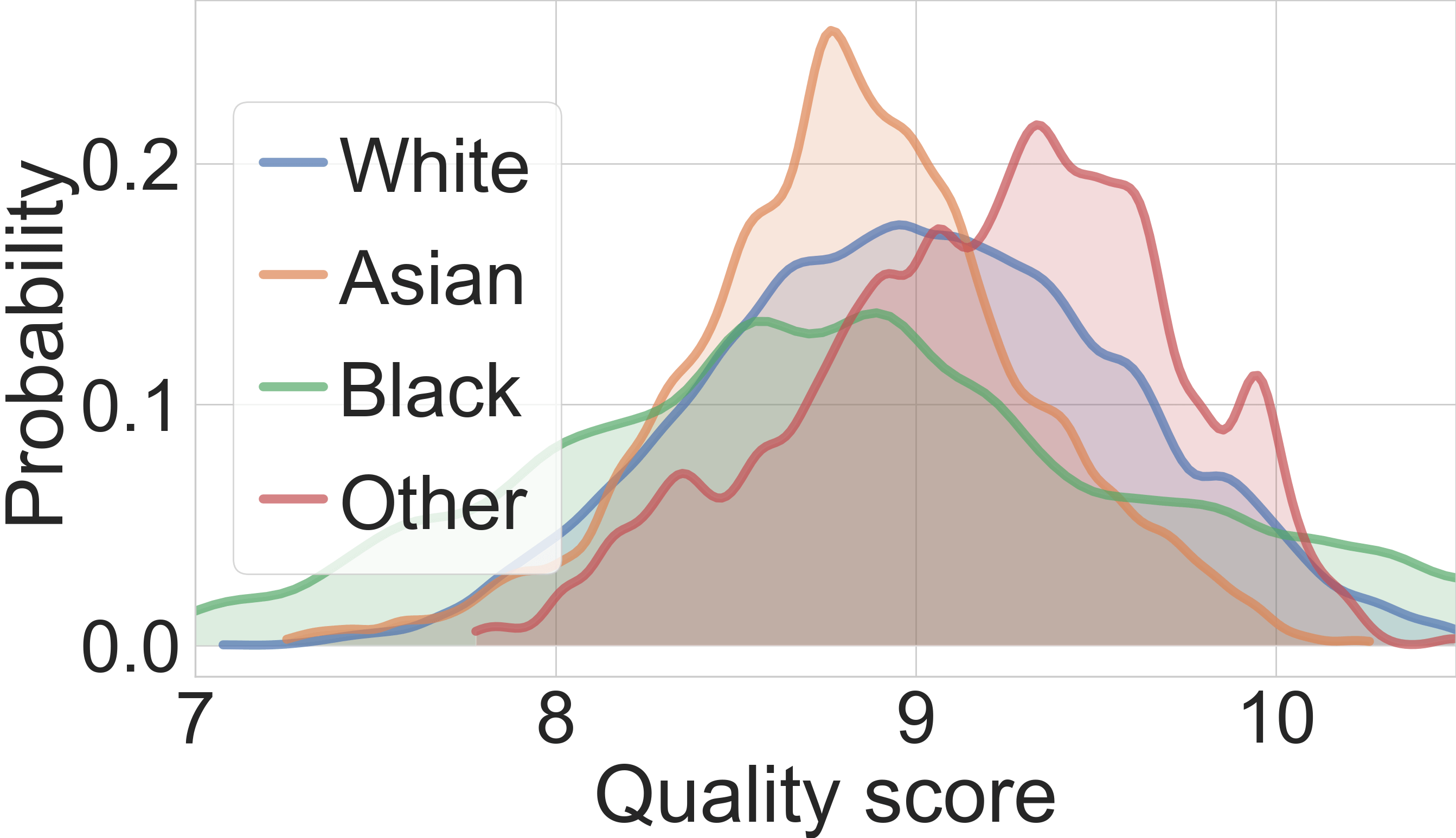}}  \hspace{5mm}     
\subfloat[Age - Best-Rowden - FaceNet \label{fig:QualityDitribution_Adience_Age_FaceNet_BestRowden}]{%
       \includegraphics[width=0.24\textwidth]{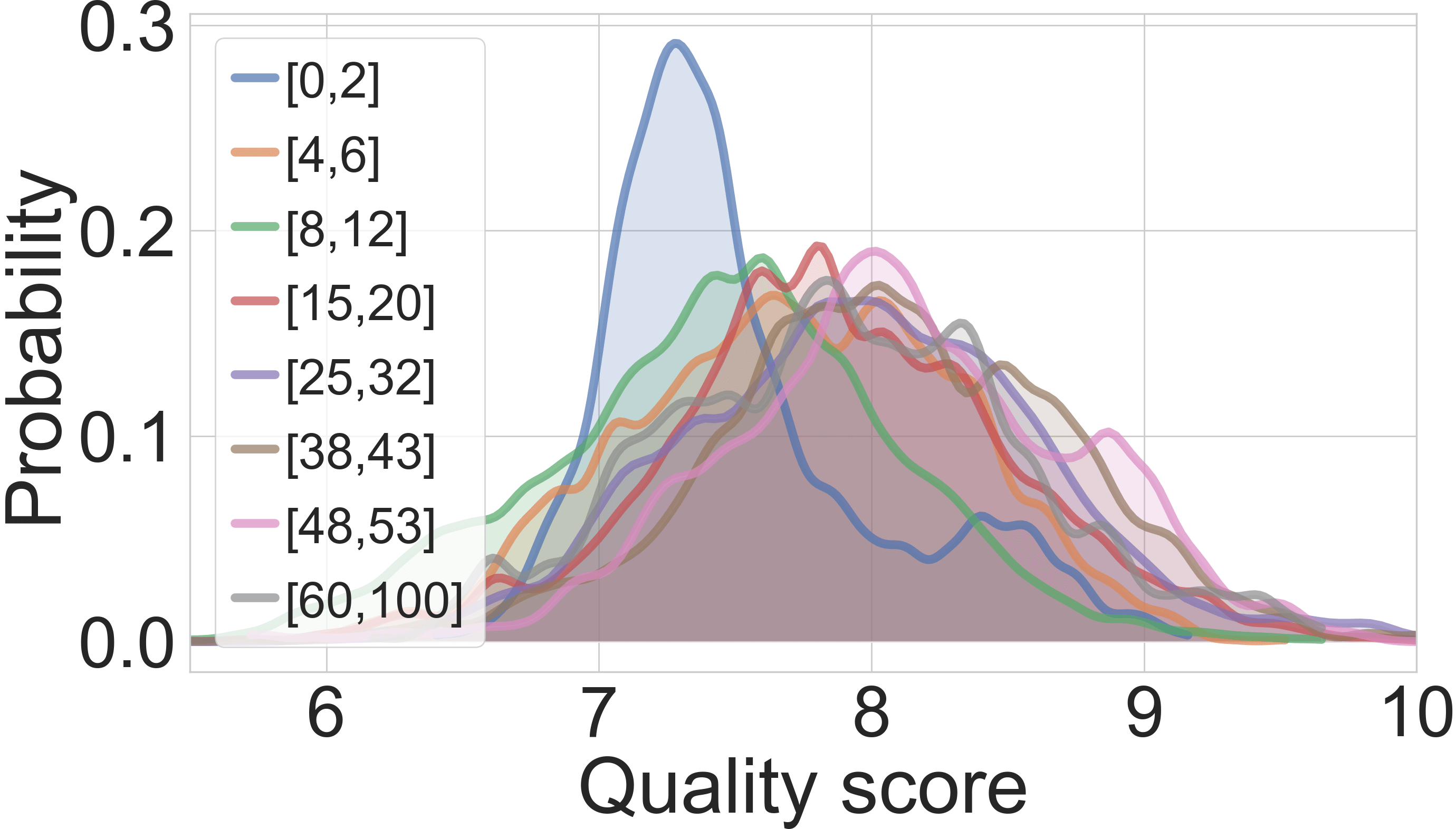}}       
    
\subfloat[Pose - Best-Rowden - ArcFace \label{fig:QualityDitribution_ColorFeret_Pose_ArcFace_BestRowden}]{%
       \includegraphics[width=0.24\textwidth]{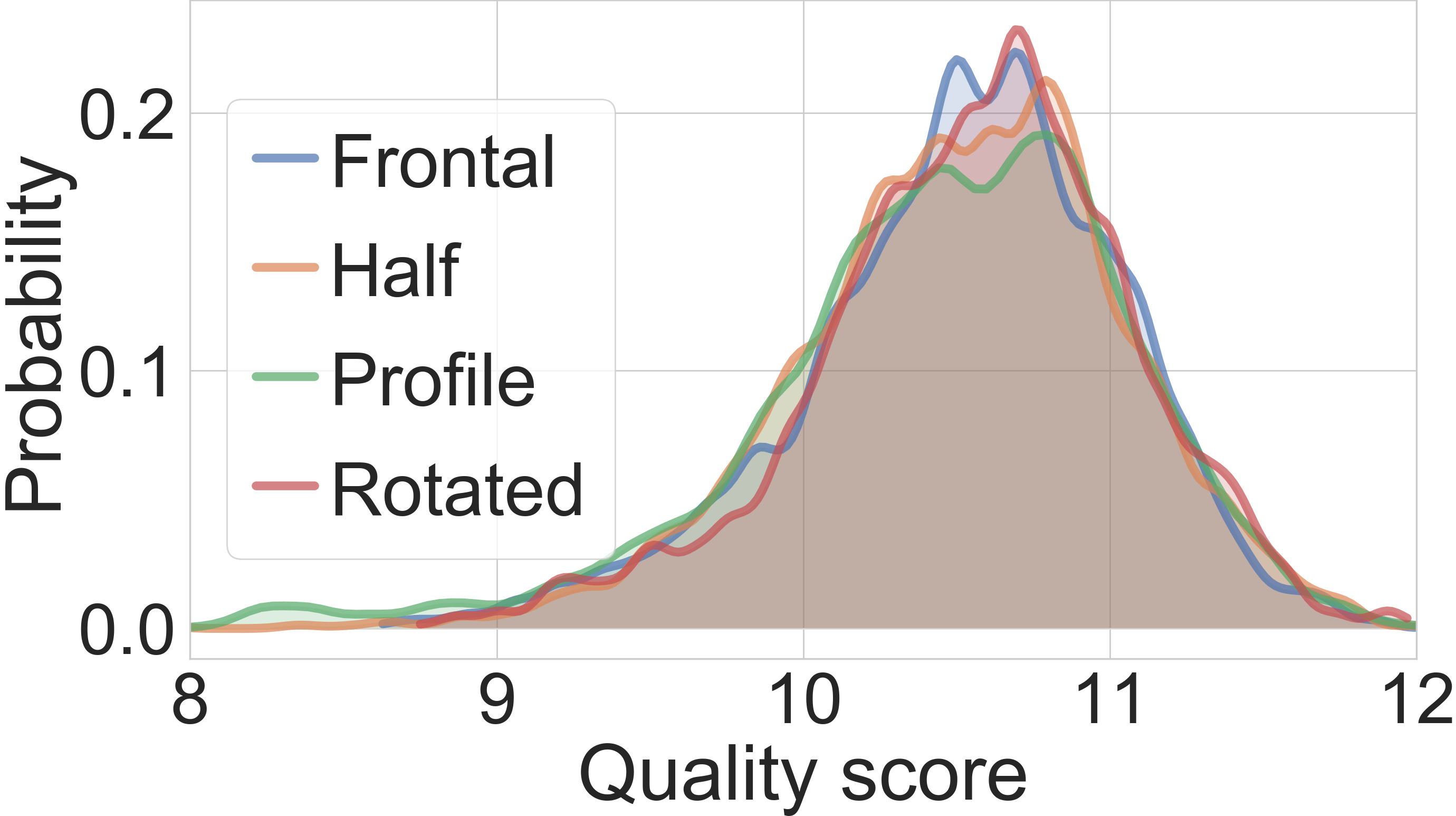}}\hspace{5mm}
\subfloat[Ethnics - Best-Rowden - ArcFace \label{fig:QualityDitribution_ColorFeret_Ethnicity_ArcFace_BestRowden}]{%
       \includegraphics[width=0.24\textwidth]{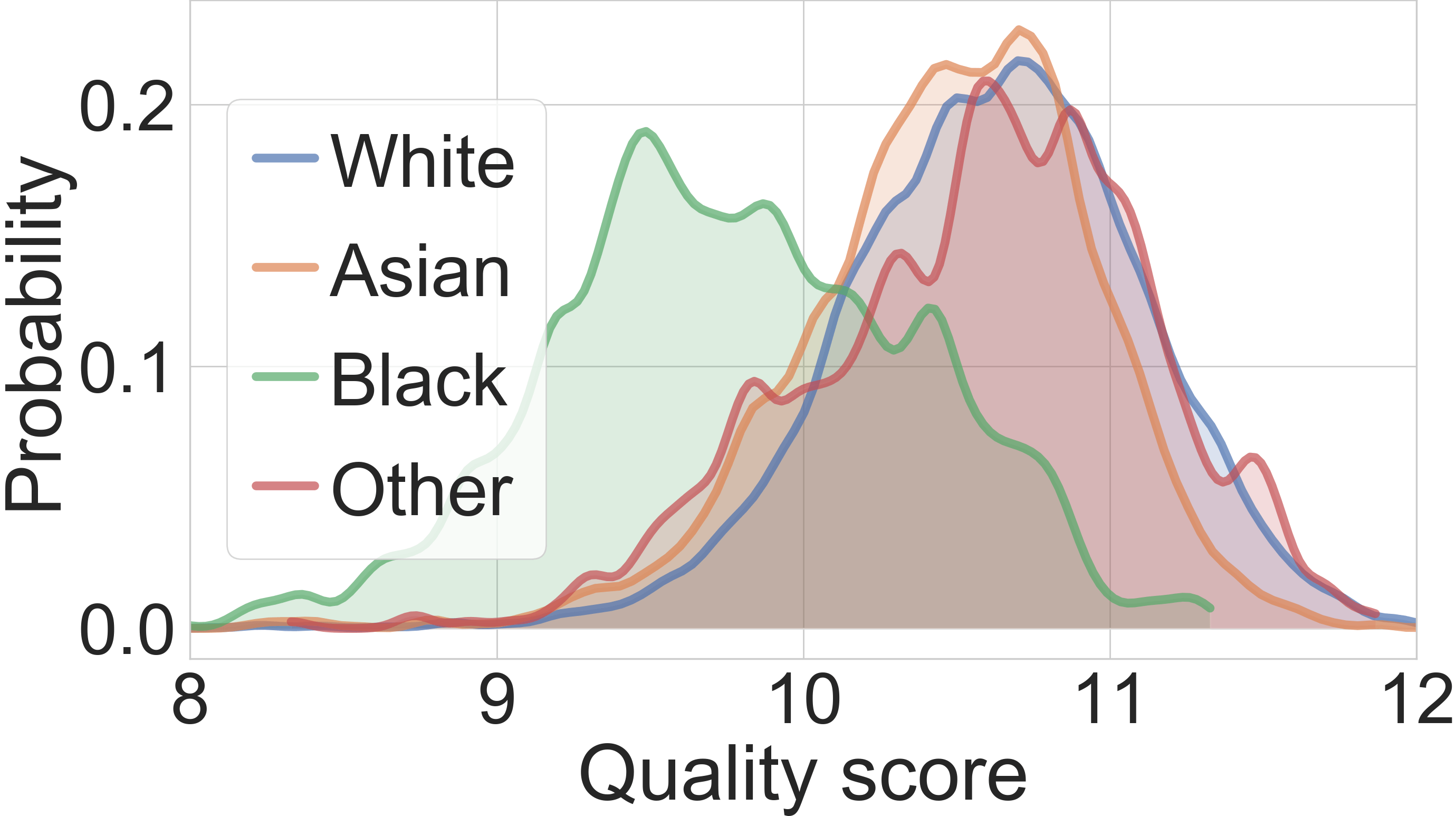}}\hspace{5mm}
\subfloat[Age - Best-Rowden - ArcFace \label{fig:QualityDitribution_Adience_Age_ArcFace_BestRowden}]{%
       \includegraphics[width=0.24\textwidth]{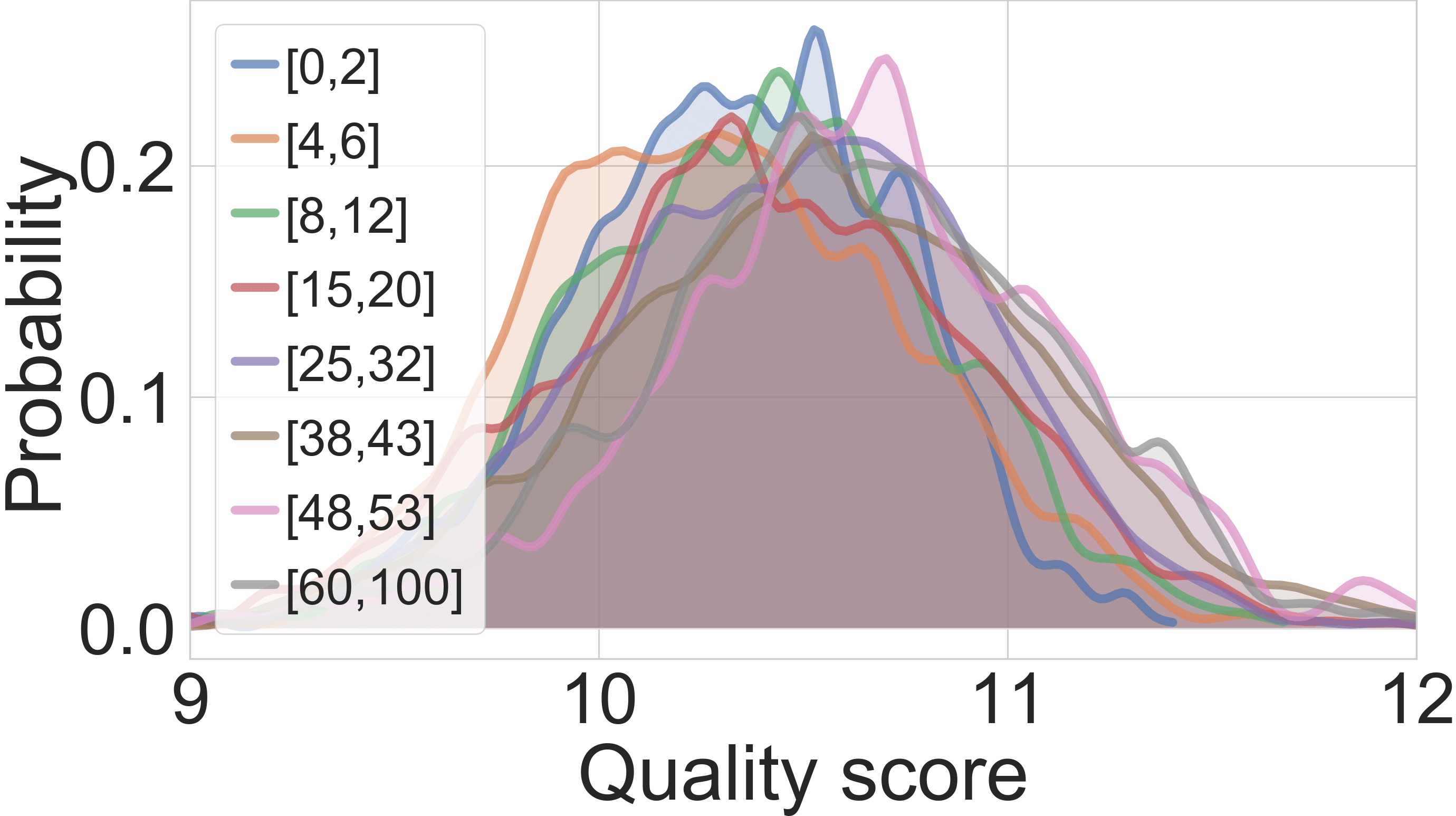}}
\vspace{-2mm}

\subfloat[Pose - FaceQnet \label{fig:QualityDitribution_ColorFeret_Pose_ArcFace_FaceQnet}]{%
       \includegraphics[width=0.24\textwidth]{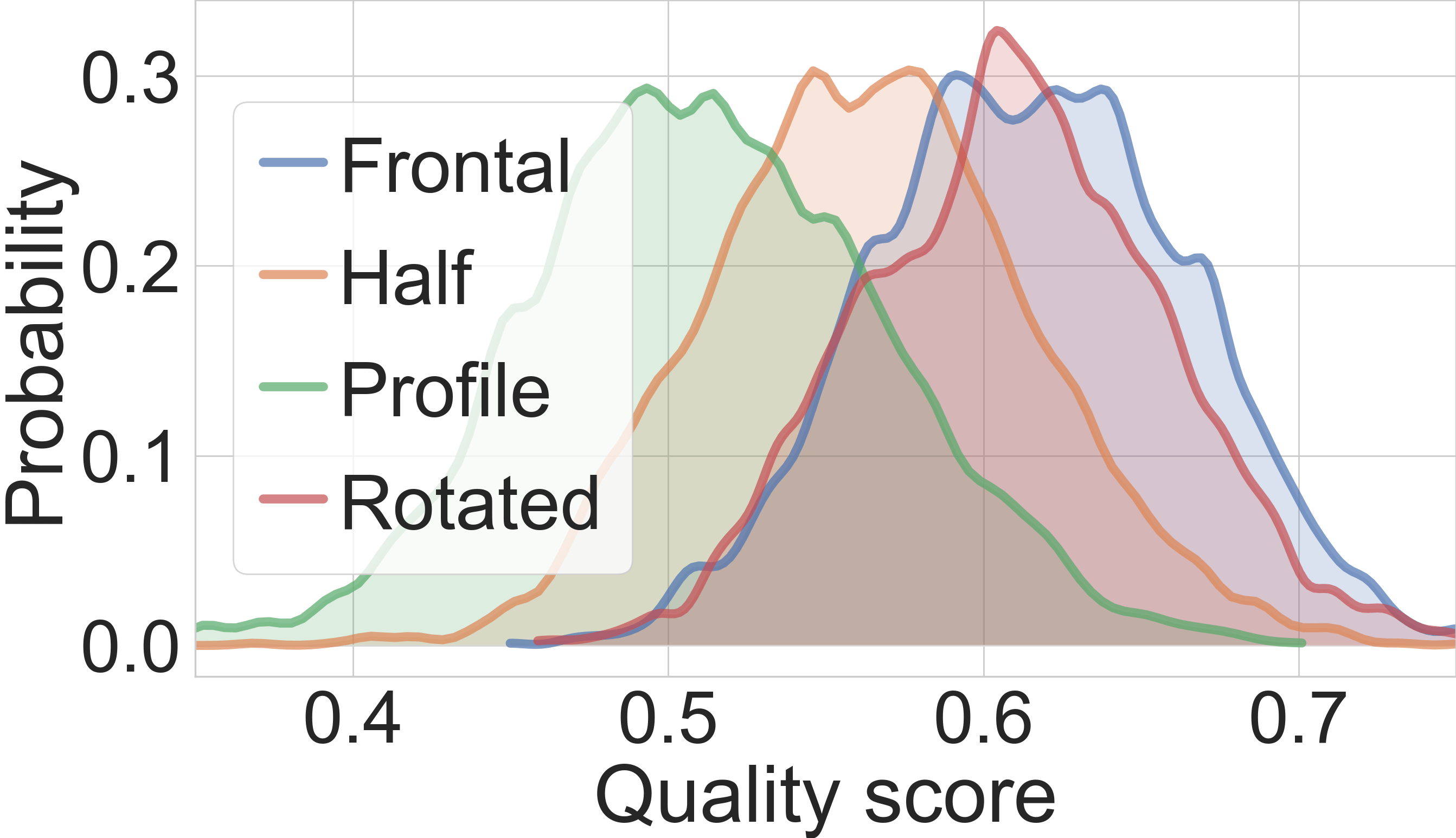}}\hspace{5mm}
\subfloat[Ethnics - FaceQnet \label{fig:QualityDitribution_ColorFeret_Ethnicity_ArcFace_FaceQnet}]{%
       \includegraphics[width=0.24\textwidth]{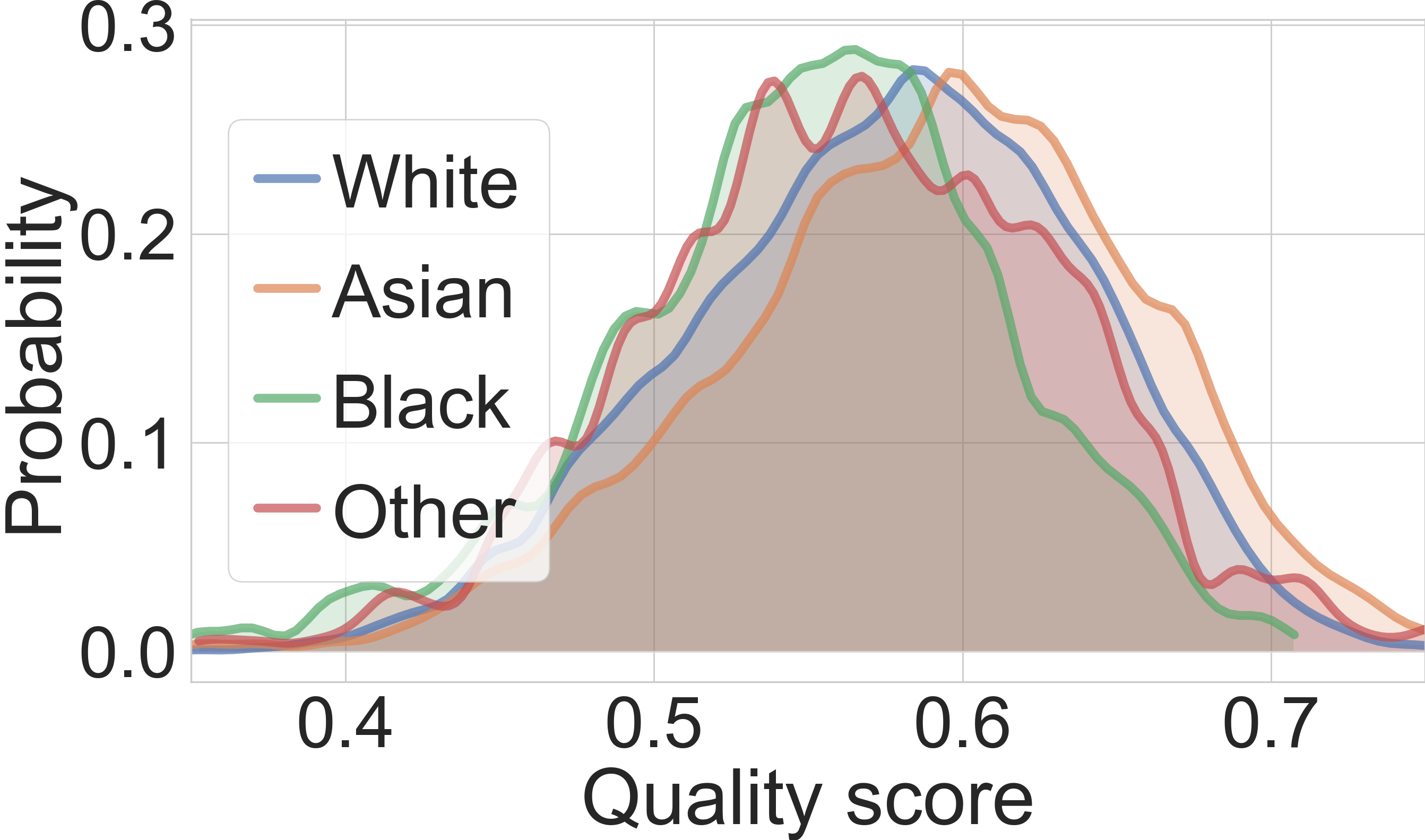}} \hspace{5mm}      
\subfloat[Age - FaceQnet \label{fig:QualityDitribution_Adience_Age_ArcFace_FaceQnet}]{%
       \includegraphics[width=0.24\textwidth]{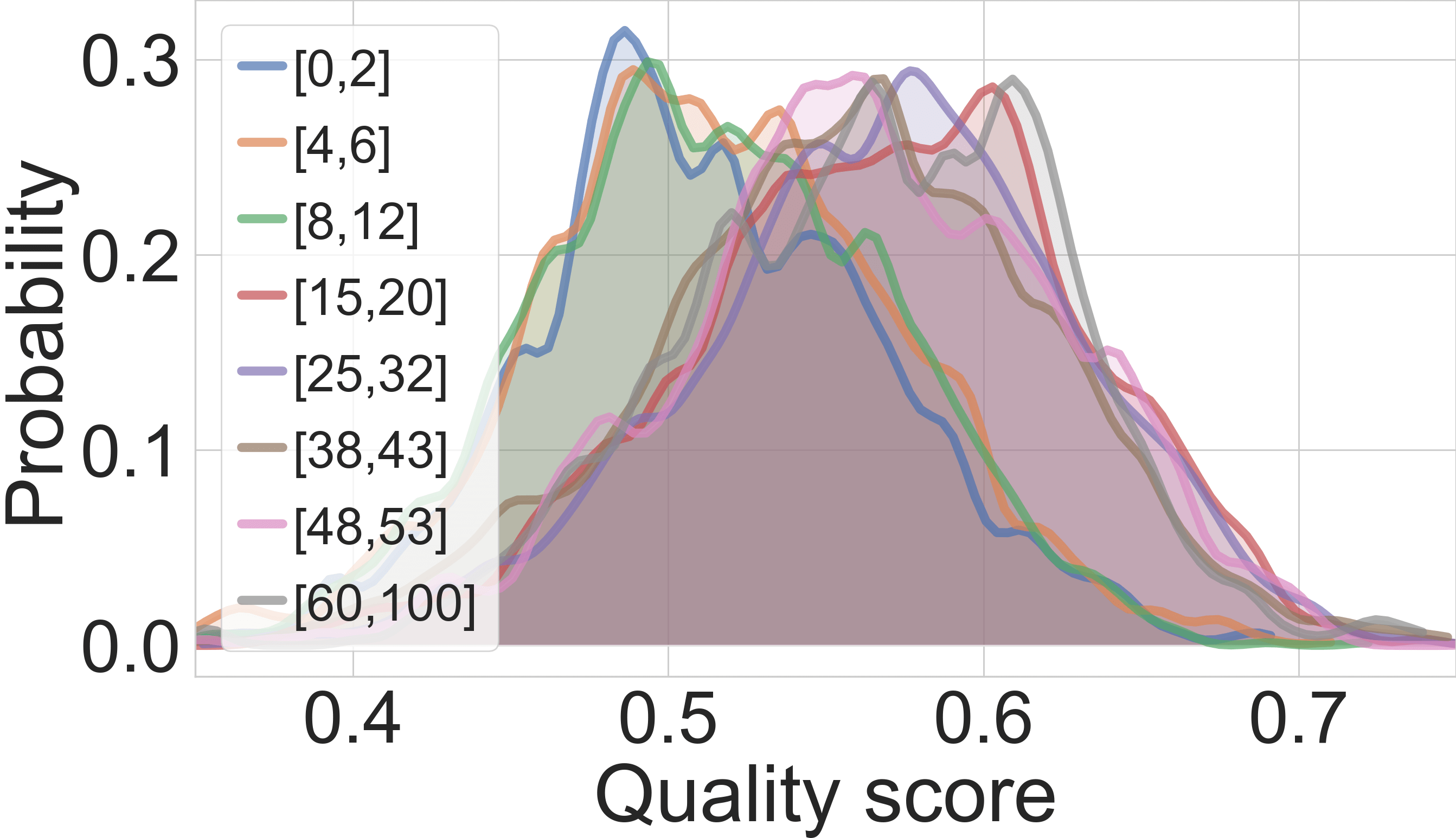}}
       
\subfloat[Pose - SER-FIQ - FaceNet  \label{fig:QualityDitribution_ColorFeret_Pose_FaceNet_SER-FIQ}]{%
       \includegraphics[width=0.24\textwidth]{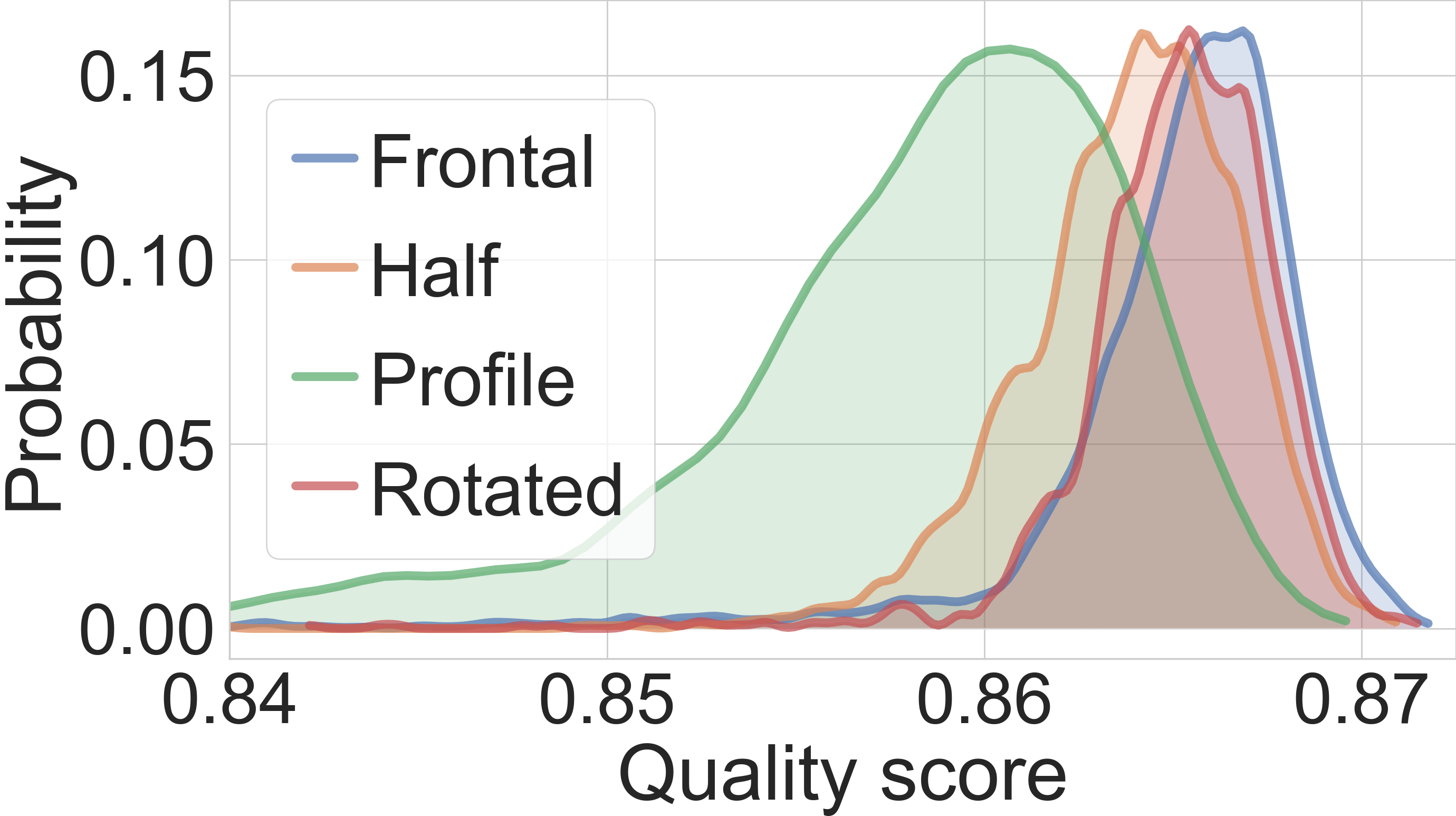}}\hspace{5mm}
\subfloat[Ethnics - SER-FIQ - FaceNet   \label{fig:QualityDitribution_ColorFeret_Ethnicity_FaceNet_SER-FIQ}]{\includegraphics[width=0.24\textwidth]{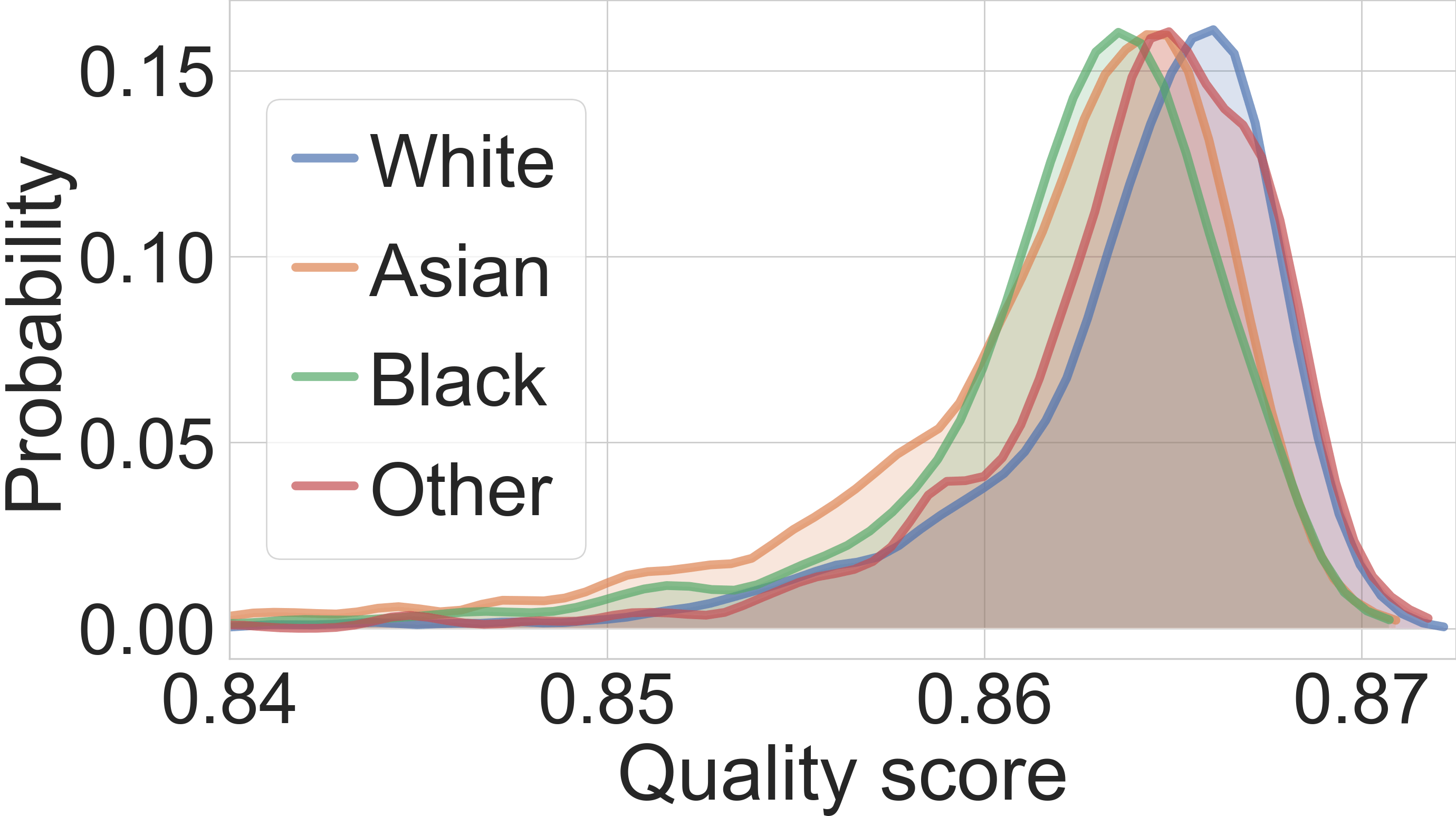}}    \hspace{5mm}   
\subfloat[Age - SER-FIQ - FaceNet   \label{fig:QualityDitribution_Adience_Age_FaceNet_SER-FIQ}]{%
       \includegraphics[width=0.24\textwidth]{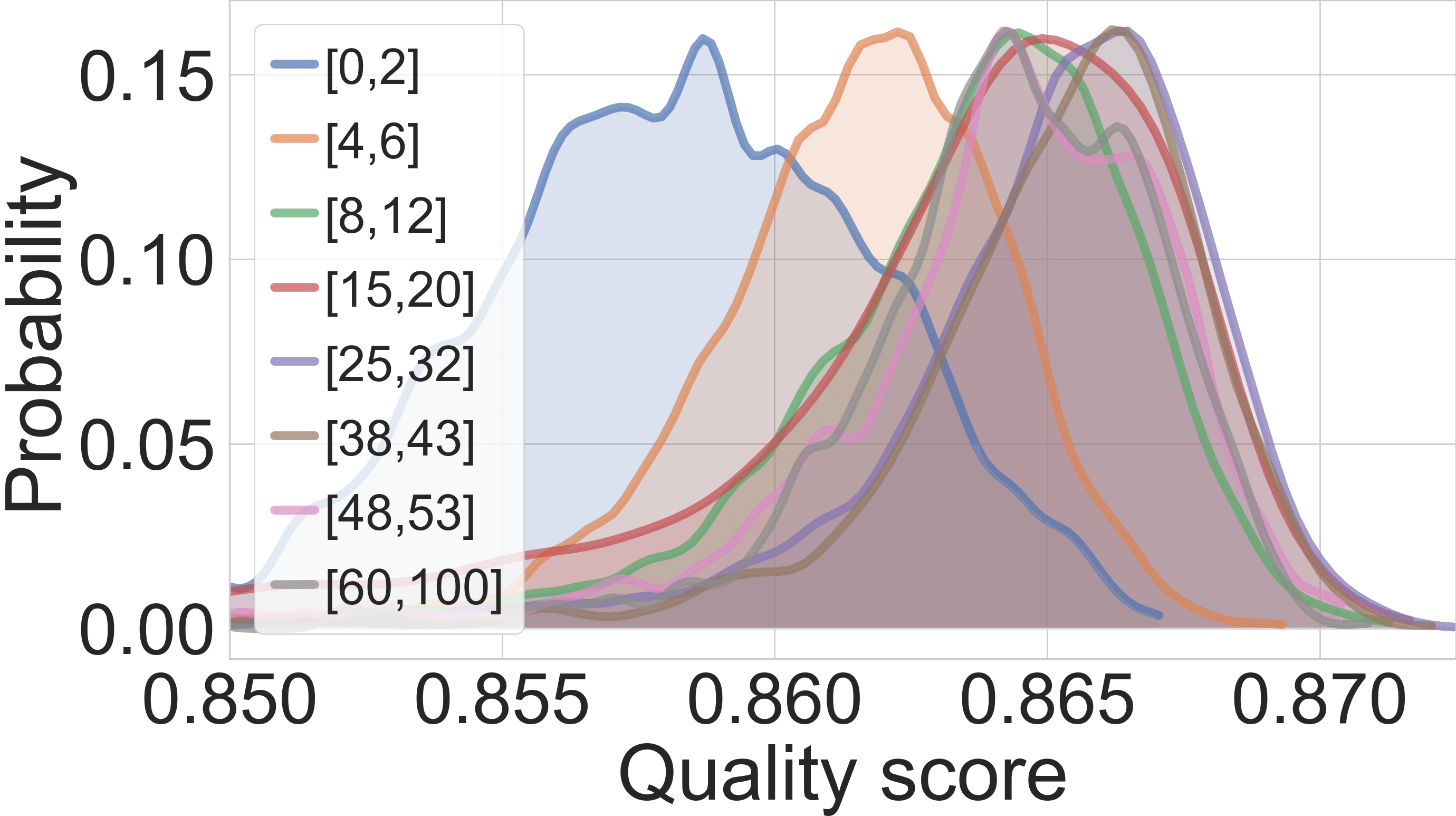}} 

\subfloat[Pose - SER-FIQ - ArcFace   \label{fig:QualityDitribution_ColorFeret_Pose_ArcFace_SER-FIQ}]{%
       \includegraphics[width=0.24\textwidth]{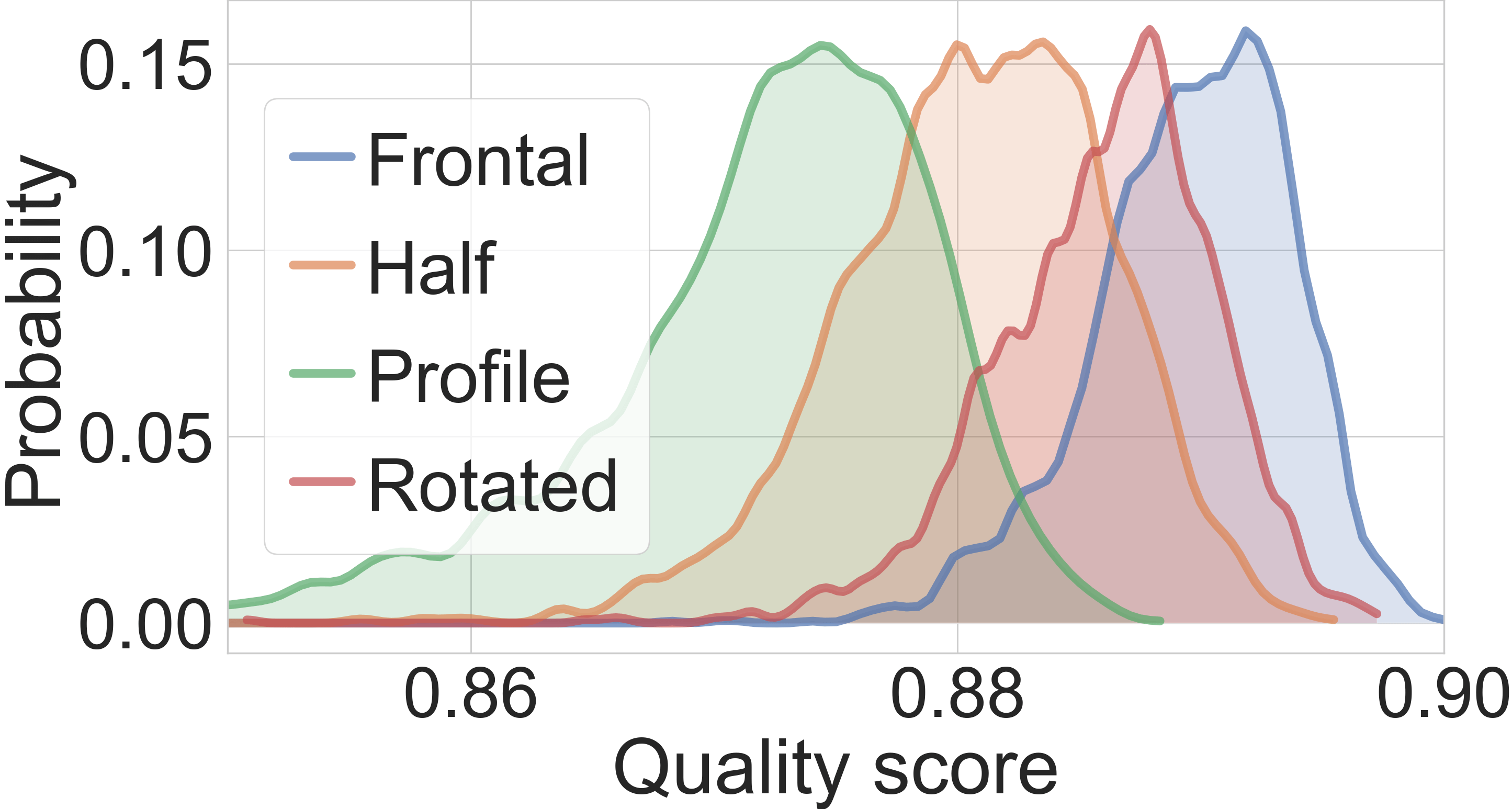}}\hspace{5mm}
\subfloat[Ethnics - SER-FIQ - ArcFace   \label{fig:QualityDitribution_ColorFeret_Ethnicity_ArcFace_SER-FIQ}]{
       \includegraphics[width=0.24\textwidth]{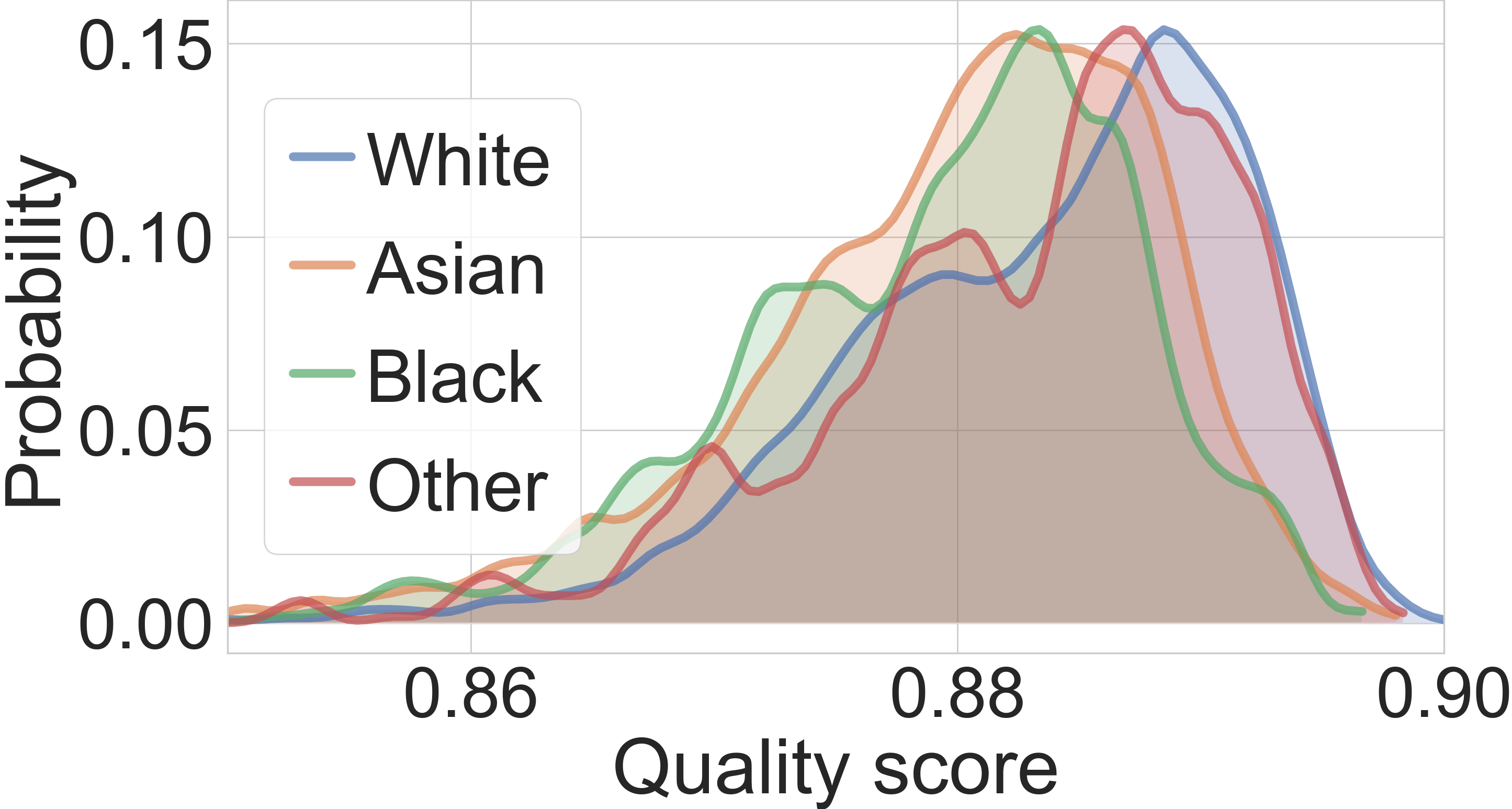}} \hspace{5mm}
\subfloat[Age - SER-FIQ - ArcFace   \label{fig:QualityDitribution_Adience_Age_ArcFace_SER-FIQ}]{%
       \includegraphics[width=0.24\textwidth]{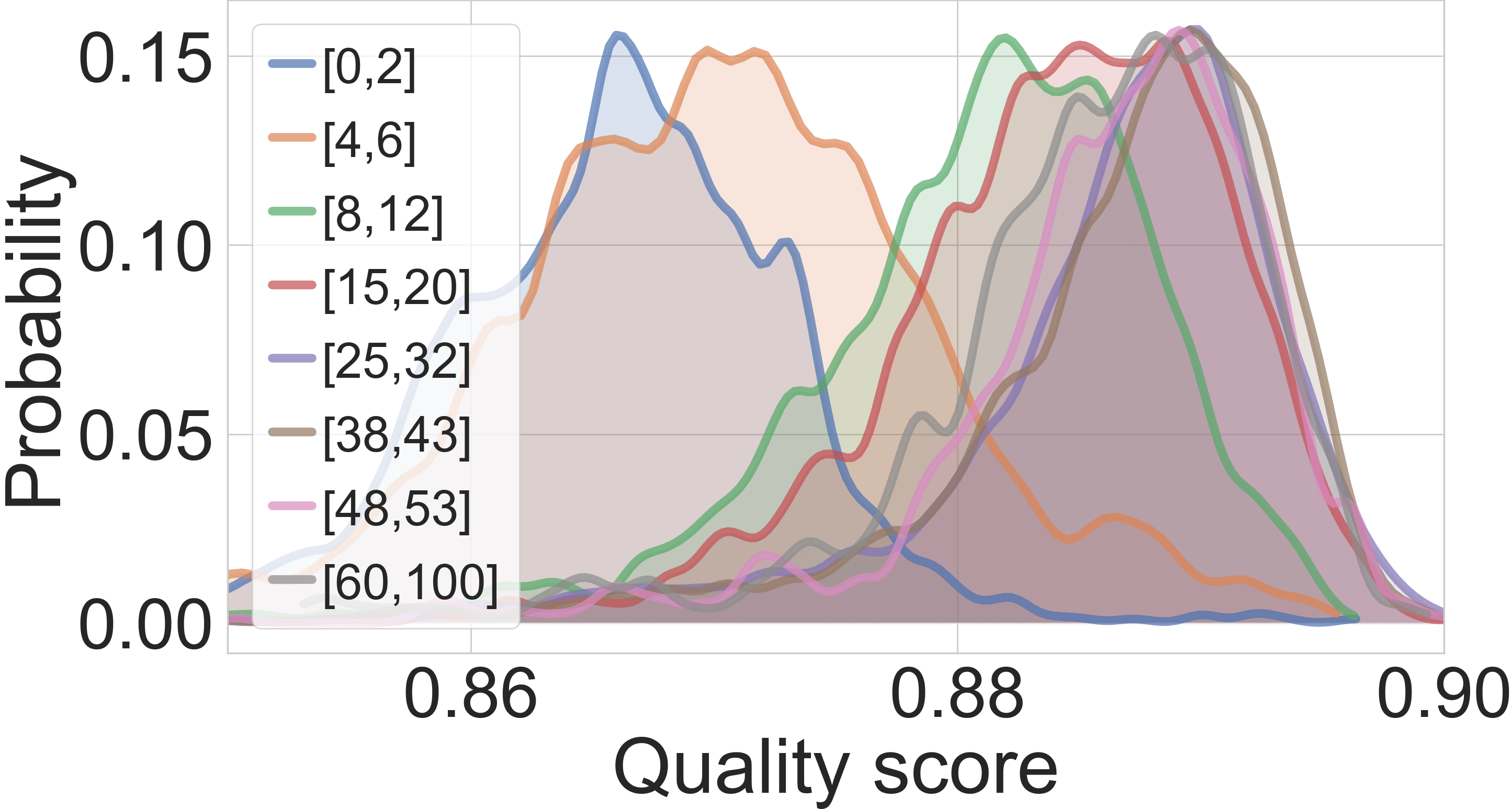}}  
       \vspace{-2mm}                   
\caption{Quality score distributions for several poses (left), ethnicities (middle), and age classes (right). The quality scores are shown for the different face quality assessment approaches. While COTS and FaceQnet work on image level, Best-Rowden and SER-FIQ are applied on FaceNet and ArcFace features.}
\label{fig:QualityDitributions}
\vspace{-3mm}
\end{figure*}

In order to analyse which kind of images will be assigned low face image qualities, Figure \ref{fig:StackedCharts_FaceNet} and \ref{fig:StackedCharts_ArcFace} show an analysis of the proportion of subclasses remaining when applying several quality thresholds.
An unbiased face quality estimator will result in a stable proportion of subclasses over different quality thresholds.
A biased estimator will cause classes effected by the bias to shrink, since these classes are mainly assigned with low quality values.
To get a more detailed understanding of the correlation between quality scores and affected classes, Figure \ref{fig:QualityDitributions} show quality score distributions for the different subclasses.
Based on the experiment before, we observed bias to frontal poses, at asian and black ethnicities and to face images of individuals below 7 years.

\vspace{-2mm}
\paragraph{COTS}
The industry product COTS from Neurotechnology shows a very strong performance in filtering profile face images.
This can be observed with FaceNet and ArcFace embeddings.
The score distributions further show high peaks around the lowest quality values for profile images indicating a weak recognition performance for this pose.
For age, the number of samples of biased age classed affected by the bias are reduced with higher quality thresholds.
The score distributions of the class [0,2] and [4,6] are shifted towards lower qualities.
Consequently, the quality assessment is biased towards age.
For ethnicity, the proportions for the different classes mainly stagnates.
Furthermore, in the corresponding score distributions the distributions show a large overlap in both cases.
Consequently, the face quality is mainly biased towards pose and age.

\vspace{-2mm}
\paragraph{Best-Rowden}
The approach from Best-Rowden shows biased decision towards ethnicity and age.
For different head poses, the quality predictions do not differ and the quality distributions are very similar. 
This can be explained by training on the frontal face database MORPH.
For FaceNet embeddings, a slightly biased behaviour is seen for asian and black faces.
For ArcFace embeddings, a strong bias towards black faces is observable.
Despite training the approach on a database with 80.4\% black ethnics, the major influence comes from the utilized embeddings that were used for the training. 
Both embeddings were trained on MS1M, a database with mainly white ethnicities.
For age, it can be observed that age classes under 12 years are getting lower quality estimates.


\vspace{-2mm}
\paragraph{FaceQnet}
FaceQnet \cite{DBLP:journals/corr/abs-1904-01740} shows a bias in all three investigated cases.
For pose and age, the method reduces the number of samples of the classes affected by the bias showing that also the quality assessment posses the same bias.
This is support by the quality distributions in Figure \ref{fig:QualityDitributions}.
The profile distribution is clearly separated and the distributions for young individuals (till 12 years) are shifted towards smaller quality values.
For ethnicity, the number of samples from the classes effected by bias increases on FaceNet as well as ArcFace embeddings. 
Moreover, the quality score distributions strongly overlap and assign the asian distributions to the highest qualities.
The age-bias can be explained by the used training database VGGFace2, which consists of mainly young adults.
However, this does not explain the quality prediction differences for pose and ethnicity, since VGGFace2 contains more non-frontal than frontal images and contains a large variance of ethnics.
The resulting bias can be better explained by the utilized embeddings. 
The FaceQnet model was trained on comparison scores from FaceNet embeddings based on the MS1M dataset.
MS1M contains mostly frontal faces of white adults.

\vspace{-4mm}
\paragraph{SER-FIQ}
SER-FIQ shows the best face quality assessment performance in all investigated cases (see Chapter \ref{chap:FQA-performance}), since it directly measures the quality based on the deployed face recognition model.
Therefore, it is able to consider the model decision patterns including biased decisions.
This effect can be observed in all evaluated cases for face quality assessment.
In all of these cases, the classes affected by the bias are strongly reduced with a growing quality threshold, while the ratio of the classes with a good face verification performance increases.
This can be observed for frontal head poses, asian and black faces, and faces from individuals below 7 years.
The quality score distributions in Figure \ref{fig:QualityDitributions} further strengthen the suspicion of bias.
In all cases, the distributions are clearly separated from each other.
Consequently, SER-FIQ adapts to the bias from the deployed face recognition model, which arises from the unbalanced MS1M training data.
For non-demographic attributes, a potential bias transfer fulfils the task of quality estimation in a non-discriminative manner.
However, for demographic attributes, SER-FIQ exactly fulfils the utility definition of face quality estimation including a discriminating bias transfer.
Future works have to come up with a solution to this problem.

\vspace{-5mm}
\paragraph{Summary}
For all evaluated face quality assessment algorithms, biased quality estimates are observed.
We point out that if the face quality assessment approach is trained on face embeddings, the major influence of the quality estimation bias was observed to originate from the face embeddings, not the training data.
It was shown that the classes that are affected by face recognition bias are also getting lower quality assignments.
The utility definition of face quality assessment causes this bias transfer and future work have to come up with a solution to this problem.
This (a) might be a development of face quality assessment solution that does not adapt demographic bias or (b) strengthen the focus on bias mitigating face recognition models, since an unintended bias transfer will not happen with an unbiased face recognition model.

\section{Conclusion}
\label{sec:Conclusion}










Current definitions of face quality assessment are based on the suitability of a face image for the task of face recognition.
Optimizing this suitability estimation can be achieved when the face quality assessment is build on the deployed face recognition. 
This leads to more robust and accurate quality predictions as recent work has shown.
However, this can lead to an unintended bias transfer towards the face quality assessment including its discriminatory effects on the society.
In this work, we presented a profound investigation between face recognition bias and face quality estimation.
The experiments were conducted on two publicly available databases and involved four state-of-the-art face quality assessment algorithms from academia and industry and two widely-used face recognition systems.
The results showed that face image quality highly correlates with demographic, as well as non-demographic, bias by demonstrating that current face quality assessment methods already adapted the bias.
Consequently, every enrolment process, as well as quality-based fusion approach, possess the bias as well.
The current definition of face quality allows this bias transfer.
The ethical questions concerning fairness and discrimination that arises with this definition, however, have to be discussed by future work.
Possible solutions for this problem include (a) a development of face quality assessment approach that, by design, prevents a demographic bias transfer or (b) a strong focus on bias mitigating face recognition models, since an unintended bias transfer will not happen with an unbiased face recognition model.

\section*{Acknowledgements}
This research work has been funded by the German Federal Ministry of Education and Research and the Hessen State Ministry for Higher Education, Research and the Arts within their joint support of the National Research Center for Applied Cybersecurity ATHENE.
Portions of the research in this paper use the FERET database of facial images collected under the FERET program, sponsored by the DOD Counterdrug Technology Development Program Office.


{\small
\bibliographystyle{ieee}
\bibliography{egbib}
}

\end{document}